\documentclass[accepted]{uai2024} % after acceptance, for a revised version; 
\usepackage[american]{babel}
\usepackage{siunitx}
\usepackage{graphicx} % Required for inserting images
\usepackage[dvipsnames]{xcolor}
\usepackage{amsmath,bm}
\usepackage[mathscr]{eucal}
\usepackage{float}
\usepackage{subcaption}
\usepackage{tikz}
\usetikzlibrary{tikzmark}
\usepackage{listings}
\usepackage{caption}
\usepackage{tabularx}
\usepackage{adjustbox}
\usepackage{cite}
\usepackage[nosectionbib]{apacite}
\usepackage{natbib}
\usepackage{soul}
\usepackage[title]{appendix}
\usepackage{pdflscape,multicol,blindtext}
\usepackage{afterpage} %prevent blank space after landscape page
\usepackage{array, makecell} 
\usepackage{booktabs} % professional-quality tables
\usepackage{soul}
\usepackage[ruled,vlined]{algorithm2e}

\newcommand{\codepackage}[1]{{\normalfont\ttfamily #1}}

\newcommand{\code}[1]{{\ttfamily\bfseries #1}}

\newcommand{\codebold}[1]{{\ttfamily\bfseries #1}}

\newcommand{\shrink}{\vspace{-2mm}}
\usepackage{todonotes}

\newcommand{\secref}[1]{Section~\ref{#1}}

\title{The Real Deal Behind the Artificial Appeal: \\ Inferential Utility of Tabular Synthetic Data}

\author[$^\ast$,1]{\href{mailto:<alexander.decruyenaere@ugent.be>; <heidelinde.dehaene@ugent.be>?Subject=Your UAI 2024 paper}{\vspace{-0.5em}Alexander Decruyenaere %\thanks{Joint first author} $^\dag$
}{}}
\author[$^\ast$,1]{\href{mailto:<alexander.decruyenaere@ugent.be>; <heidelinde.dehaene@ugent.be>?Subject=Your UAI 2024 paper}{Heidelinde Dehaene %$^\ast$ \thanks{Corresponding author}
}{}}
\author[2]{Paloma Rabaey}
\author[1]{Christiaan Polet}
\author[1]{\\Johan Decruyenaere}
\author[3]{Stijn Vansteelandt}
\author[2]{Thomas Demeester\vspace{-0.5em}}
\affil[$^\ast$]{%
    Joint first authors and corresponding authors
  }  
\affil[1]{%
    Ghent University Hospital -- SYNDARA research group\\
    Belgium
}
\affil[2]{%
    Ghent University -- imec\\
    Belgium
}
\affil[3]{%
    Ghent University\\
    Belgium
  }

\begin{document}
\maketitle

\begin{abstract}
  Recent advances in generative models facilitate the creation of synthetic data to be made available for research in privacy-sensitive contexts. 
  However, the analysis of synthetic data raises a unique set of methodological challenges. In this work, we highlight the importance of inferential utility and provide empirical evidence against naive inference from synthetic data, whereby synthetic data are treated as if they were actually observed. 
  Before publishing synthetic data, it is essential to develop statistical inference tools for such data. By means of a simulation study, we show that the rate of false-positive findings (type 1 error) will be unacceptably high, even when the estimates are unbiased. Despite the use of a previously proposed correction factor, this problem persists for deep generative models, in part due to slower convergence of estimators and resulting underestimation of the true standard error. We further demonstrate our findings through a case study.
\end{abstract}

\section{Introduction}\label{sec:introduction}
Data lie at the core of various disciplines that have a substantial impact on our daily life. Transparency and access to these data are therefore beneficial for an open society. However, alongside great opportunities, great precaution should be taken regarding the possible sensitive nature of these data and related privacy concerns \citep{raghunathan2021synthetic,synthpop2016}. 

Over the last decades, there is an increased awareness that conventional methods for anonymisation or deidentification are insufficient in terms of protecting the privacy and confidentiality of individuals \citep{ohm2009broken, bellovin2019privacy}. Due to numerous well-documented statistical disclosure control failures resulting from the use of these methods, synthetic data are being put forward as an alternative. The idea of creating synthetic data was first proposed by \cite{rubin1993statistical} as an example of multiple imputation and has been further explored, culminating in extensive literature on this topic \citep{raghunathan2003multiple, raghunathan2021synthetic, drechsler2011synthetic, raab2016practical, reiter2005releasing}.
Although the idea itself is thus not novel, the advances in computing power and in the dynamic field of deep generative modelling caused a steep rise in research interest towards synthetic data \citep{raghunathan2021synthetic, drechsler30yearsof, van2023synthetic}. 

Synthetic data are artificial data that (attempt to) mimic the original data in terms of statistical properties, without revealing % any 
individual records \citep{chen2021synthetic}. As such, synthetic data might be able to replace the original data in % statistical
analysis, while preserving the privacy of %the 
individual members of the original data and thereby fulfilling the regulatory privacy constraints \citep{van2023synthetic, zhang2020ensuring, kaloskampis2020synthetic}.
This could enable data sharing with the scientific community and therefore accelerate research, making synthetic data particularly appealing \citep{yan2022multifaceted, van2023synthetic}. 
Synthetic data can be generated using a broad spectrum of methods, ranging from statistical modelling techniques to highly innovative deep learning (DL) techniques such as Generative Adversarial Networks (GANs) and Variational Autoencoders (VAEs) \citep{wan2017variational, yan2022multifaceted, synthpop2016, endres2022synthetic, hernandez2022synthetic}. 
% liu2022goggle, 
These methods have also been extended to offer formal privacy guarantees by imposing differential privacy \citep{algorithmic2014dwork} as an additional constraint during model training \citep{zhang2017privbayes, jordon2018pate, xie2018differentially}.

Consistent in the literature is the conclusion that the trade-off between taking steps to prevent disclosure of the identity of the individuals and preserving the data utility remains \citep{raghunathan2021synthetic}. 
Moreover, there is a wide variation of metrics to assess data utility, % which are 
often related to a quantification of how well the synthetic data resemble the real data or preserve their statistical information (e.g. in terms of distribution, data types, or uni- and bivariate associations%between variables
), also referred to as fidelity, and whether performance and feature selection in modelling tasks are congruent \citep{yan2022multifaceted, el2020seven, ghosheh2022review, kaloskampis2020synthetic}. Within the concept of data utility, we notice that inferential utility is often unmentioned, especially in the DL community \citep{drechsler30yearsof}. Inferential utility captures whether a synthetic sample can be used to obtain valid estimates for a population parameter and to test hypotheses. Therefore, it describes whether one can make valid inferences 
% and generalise to 
concerning the population \citep{raghunathan2021synthetic}. \cite{wilde2021foundations} argue that when synthetic data are used as if they were real data, inferential statements are only related to the synthetic and not the real data generating process, thereby compromising inferential utility. However, they only focus on differentially private synthetic data and state that the fundamental problem of inference is that the synthetic data generating process is misspecified by design, resulting from the additional constraints that are added to guarantee differential privacy. In our work, we will further prove empirically that the problem expands beyond differentially private synthetic data.

%\hd{The work of \cite{wilde2021foundations} focuses on differentially private (DP) synthetic data, postulating that when synthetic data are used as if it were real data, the conclusions are only related to the synthetic and not the real data generating process. They state that the fundamental problem of inference from synthetic data is that the synthetic data generating process is misspecified by design given the additional constrains that are used to guarantee differential privacy.} \pr{In our work, we will further prove empirically that the problem expands beyond differentially private synthetic data.}
% To adress these issues, they adopt a general Bayesian, minimum divergence paradigm for inference. 
%\textcolor{red}{Also \cite{raisa2023noise} focus on DP data and briefly show that the empirical coverage of confidence intervals is below the nominal level and propose a pipeline to generate multiple DP synthetic datasets relying on a Bayesian approach and subsequently conducting inference based on the field of multiple imputation.}
While \cite{drechsler2011synthetic}, \cite{raghunathan2003multiple}, \cite{raghunathan2021synthetic} and \cite{raisa2023noise} developed procedures to obtain valid inferences from \textit{multiple} synthetic datasets, we instead focus on inference from a \textit{single} synthetic dataset, created by both statistical and DL techniques. 
This choice is motivated by previous research showing that the risk of disclosure increases with the number of synthetic datasets % \citep{drechsler30yearsof, reiter2009estimating, drechsler2009disclosure, reiter2010releasing, raab2016practical}. 
\citep{reiter2009estimating, drechsler2009disclosure, klein2015likelihood}.
The work of \cite{raab2016practical} is closely related to the work presented here, since they derive %new 
expressions for the standard error (SE) of an estimator from a \textit{single} synthetic dataset. However, they fail to take into account the implications of the regularisation bias prevalent in DL techniques (i.e.~their bias-variance trade-off being optimised with respect to the prediction error instead of the error in the estimator).

Unfortunately, the regularisation bias introduced by data-adaptive DL techniques to prevent overfitting makes it impossible to guarantee close agreement between all functionals calculated on the real vs. synthetic data, thereby leaving overly optimistic impressions of data utility. Complex functionals involving higher-dimensional associations are arguably more vulnerable to this \citep{van2011targeted}. Moreover, naive SEs calculated on the synthetic data ignore the uncertainty (and regularisation bias) induced by the generative model. While this excess uncertainty and regularisation bias shrink with sample size (at different rates for different techniques \citep{brain1999effect, demystifying2022hines}), they may be large relative to the size of naive SEs at each sample size, resulting in naive confidence intervals that (almost) never contain the population parameter. 
This excess variability is difficult to systematically account for and, as far as we are aware, this has not yet been studied in the context of synthetic data generated by DL techniques.

In this work, we will focus on the \textit{inferential utility} of \textit{tabular} synthetic data. We identify three key contributions. First, we empirically investigate the behaviour of various estimators in terms of bias, SE and their convergence rates when estimated in synthetic data generated by both statistical and DL approaches. Second, we demonstrate how deviations from default behaviour in these properties lead to overly optimistic or even wrong conclusions through an inflation of the type 1 error rate. These issues are especially apparent for DL approaches. 
Finally, we show by means of a simulation and case study that the inferential utility of synthetic data remains compromised despite the use of a correction to the SE, as previously proposed by \cite{raab2016practical}.
Overall, we aim to raise awareness that the current correction factors for the SE of an estimator are not routinely capable of capturing all added variability inherent to synthetic data generated by DL approaches.

The paper is organised as follows. \secref{sec:statproperties} elaborates on the statistical properties that we examine in the context of inferential utility and the corrected SE proposed by \cite{raab2016practical} for estimation with synthetic data. To assess the uncertainty in the estimates obtained from synthetic data and to explore their convergence rate, we conduct a simulation study. We also investigate the impact of deviations from the default behaviour of estimators on null hypothesis significance testing. \secref{sec:Experimental_setup} outlines our experimental setup and the generative models and statistical estimators considered. The findings of the simulation study are summarised and discussed in \secref{sec:simulationstudy}. Finally, to emphasise the relevance of this paper, we illustrate our key contributions by a case study using a well-known dataset in \secref{sec:case_study}.

% \shrink
\section{Evaluating Statistical Properties based on Synthetic Data} \label{sec:statproperties}
% \shrink

In the literature, there is lack of consensus on the metrics that should be applied when evaluating synthetic data because of the complexity of synthetic data and the specific demands each use case has \citep{alaa2022faithful, yan2022multifaceted, van2023synthetic}. 
Most metrics are generally developed with the aim of assessing utility and/or privacy, where e.g. \cite{yan2022multifaceted} proposed a benchmarking framework that incorporates both facets. 
However, in this work, we investigate the impact of estimating population parameters from synthetic data, which may no longer have the same inferential utility when they are estimated as if the data were really observed. 
We deem it important to stress that our purpose is not to propose another utility measure, but rather to evaluate the inferential utility itself for different generative models. More specifically, focus lies on studying the validity of estimators that are well-established on original data, but remain understudied in synthetic data, especially when these data are created by a DL approach.

The \textit{estimands} considered in our simulation study range from the population mean to various regression coefficients.
Commonly used estimators for these estimands (such as the sample mean and logistic regression coefficients) are further referred to as \textit{estimators}, the obtained values of those estimators in a specific sample as \textit{estimates}.

% \shrink
\subsection{Large Sample Behaviour of Estimators} \label{subsec:asymptotics}
% \shrink

The purpose of the simulation study in Section \ref{sec:simulationstudy} is to evaluate the quality of estimators of finite-dimensional parameters calculated on synthetic data. 
When an inferential statement is made, we rely on the test statistic and its properties to obtain a \textit{p}-value. The formula for a test statistic is typically made up of the estimate divided by its SE. Therefore,
%we start with evaluating the estimate (and its corresponding bias) and the standard error of the estimator. 
%Specifically, 
we look at %their 
the empirical \textbf{bias} and \textbf{standard error (SE)} of the estimator by means of a simulation study, and how these evolve with increasing sample size. In standard statistical analyses, both are supposed to diminish as the sample size tends to infinity.
It is typically seen that the \textbf{convergence rate} of the SE is of the order $1/\sqrt{n}$ while the bias converges faster \citep{lehmann2006theory}. Such estimators are called roughly $\sqrt{n}$-consistent (with $n$ referring to the size of the original data). When this is not the case, standard statistical inference will be compromised. Therefore, we first assess how the estimators behave in terms of bias and SE when estimated in synthetic data. 
As indicated in \secref{sec:introduction}, we foresee atypical behaviour for the estimators given the additional variability inherent to the generation process of the synthetic data and the regularisation bias, which should ideally be accounted for \citep{brain1999effect}. 
Hence, we subsequently map the inferential repercussions of this deviant behaviour in the context of null hypothesis testing by quantifying the empirical \textbf{type 1 error rate}, i.e.~the probability to find a significant effect when in truth there is none, and the empirical \textbf{power}, i.e.~the probability to find a true significant effect.

% \shrink
\subsection{Minimal Correction for Estimation}
\label{subsec:correctionfactor}
% \shrink

Even when synthetic data are generated based on a correct statistical model (i.e.~without model misspecification) and without data-adaptive modelling (unlike DL methods), the regular expressions for the SE of estimators are insufficient when used in synthetic data. 
Within a statistical approach, 
% When a statistical approach is applied to create synthetic data, 
the original sample is used to obtain a parametric representation of the dependency structure of the data.
Based on these representations, synthetic %data 
samples are generated 
and then 
% which are then 
used to estimate the %desired 
population parameters. When it is silently assumed that synthetic data can be treated as real data, as would be the case if standard expressions for the SE are
% would be 
used, this will lead to an underestimation of the SE due to ignoring the uncertainty in the generation process.
%of synthetic data.

We foresee that the added variability of estimation based on synthetic data will diminish when the sample size of the original training sample increases. Intuitively, there is more uncertainty (and thus model variability) when a synthetic sample of e.g. 200 instances is created based on an original sample of 100 vs.~\num{100000} observations. Therefore, in the absence of both model misspecification and data-adaptive modelling, this added variability in a statistical approach will not induce large sample bias and will decrease with increasing sample size \citep{vansteelandt2022assumption}.
Consequently, when using a (pre-specified) parametric statistical approach to create synthetic data (where the size of the synthetic data is a fixed fraction $\in~]0,1]$ of $n$) and an estimator that is $\sqrt{n}$-consistent on the original data, it is expected that the estimators will remain unbiased and $\sqrt{n}$-consistent (relative to the original data).
It should be emphasised that this behaviour will not occur when synthetic data are created with a DL approach, due to the added variability and regularisation bias (with the latter not present in parametric statistical approaches) converging at slower rates.

In Appendix \ref{sec:APPENDIX_derivation}, we provide an analytic derivation for a correction to the SE that is valid with a single synthetic dataset for any $\sqrt{n}$-consistent estimator $\hat{\theta}$. Originally proposed by \cite{raab2016practical}, %but applied to our setting, 
the corrected SE is defined as follows:
%this corrected SE estimator coincides with the adaptation proposed by  and equals}
\begin{equation}\label{correction_SE}
\sigma_{\hat{\theta}\text{, corrected}} = \sigma_{\hat{\theta}\text{, naive}} \sqrt{1+ \frac{m}{n}},
\end{equation}
where $\sigma_{\hat{\theta}\text{, naive}}$ is the model-based SE of the estimator $\hat{\theta}$ in the synthetic data, $m$ the sample size of the synthetic data, and $n$ the sample size of the original data. This adaptation to the model-based SE will henceforth be referred to as the corrected SE.
%It is important to stress that this is a \textit{minimal} correction since it only applies to $\sqrt{n}$-consistent estimators, hence the added variability resulting from the regularisation bias is not accounted for. Therefore, it will not be sufficient when synthetic data are created with a DL approach. Looking at the proof provided in Appendix \ref{sec:APPENDIX_derivation}, this is reflected in the explicit assumption that the estimator is $\sqrt{n}$-consistent when used in the original data. }
It is important to stress that the proof assumes a $\sqrt{n}$-consistent estimator when used in the original data.
%It is important to stress that the proof in Appendix \ref{sec:APPENDIX_derivation} assumes that the estimator is $\sqrt{n}$-consistent when used in the original data. 
% For this reason, 
Therefore, Equation (\ref{correction_SE}) %presents 
is a \textit{minimal} correction which does not account for the added variability resulting from the regularisation bias, making it insufficient when synthetic data are created using a DL approach. While \cite{raab2016practical} implicitly assume $\sqrt{n}$-consistency and do not provide a formal derivation for the corrected SE with a single synthetic dataset, 
%of the corrected SE, 
our contribution is to make this assumption explicit through a formal derivation, as well as to show empirically that it indeed does not sufficiently correct the SE in cases where $\sqrt{n}$-consistency cannot be guaranteed.

\section{Experimental Setup} \label{sec:Experimental_setup}
% \shrink

To increase awareness that traditional statistical analyses, as well as corrected alternatives like the one discussed in the previous section, may fail when applied to synthetic data, we developed a general framework that will be used on toy data (simulation study in \secref{sec:simulationstudy}) and real-world data (case study on Adult Census Income dataset in \secref{sec:case_study}). 
In this section, we elaborate on this framework and the generators used to create synthetic data.

% \shrink
\subsection{General framework}\label{subsec:Setup_framework}
% \shrink

\begin{figure*}[t]
	\centering
    \includegraphics[width=0.85\textwidth]{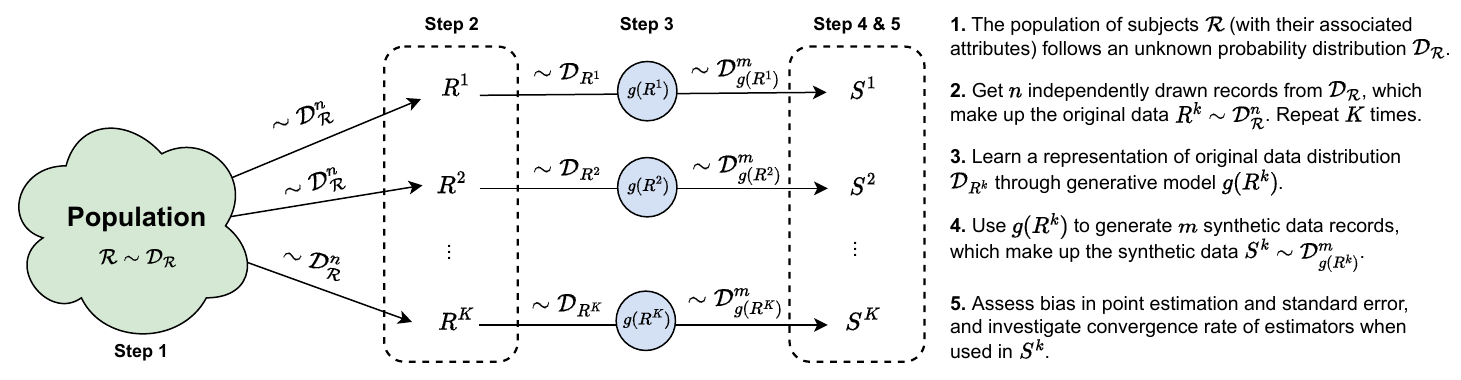}
    % \vspace{.1in}
    % \shrink
	\caption{General experimental framework, applied in both the simulation study and case study.}
	\label{fig:experimental_setup}
\end{figure*}

Some notation is introduced for the remainder of this paper, in line with the notation used in \cite{stadler2022synthetic}. A visualisation of the framework is given in Figure \ref{fig:experimental_setup}. We are interested in $\mathscr{R}$, a population of subjects where each data record $\bm{r} \in \mathscr{R}$ encompasses information on $p$ variables: %or attributes: 
$\bm{r} = (r_1, \ldots, r_p)$. These data follow an unknown joint probability distribution, % indicated by 
$\mathscr{R}\sim \mathscr{D}_\mathscr{R}$.
In reality, this population cannot be observed and hence a random sample is taken. We refer to this observed original data as $R\sim \mathscr{D}_\mathscr{R}^n$, which are $n$ independent data records from $ \mathscr{D}_\mathscr{R}$, and these original data define the data distribution $\mathscr{D}_R$. This process is repeated $K$ times, further referred to as $K$ Monte Carlo runs. 

Next, in the process of generating tabular synthetic data, a generative model will aim to learn a representation of the joint probability distribution $\mathscr{D}_R$ based on the original data $R$. Formally, a model training algorithm will learn a representation of the distribution $\mathscr{D}_R$, which is then denoted as $\mathscr{D}_{g(R)}$, and outputs a trained, yet stochastic generative model $g(R)$. 
Subsequently, synthetic data are generated based on this model $g(R)$. Synthetic data records are distributed according to $\mathscr{D}_{g(R)}$ and form the synthetic dataset $S = (\bm{s}_1, \ldots, \bm{s_m})$ of size $m$, $S \sim \mathscr{D}_{g(R)}^m$.

Finally, we examine the performance of diverse estimators when estimated in the $K$ %obtained 
synthetic datasets. This includes an evaluation of their bias and SE, their convergence rates, and some inferential metrics typically seen in null hypothesis testing (i.e.~type 1 error rate and power).

% \shrink
\subsection{Synthetic Data Generators}\label{subsec:Setup_generators}
% \shrink

We chose diverse data generation methods that are representative in terms of use and that enable us to examine the impact of added variability in the generation process and the extra layer of complexity due to regularisation bias.
Following the categorisation suggested in \cite{hernandez2022synthetic}, we study both statistical (classical) approaches and DL approaches.
In the following, a compact description of the applied methods is given, but a more detailed explanation of all these methods can be found in Appendix \ref{subsec:APPENDIX_sim_synthmethods}.

The first implementation of a statistical approach, named \codebold{Synthpop}, relies on the \codepackage{Synthpop} package for \codepackage{R}, which provides a routine to generate synthetic data \citep{synthpop2016}. This framework encompasses both parametric and non-parametric methods to sequentially fit a series of conditional distributions, based on the observed data. We restrict ourselves to the default parametric method and provide information of the dependency structure of our data via specification of a directed acyclic graph (DAG). This defines the order of the sequence and which variables need to be included as predictors in the conditional models.

A second and third implementation of a statistical approach are based on Bayesian Networks (BNs) \citep{pearl2011BN}. We opted to implement both a method where the dependency structure was pre-specified by the user through a DAG (\codebold{BN DAG}) and a method where the DAG was estimated automatically using the Chow-Liu algorithm \citep{chowliu} (\codebold{BN}).
We refer to both methods as statistical % or parametric 
approaches given that they rely on the Maximum Likelihood Estimator to estimate the conditional probability distributions. However, \codebold{BN} includes a data-adaptive component since the structure of the Bayesian Network is learned non-parametrically via the Chow-Liu algorithm.

Two commonly used generative DL approaches are GANs \citep{goodfellow2014generative} and VAEs \citep{kingma2013auto}. We decided to focus on these methods, as they have been specifically adapted towards tabular data \citep{xu2019modeling}, and have been frequently used in recent literature on synthetic data %generation
\citep{van2023synthetic, assefa2021finance, tao2021benchmarking, bourou2021tabular, rajabi2022tabfairgan, figueira2022survey, chale2022intrusion, liu2022goggle, el2024evaluation, akiya2024comparison}. 
Tabular data impose several challenges to the design of generative models, such as mixed data types, non-normality, and highly imbalanced categorical variables. To overcome these difficulties, \cite{xu2019modeling} propose \codebold{CTGAN} and \codebold{TVAE}.
Training details for all DL approaches, including hyperparameter tuning, can be found in Appendix \ref{subsubsection:APPENDIX_sim_synthmethods_DL_hpo}. To study the effect of hyperparameter tuning on the inferential utility of %the generated 
synthetic data, we also considered untuned versions, further denoted as \codebold{Default CTGAN} and \codebold{Default TVAE}, where the hyperparameters were set to their default values as suggested by the \codepackage{Synthcity} library \citep{Synthcity}. 

% several privacy-focused generators

Both privacy and utility are essential for synthetic data and their trade-off should be optimised. In our study, inferential utility was the starting point and as such, we did not formally define and assess privacy. However, by imposing differential privacy as an additional constraint during model training, some
generative models provide formal privacy protection guarantees. Complementary to the approaches described above, we additionally study three state-of-the-art privacy-focused generators, i.e.~\codebold{PrivBayes} \citep{zhang2017privbayes}, \codebold{DP-GAN} \citep{xie2018differentially}, and \codebold{PATE-GAN} \citep{jordon2018pate}. All results pertaining to this class of generative models are presented in Appendix \ref{sec:app_dp_methods}, confirming that the conclusions of our study extend to privacy-focused generators.

% \adc{In the main analysis, we did not include methods that focus specifically on privacy since these also need to tackle this additional layer of complexity. In Appendix \ref{sec:app_dp_methods}, we present the results for three state-of-the-art differentially private models, i.e.~\codebold{PrivBayes} \citep{zhang2017privbayes}, \codebold{DP-GAN} \citep{xie2018differentially}, and \codebold{PATE-GAN} \citep{jordon2018pate}, noting that this did not impact the conclusions of our study.}

% \shrink
\section{Simulation Study} \label{sec:simulationstudy}
% \shrink

In order to assess the finite sample performance of the different estimators and the correction for the SE proposed by \cite{raab2016practical} in the context of synthetic data, a Monte Carlo simulation study is performed. 
In line with the experimental setup from \secref{subsec:Setup_framework}, the sections below introduce the data on population level (\secref{subsec:sim_data}), and list the experimental details of our simulation study, after which we dive into the results (\secref{sec:results}).

% \shrink
\subsection{Data generating mechanism} \label{subsec:sim_data}
% \shrink

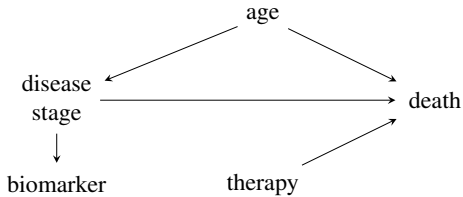
\begin{figure}[t]
\centering
\resizebox{.37\textwidth}{!}{
\begin{tikzpicture}[
    baseline=(current bounding box.north), % top alignment with A., B., etc.
    ->,
	>=stealth, % arrow head style
	node distance = 0.5cm and 2cm, % distance between nodes (vertical and horizontal)
	shorten >= 2pt, % don't touch arrow head to node
	pil/.style={ % define arrow style
		->,
		thick, % linestyle
		shorten = 2pt,}
	]
    \node[align=center] (D) {disease\\stage};
    \node[above right=of D] (A) {age};
	\node[below=of D] (B) {biomarker};
	\node[right=of A] (M) at (A |- D) {death};
	\node[] (T) at (A |- B) {therapy};  %aligned with both A and B
	\draw [->] (A) to (D);
	\draw [->] (A) to (M);
	\draw [->] (D) to (M);
	\draw [->] (D) to (B);
	\draw [->] (T) to (M);
\end{tikzpicture}
}
% \vspace{.1in}
\caption{ DAG for
% Illustration of the relationship between 
the variables in the simulation study.}
\label{DAG_simulated_data}
\end{figure}

We opted to work with low-dimensional tabular data given their frequent use in applied medical research. % Different regression models that are commonly used, 
Commonly used regression models were taken into account when choosing the nature of the variables. We included %wish to have 
a mix of continuous (normally distributed or skewed), binary, and ordinal variables.
To obtain these requirements and reflect a generic clinical setting, the data generating process consists of the following five variables: \emph{age} (continuous with a normal distribution), \emph{disease stage} (ordinal with four categories), \emph{biomarker} (continuous with a skewed distribution), \emph{therapy} (binary), and \emph{death} (binary).
% This design is a simplification of reality since we assume that there are no missing data and we do not consider the data as longitudinal.
A DAG representing the dependency structure is shown in Figure \ref{DAG_simulated_data}. We refer to Appendix \ref{sec:app_data_gen_mechanism} for the exact data generating mechanism. 

% \shrink
\subsection{Specifications and Results} \label{sec:results}
% \shrink

In line with Figure \ref{fig:experimental_setup}, we simulate $n$ independent records from the population $\mathscr{R}$ that form the observed original data $R$. This process is repeated $K$ = 200 times, with the sample size $n$ varying log-uniformly between $50$ and $5000$ (i.e.~$n \in \{50, 160, 500, 1600, 5000\}$). Per generation method as introduced in Section \ref{subsec:Setup_generators}, a generative model $g(R^k)$ is trained, from which $m$ synthetic data records are sampled. In our study, we set $m = n$ to retain the dimensionality of the original data and to facilitate an equal comparison between original and synthetic data. This process results in 200 synthetic datasets $S$ for each of the generator methods and each value of $n$.

We then evaluate a variety of statistical estimators in these synthetic datasets. 
Motivated by commonly used analyses in applied medical research, and the variety of mixed data types in our setup, we opted to work with the following estimators: mean, proportion, and regression coefficients from a main effects proportional odds cumulative logit model (effect of \emph{age} on \emph{disease stage}), a main effects gamma regression model (effect of \emph{disease stage} on \emph{biomarker}), and a main effects binomial logistic regression model (effect of \emph{age}, \emph{disease stage} and \emph{therapy} on \emph{death}). 

We now present the results of our simulation study. In Section \ref{subsec:results_quality} we evaluate the quality of our synthetic data. Then, Section \ref{subsec:bias_SE} investigates the bias and SE of the estimators, after which Section \ref{subsec:convergence_rate} discusses their convergence rate. Finally, Section \ref{subsec:NHST_results} addresses the impact of the atypical behaviour of the estimators on null hypothesis testing.
% The code to recreate all results can be found on our Github repository\footnote{\url{github.com/syndara-lab/inferential-utility}}. 
The code to reproduce all results is available on Github: \url{https://github.com/syndara-lab/inferential-utility}.
% \footnote{\url{github.com/syndara-lab/inferential-utility}}. 

% \shrink
\subsubsection{Quality of Synthetic Data}
\label{subsec:results_quality}
% \shrink

The average inverse of the Kullback–Leibler divergence (IKLD) is often used to assess the statistical similarity between distributions \citep{espinosa_2023}. This metric was used to tune the hyperparameters of \code{CTGAN} and \code{TVAE}. The average IKLD over all Monte Carlo runs is presented per generator in Table \ref{table:appendix_IKLD} in the appendix, where the tuned \code{CTGAN} and \code{TVAE} have higher IKLD than their default versions. However, the statistical approaches still seem to perform slightly better. Additional analyses on synthetic data quality are included in Appendix \ref{sec:app_quality_synth_data}.

% \shrink
\subsubsection{Bias and Standard Error} \label{subsec:bias_SE}
% \shrink

\begin{figure}[t]
    \centering
    \includegraphics[width=0.45\textwidth]{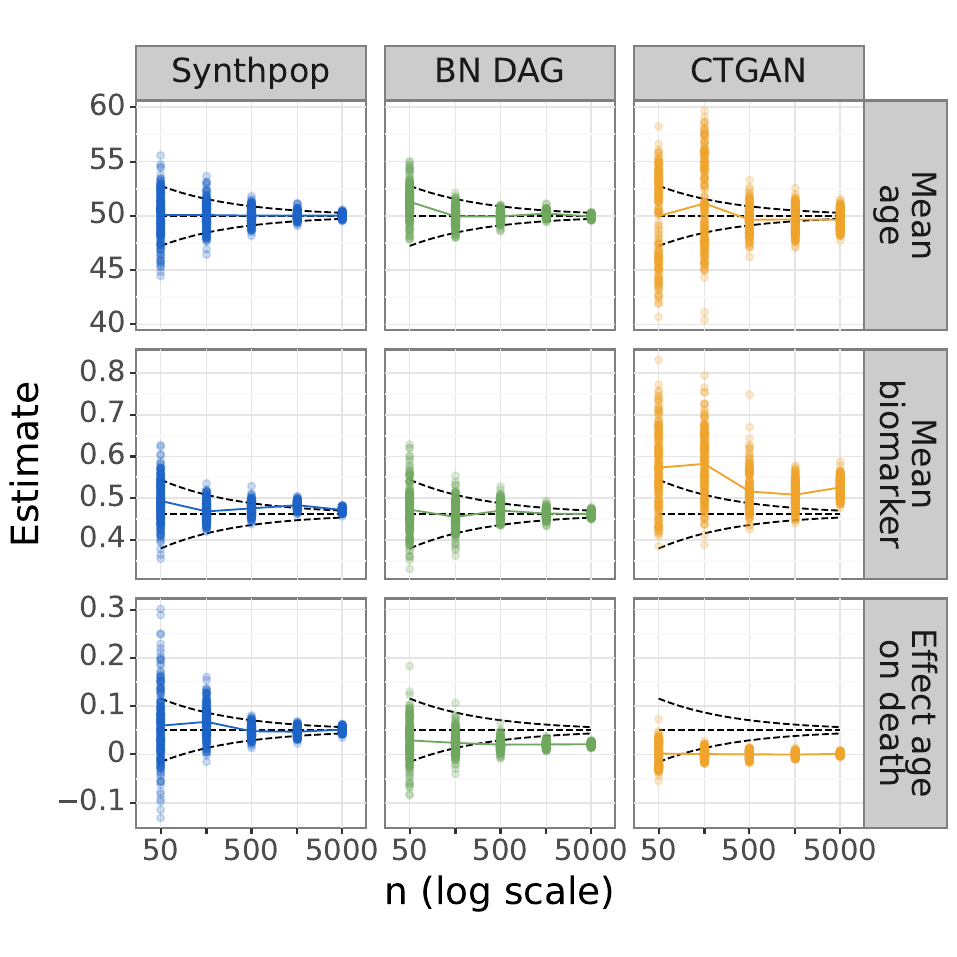}
    %{plot_bias_PAPER.pdf}	
    % \shrink
    \caption{The horizontal dashed line represents the population parameter and each dot is an estimate per Monte Carlo run (200 dots in total per value of $n$).
    % Results for a selection of estimators and generators. Each dot is an estimate per Monte Carlo run (200 dots in total per value of $n$). The population parameter is represented by the horizontal dashed line. 
    The dashed funnel indicates the behaviour of an unbiased and $\sqrt{n}$-consistent estimator based on observed data.}
    \label{fig:bias_funnel}
\end{figure}

% The estimates are depicted in Figure \ref{fig:bias_funnel} for a selection of estimators 
Figure \ref{fig:bias_funnel} depicts the estimates for a selection of estimators 
(the sample mean of $age$ and $biomarker$, and the logistic regression coefficient of $age$ on $death$) and generators (\code{Synthpop}, \code{BN DAG} and \code{CTGAN}). Figures \ref{appendix:bias_plot_1} and \ref{appendix:bias_plot_2} in the appendix show these for all estimators and generators. Each dot is an estimate per Monte Carlo run and the true population parameters are represented by the horizontal dashed line. This figure allows a qualitative assessment of two key properties of estimators: %based on synthetic data
\textbf{empirical bias} (i.e.~the average difference between the estimates and the population parameter, as represented by the solid line) and \textbf{empirical SE} (i.e.~the standard deviation of the estimates, as indicated by the vertical spread of the estimates). Table \ref{table:results_relative_error} lists the same information numerically, summarising the relative bias ($\text{RE}_{\hat{\theta}}$) and the relative underestimation of the empirical SE by the naive model-based SE ($\text{RE}_{\hat{\sigma}_{\hat{\theta}}}$). Tables \ref{tab:app_RE_all_stat} and \ref{tab:app_RE_all_DL} in the appendix show these metrics for all estimators and generators.

Ideally, both the bias and SE converge to zero as the sample size grows larger. The convergence rate conveys the rate at which this happens. The funnel in Figure \ref{fig:bias_funnel} represents the default behaviour of an unbiased estimator based on original data of which the SE diminishes at a rate of $1/\sqrt{n}$. We observe % in this figure 
that the bias and SE of estimators based on synthetic data often deviate from this default behaviour, the extent of which differs between generative models.

First, generative model misspecification will introduce bias. This is for example the case with \code{Synthpop}, where the sample mean of $biomarker$ consistently overestimates the population mean. $Biomarker$ is gamma-distributed in the original data, but \code{Synthpop} fails to reconstruct this marginal distribution since it uses (by default) an ordinary least squares regression model during the generation process, leading to reconstruction error. Generative model misspecification also occurs for the logistic regression coefficients based on synthetic data generated by \code{BN DAG}, resulting in bias towards the null effect. This arises from the full discretisation of the continuous variable \emph{age}, along with discrete variables \emph{stage} and \emph{therapy}, during the conditional generation of \emph{death}, introducing non-negligible overfitting bias in the latter. By contrast, the parsimonious nature of the Chow-Liu algorithm applied in \code{BN} provides some immunity against the impact of this discretisation on overfitting.
\code{CTGAN}, despite being more flexible yet tuned to prevent overfitting, also fails to adequately fit the joint distribution in our simulation study, leading to bias for several estimators including a biased null effect of $age$ on $death$. We also observe some bias for the other DL approaches in Table \ref{tab:app_RE_all_DL} in the appendix. Note that this might be partially attributed to the fact that these methods do not receive prior knowledge on the dependencies between the variables, while \code{Synthpop} and \code{BN DAG} do. Further, this model misspecification is not captured well by the IKLD metric described in \secref{subsec:results_quality}. Moreover, the model misspecification for \code{CTGAN} could even directly result from the tuning objective being based on the average IKLD, because this will not prioritise the preservation of multivariate relations as the divergences are only calculated per single variable and then averaged across all variables.

Second, the empirical SEs are larger for synthetic data than for original data and may vary over generative models. Larger SEs reflect the additional (predictive) uncertainty in the generation of synthetic data, which seems most pronounced with DL approaches. This uncertainty is discarded in the naive use of model-based SEs, leading to underestimation of the empirical SE, as is evident from Table \ref{table:results_relative_error}. 

\begin{table*}[t]
    \caption{\label{table:results_relative_error}  
    Relative error (RE) for \code{Synthpop}, \code{BN DAG} and \code{CTGAN} for a selection of estimators, averaged over 200 Monte Carlo runs. $\text{RE}_{\hat{\theta}}$ is the relative bias of the estimates $\hat{\theta}$. $\text{RE}_{\hat{\sigma}_{\hat{\theta}}}$ is the RE between the naive model-based ($\hat{\sigma}_{\hat{\theta}, naive})$ and the empirical standard error. Positive and negative values indicate a relative over- and underestimation. } 
    %\begin{tabular}{lrr|rr|rr|rr}
    % \shrink \shrink
    \begin{center}
    \begin{adjustbox}{max width=\textwidth}
    \begin{tabular}{lcccccccccccc}
    \toprule
    & \multicolumn{4}{c}{\textbf{Synthpop}} & \multicolumn{4}{c}{\textbf{BN DAG}} & \multicolumn{4}{c}{\textbf{CTGAN}} \\ 
    {} & \multicolumn{2}{c}{$\text{RE}_{\hat{\theta}}$ (\%)} & \multicolumn{2}{c}{$\text{RE}_{\hat{\sigma}_{\hat{\theta}}}$ (\%)} & \multicolumn{2}{c}{$\text{RE}_{\hat{\theta}}$ (\%)} & \multicolumn{2}{c}{$\text{RE}_{\hat{\sigma}_{\hat{\theta}}}$ (\%)} & \multicolumn{2}{c}{$\text{RE}_{\hat{\theta}}$ (\%)} & \multicolumn{2}{c}{$\text{RE}_{\hat{\sigma}_{\hat{\theta}}}$ (\%)} \\
     \textbf{Estimator}  & $n=50$ & $n=5000$ & $n=50$ & $n=5000$ &  $n=50$ & $n=5000$ &  $n=50$ & $n=5000$ &  $n=50$ & $n=5000$ &  $n=50$ & $n=5000$ \\
    \midrule
    Mean age & 0.15 & 0.03 & -40.31 & -26.55 & 2.59 & -0.10 & -6.17 & -2.32 & -0.03 & -0.73 & -46.38 & -78.85 \\
    Mean biomarker & 6.86 & 2.16 & -14.23 & 1.16 & 2.32 & 0.05 & -21.31 & -26.83 & 24.14 & 13.87 & -35.77 & -76.00 \\[0.15cm]
    Proportion therapy & 0.18 & -0.12 & -33.26 & -31.47 & -25.28 & -0.93 & 7.00 & -4.56 & -0.42 & -0.18 & 131.41 & -56.79 \\
    Proportion death & 31.39 & -2.35 & -25.40 & -14.07 & 7.27 & 3.23 & 45.38 & 25.23 & 6.04 & 5.29 & -9.52 & -47.44 \\[0.15cm]
    Effect age on death & 19.07 & 2.35 & -33.90 & -30.04 & -41.76 & -57.81 & -9.98 & -3.13 & -96.42 & -97.81 & -9.41 & -1.18 \\
    Effect therapy on death & 38.53 & -2.74 & -32.78 & -30.45 & -50.16 & -55.24 & -11.63 & -15.83 & -104.72 & -101.02 & 17.50 & -13.89 \\
    \bottomrule
    \bottomrule
    \end{tabular}
    \end{adjustbox}
    \end{center}
\end{table*}

\shrink
\subsubsection{Convergence Rate} \label{subsec:convergence_rate}

\begin{table}[t]
    \caption{\label{table:results_convergence} Estimated exponent $a$ for the power law convergence rate $n^{-a}$ for empirical bias and standard error (SE).
    }
    % \shrink \shrink
    \begin{center}
    \resizebox{0.475\textwidth}{!}{
    \begin{tabular}{lcccc}
    \toprule
    && \multicolumn{3}{c}{\textbf{Generator}} \\ 
    \cmidrule(r){3-5} 
     &  \textbf{Original} &  \textbf{Synthpop} & \makecell[t]{\textbf{BN DAG}} &   \textbf{CTGAN}  \\
    \textbf{Estimator, bias/SE} &  &  &  & \\
    \midrule
    Mean age &      0.64 / 0.49 &  0.38 / 0.53 &          0.45 / 0.49 &  -0.42 / 0.40 \\
    Mean biomarker &      0.47 / 0.48 &  0.10 / 0.51 &          0.81 / 0.49 &   0.18 / 0.34 \\[0.15cm] 
    Proportion therapy &      0.42 / 0.50 &  0.12 / 0.50 &          0.81 / 0.46 &   0.10 / 0.19 \\
    Proportion death &      1.24 / 0.51 &  0.45 / 0.52 &          0.02 / 0.48 &   0.04 / 0.41 \\[0.15cm] 
    Effect age on death  &      0.76 / 0.56 &  0.52 / 0.59 &         -0.07 / 0.57 &  -0.00 / 0.47 \\
    Effect therapy on death &      0.70 / 0.53 &  0.52 / 0.56 &         -0.02 / 0.53 &   0.00 / 0.51 \\    
    
    \bottomrule
    \bottomrule
    \end{tabular}
    }
    \end{center}
\end{table}

Assuming a power law $n^{-a}$ in convergence rate for the empirical bias and SE, we estimated the exponent $a$ from five logarithmically spaced sample sizes $n$ between 50 and 5000, shown for a selection of estimators in Table \ref{table:results_convergence}. The %full 
results (with 95\% confidence interval) for all estimators and generators can be found in Tables \ref{tab:app_convergence_rate_SE} and \ref{tab:app_convergence_rate_bias} and Figure \ref{appendix:plot_convergence_se} in the appendix. Standard statistical analysis assumes that the bias converges faster than the SE with the latter diminishing at a rate of $1/\sqrt{n}$. The corrected SE proposed by \cite{raab2016practical}, though taking into account the added variability of the synthetic data generating process, still relies on the same assumption, thus remaining invalid for deviating convergence rates.

As shown in the table, the empirical SE of estimators based on original data indeed converges at a rate of $1/\sqrt{n}$ (i.e.~$a_{SE}\approx0.5$). Fully parametric generative models are also expected to yield estimators of which the SE decreases at a rate of $1/\sqrt{n}$. This seems confirmed by \code{Synthpop}, \code{BN DAG}, and \code{BN}. %However, the SEs produced by and \code{BN} converge a bit slower (i.e.~$a_{SE}<0.5$), as non-parametric components are built into their generation process.
By contrast, the SEs produced by the more data-adaptive DL approaches converge much slower (i.e.~$a_{SE}<<0.5$). The slower-than-$\sqrt{n}$-convergence leads to a progressively increasing underestimation of the empirical SE by the model-based SE (which assumes $\sqrt{n}$-convergence) as the sample size grows larger, as seen in Table \ref{table:results_relative_error}. Furthermore, 
% both Table \ref{table:results_convergence} and Tables \ref{tab:app_convergence_rate_SE} \& \ref{tab:app_convergence_rate_bias} indicate that, 
as opposed to default behaviour, the bias converges slower than the SE ($a_{bias} \le a_{SE}$) for some estimators and generators. This is problematic, as elaborated on in the next section.

% \shrink
\subsubsection{Null Hypothesis Significance Testing} \label{subsec:NHST_results}
% \shrink

\begin{figure}[t]
    \centering
    \includegraphics[width=0.45\textwidth]{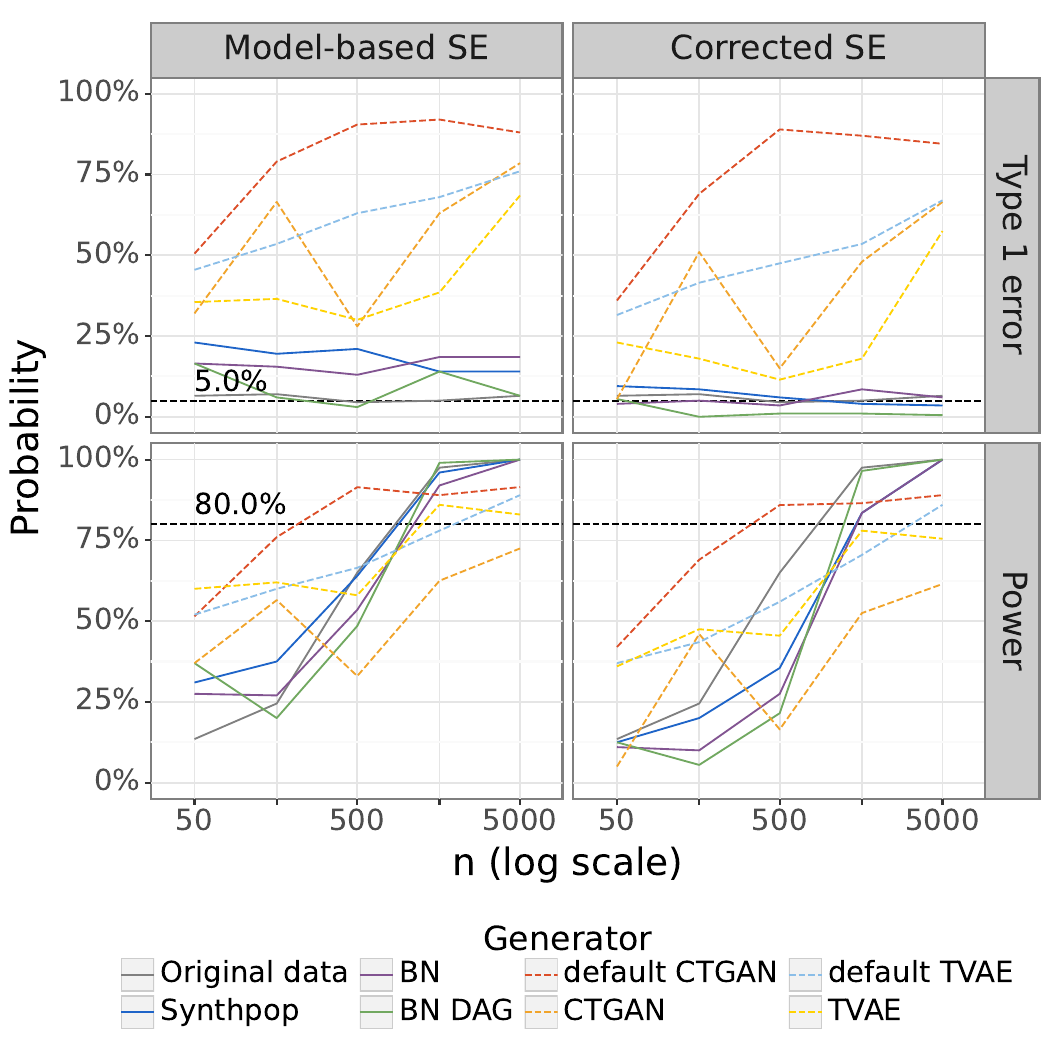}	
    % {plot_typeI_II_PAPER.pdf}	
    % \shrink
    \caption{Type 1 error rate and power of a one-sample t-test at $\alpha=5\%$ for the population mean of $age$ with naive model-based and corrected standard errors (SEs). %(using the correction factor proposed in Section \ref{subsec:correctionfactor}). 
    %For the type 1 error rate, the null hypothesis states that the population mean of $age$ is equal to the ground truth, whereas for the power, it states that the population mean of $age$ is equal to $98\%$ of the ground truth.
    }
    \label{fig:typeI_II}
\end{figure}

The null hypothesis significance testing (NHST) framework typically uses an estimate divided by its associated (un)certainty (reflected by the SE) as test statistic to reject a null hypothesis. We foresee the following problems with NHST on synthetic data. First, if the bias of the estimator % based on synthetic data 
converges slower than its SE, the test statistic will become more extreme, thereby increasing the type 1 error rate. Second, even if the bias converges faster than the SE, the naive model-based SEs %obtained from synthetic data 
are still optimistically small, also inflating the type 1 error rate. Both scenarios are highly concerning, since they will lead to a flurry of false-positive findings. Acknowledgement of the extra uncertainty 
% of the data generating process 
by using a larger yet valid SE will control the type 1 error rate at the nominal level. 
In turn, this will decrease the power, reflecting the loss of information when working with synthetic data. 

Figure \ref{fig:typeI_II} shows the empirical type 1 error rate and power of a one-sample t-test at $\alpha=5\%$ for the population mean of $age$, separately tested with naive model-based SEs and corrected SEs as suggested by \cite{raab2016practical}.
For the type 1 error rate, the null hypothesis states that the population mean of $age$ is equal to the ground truth, whereas for the power, it states that the population mean of $age$ is equal to $98\%$ of the ground truth. 

Hypothesis tests with naive SEs lead to type 1 error rates larger than $5\%$: the more the empirical SE is underestimated by the naive model-based SE (as is especially the case for the DL approaches with increasing $n$), the larger the inflation of the type 1 error rate. Use of corrected SEs will control the type 1 error rate at approximately $5\%$, but only for statistical approaches.
However, this comes with a loss of power, a trade-off most pronounced in small sample sizes.
%\delete{Finally,}
More importantly, the corrected SE does not sufficiently account for the predictive uncertainty of the DL approaches, so the type 1 error rate remains uncontrolled for. It is essential to highlight that the low inferential utility of synthetic data generated by DL approaches will be observed regardless of whether the estimator is biased, so this limitation cannot be explained in terms of bias or poor performance of one DL approach.

% \shrink
\section{Case study} \label{sec:case_study}
% \shrink

To illustrate our findings and their implications for the applied researcher, we perform a case study on the Adult Census Income dataset \citep{misc_adult_2} following the framework discussed in \secref{sec:Experimental_setup}. The dataset with \num{45222} complete cases constitutes our \textit{population}. We assume that the researcher only has access to a limited \textit{sample} of \num{5000} observations. In order to % be able to 
share their data without privacy issues, the researcher generates a synthetic dataset, with $n$ = $m$ = \num{5000}, once using the statistical method \code{Synthpop}, and once by training a \code{Default CTGAN}. 
For this % specific 
case study, it was computationally too intensive to create synthetic values for %categorical 
variables with a large number of categories based on a parametric method in \code{Synthpop}. Therefore, we used the non-parametric method \textit{CART} for these unordered categorical variables.
%\delete{, rather than the default method \textit{polytomous logistic regression}}.
Additional analyses on synthetic data quality are included in Section \ref{sec:app_quality_synth_data_case_study} of the appendix.
 
\begin{figure}[t]
\centering
    \includegraphics[width=0.45\textwidth]{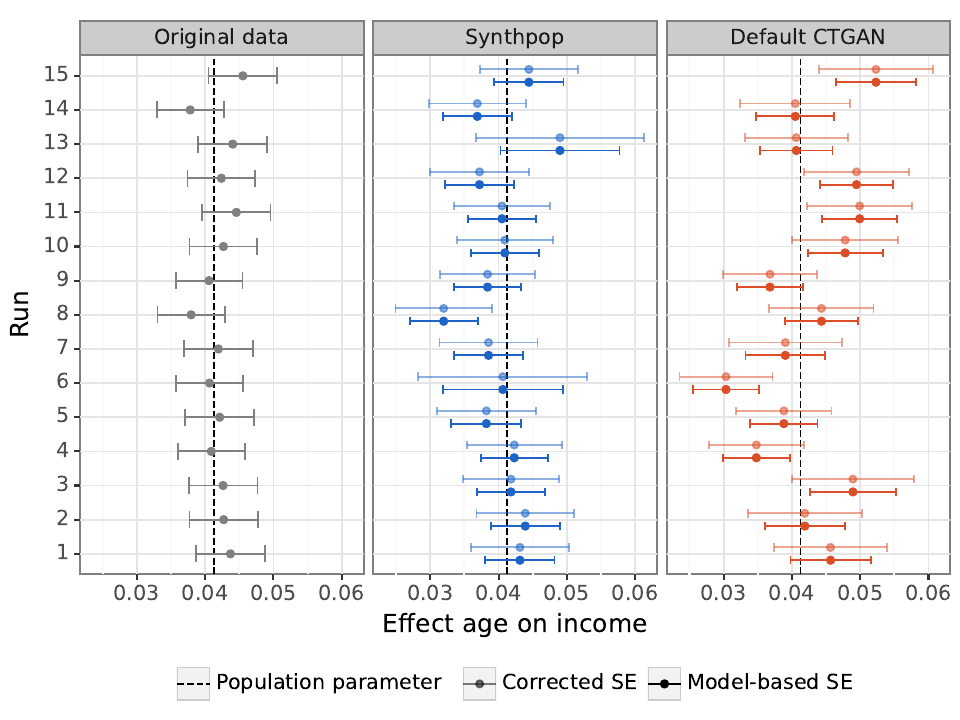}
    % \shrink
    \caption{Empirical coverage of $95\%$ confidence intervals for effect of $age$ on $income$, with model-based and corrected standard error (SE).}
    \label{fig:case_study_coverage}
\end{figure}

For simplicity, we assume the researcher's interest lies in inferring the population mean of $age$, and the effect of $age$ on $income$ (estimated via a logistic regression model) from a single synthetic dataset. When an estimate for these targets is obtained, an inferential statement can be made by using a \num{95}\% confidence interval (CI). A CI for the population parameter is constructed by using the estimate and its model-based SE obtained from that synthetic set of \num{5000} instances, in our case resulting in \num{200} CIs. %In theory, 
If we repeated the construction of CIs infinitely, %an infinite amount of times, 
then \num{95}\% of the \num{95}\% CIs should by definition cover the population parameter. Figure \ref{fig:case_study_coverage} depicts the first \num{15} CIs obtained from both original and synthetic samples for the effect of $age$ on $income$. The vertical dashed lines represent the true parameter value as obtained from the \textit{population}. Comparing the point estimates with this dashed line, one can see that the estimates obtained in the synthetic samples are positioned around the true population parameter, but that the variability is much higher than in the original samples. This is in accordance with the results from the simulation study and hence endorses the claim that the SE should incorporate the extra variability caused by the synthetic data generation process. Table \ref{table:results_case_study_RE} also shows that the SE for all estimators is now highly underestimated when estimated in the synthetic samples.

\begin{table}[t] \caption{\label{table:results_case_study_RE}
    Results for the case study:
    % Case study results, averaged over \num{200} Monte Carlo runs: 
    relative error (RE) for the estimates $\hat{\theta}$ and the model-based standard errors (SEs) $\hat{\sigma}_{\hat{\theta}}$, and empirical coverage of \num{95}\% confidence intervals with the model-based (Cov) and corrected SE ($\text{Cov}_{\text{corr}}$) (in \%).}
%\shrink \shrink
\begin{center}
\scalebox{0.68}{
\begin{tabular}[b]{lcccccc}
      \toprule
 &  \multicolumn{2}{c}{\textbf{Original}} &  \multicolumn{2}{c}{\textbf{Synthpop}} &  \multicolumn{2}{c}{\textbf{\makecell[t]{Default\\CTGAN}}} \\
\textbf{Estimator} & $\text{RE}_{\hat{\theta}}$ & $\text{RE}_{\hat{\sigma}_{\hat{\theta}}}$ & $\text{RE}_{\hat{\theta}}$ & $\text{RE}_{\hat{\sigma}_{\hat{\theta}}}$ & $\text{RE}_{\hat{\theta}}$ & $\text{RE}_{\hat{\sigma}_{\hat{\theta}}}$ \\
\midrule
Mean age& 0.09 &  4.76 & -0.03 & -28.98 & -1.58 & -93.65 \\[0.15cm]
 \makecell[l]{Effect age on income} &  -0.09 & 20.39 & -1.96 & -50.11 & 3.87 & -56.51 \\
\midrule
\midrule
\textbf{Estimator} & Cov & \makecell[t]{$\text{Cov}_{\text{corr}}$} & 
Cov & \makecell[t]{$\text{Cov}_{\text{corr}}$} & 
Cov & \makecell[t]{$\text{Cov}_{\text{corr}}$} \\
\midrule
Mean age& 95.50 & -- & 83.50 & 95.00 & 12.18 & 13.20 \\[0.15cm]
 \makecell[l]{Effect age on income} &  98.00 & -- & 70.50 & 87.00 & 59.39 & 77.16 \\
\bottomrule
\bottomrule
    \end{tabular}}
\end{center}
\end{table}

More strikingly, we find that for synthetic data created with \code{Default CTGAN}, only 8 out of 15 CIs depicted in Figure \ref{fig:case_study_coverage} contain the true parameter value. This is also quantified by the low empirical coverage levels reported in Table \ref{table:results_case_study_RE}, and is even more pronounced for the mean of $age$ (see Figure \ref{fig:adult_dataset} in the appendix). Combining all results from Table \ref{table:results_case_study_RE} (i.e.~limited bias in point estimations but substantial underestimation of the SE and thus low empirical coverage levels), we can state that naive CIs based on a synthetic sample are too narrow (or \textit{permissive}), hence overestimating the confidence about the estimated mean of $age$ and effect of $age$ on $income$ in the \textit{population}.
Using the corrected SE improves the coverage to some extent, as seen in Figure \ref{fig:case_study_coverage} and Table \ref{table:results_case_study_RE}, but this is still highly insufficient for synthetic data created with \code{Default CTGAN}.
For \code{Synthpop}, we observe that the correction works properly on the estimator for the population mean of \textit{age}, but fails to fully elevate the empirical coverage to the nominal level for the effect of \textit{age} on \textit{income}. These mixed results were expected, given that \textit{age} was a root node and therefore synthesised based on bootstrap samples. Conversely, \textit{income} was synthesised based on all other previously 
 (non-parametrically) synthesised variables. 
 %(including non-parametric generation methods due to the computational burden mentioned earlier). 
 This reinforces our claim that the corrected SE should be seen as a minimal correction, depending on the used synthetic data generator.

% \shrink
\section{Discussion}
% \shrink

We conducted a simulation study to quantify how the naive use of statistical estimators in synthetic data (which silently assumes that these data can be treated as if directly observed) compromises the inferential utility. We studied both statistical approaches and deep learning techniques. We empirically confirmed that increased variability in the estimates leads to an underestimation of their standard error, supporting previous claims by \cite{raab2016practical}. Our simulation study moreover revealed the slower-than-expected convergence rate of both bias and standard error, which was most pronounced for deep learning approaches.

We also tested the corrected standard error as proposed by \cite{raab2016practical} for applications with only one synthetic dataset, and demonstrated that this adaptation does not sufficiently capture all added variability in the case of deep learning approaches. We argued that this is due to an extra layer of complexity introduced by their regularisation bias, which cannot readily be expressed analytically and therefore remains unaccounted for. Furthermore, the corrected standard error relies on the classical assumptions concerning the root-$n$ convergence rate of estimators, which were shown to be unfulfilled for the deep learning approaches studied in this work. The fact that these approaches fail to offer the specific guarantee of root-$n$ convergence implies that the corrected standard error will never approximate the empirical standard error, even when increasing the synthetic sample size $m\rightarrow+\infty$. Therefore, at present, this renders them not useful for statistical inference, despite their flexibility allowing them to better approximate a more complex joint distribution of the original data.

The impact of these deviations from default behaviour becomes apparent when evaluating metrics % that are essential 
in the context of null hypothesis testing, often the prime interest of applied researchers. A naive use of synthetic data leads to an inflation of false-positive findings, which can be controlled for to some extent with the use of the corrected standard error, though only for data generated using parametric statistical approaches. This comes at a cost in terms of power loss, which is an inevitable trade-off.
These practical implications were further cemented in our case study. 

The broader implications of this work for an applied researcher come into play at every stage of employing synthetic data. % positioned on different levels. 
First of all, a reader who consults (published) work that is based on synthetic data should interpret naive analysis with caution. As empirically proven in this paper, standard confidence intervals and p-values obtained on synthetic data may drastically underestimate the uncertainty in synthetic data. 
% \delete{as these ignore the size of the original data. 
% Indeed, synthetic data obtained via generators trained on small datasets will unsurprisingly deliver much worse quality than synthetic data obtained via generators trained on large datasets. Standard confidence intervals and p-values ignore this uncertainty, as they do not distinguish whether the data is synthetic or real.}
% \delete{We can now make the bridge to the second researcher,} 
Second, the analyst who 
% \delete{aims to perform analyses on synthetic data that was provided externally. They} 
only has access to the synthetic data should use an adequate correction for the standard error of an estimator when making inferential statements. However, we have shown that the current correction factors are not capable of capturing all added variability inherent to synthetic data generated by deep learning approaches. We deem it difficult to obtain a 
% \delete{practical and} 
generic correction for deep learning approaches, since the uncertainty associated with their regularisation bias cannot readily be expressed analytically.
Therefore, we can conclude that the %\delete{third researcher} 
original data holder who creates synthetic data, 
% \delete{should reflect about the goal of the synthetic data and how to work towards this goal} 
must know what analysis will be done on these data. If the goal is inference, we advise to use a parametric generation method, since the corrected standard errors are proven to be sufficient only in these settings.

Limitations of our study include the low-dimensional setting, for which deep learning approaches might be less suited. Still, the fact that deviant behaviour is already observed for a wide range of statistical estimators given the use of such a simple data generating mechanism, raises questions about what can be expected in larger-scale applications. We especially notice that deep learning approaches fall short. Our study did not cover more recently developed deep generative models, such as diffusion-based models or even large language models. While these models are popular, they are less well-established in the domain of tabular synthetic data than \codebold{CTGAN} and \codebold{TVAE}. Still, we expect the same problems to occur, since all deep learning approaches are designed to optimally balance bias and variance only w.r.t. a chosen criterion (like prediction error). Therefore, none of them can guarantee that an optimal trade-off is made simultaneously w.r.t. the mean squared error for all possible estimators. 

%\delete{Since we deem it to be impossible to train a deep learning model that would learn the observed data distribution in all possible dimensions, future work must focus on the targeted training of generative models for intended use cases of the synthetic data. In addition, if synthetic data are to be published to enable data sharing and accelerate hypothesis testing, constructing valid standard errors will be necessary.}

To improve %fully leverage 
the inferential utility of synthetic data created by deep learning approaches, we propose the following ideas for future research. 
%First, inferential utility related objectives could be tackled during model training, for example by adding a regularisation objective to the loss function that reduces the bias in a particular target estimator on a per-batch basis.
First, building on insights from the literature on debiased and targeted machine learning \citep{van_der_laan_targeted_2011, chernozhukov_double/debiased_2018}, there may be potential to eliminate bias by targeting synthetic data generators towards the considered statistical analysis.
% Second, insights from the literature on debiased and targeted machine learning \citep{van_der_laan_targeted_2011, chernozhukov_double/debiased_2018} could be applied to optimise the bias-variance trade-off in the targeted estimator. 
Second, importance weighting methods could be developed, similarly to what \cite{ghalebikesabi2022mitigating} proposed in the context of noise-related bias in differentially private synthetic datasets.

\newpage

% \paragraph{What is title case?}
% \href{https://en.wikipedia.org/wiki/Title_case}{Wikipedia} explains:
% \begin{quote}
%     Title case or headline case is a style of capitalization used for rendering the titles of published works or works of art in English.
%     When using title case, all words are capitalized except for ‘minor’ words (typically articles, short prepositions, and some conjunctions) unless they are the first or last word of the title.
% \end{quote}

% ------------ BACK MATTER ------------

% \begin{contributions} % will be removed in pdf for initial submission 
% % (without ‘accepted’ option in \documentclass)
% % so you can already fill it to test with the
% % ‘accepted’ class option
%     Briefly list author contributions. 
%     This is a nice way of making clear who did what and to give proper credit.
%     This section is optional.
% \end{contributions}

\begin{acknowledgements} % will be removed in pdf for initial submission,
% (without ‘accepted’ option in \documentclass)
% so you can already fill it to test with the
% ‘accepted’ class option
This research was funded by a grant received from the Fund for Innovation and Clinical Research of Ghent University Hospital. Paloma Rabaey's research is funded by the Research Foundation Flanders (FWO-Vlaanderen) with grant number 1170122N. This research also received funding from the Flemish government under the “Onderzoeksprogramma Artificiële Intelligentie (AI) Vlaanderen” programme.

\end{acknowledgements}

% ------------ REFERENCES ------------
% Here you have your bibliography created

%\bibliographystyle{apalike}  %apalike,phdbib.bst
% \renewcommand{\bibsection}{}
% \section*{References}
\bibliography{UAI_manuscript}

\begin{thebibliography}{}

\bibitem [\protect \citeauthoryear {%
Akiba%
, Sano%
, Yanase%
, Ohta%
\BCBL {}\ \BBA {} Koyama%
}{%
Akiba%
\ \protect \BOthers {.}}{%
{\protect \APACyear {2019}}%
}]{%
optuna_2019}
\APACinsertmetastar {%
optuna_2019}%
\begin{APACrefauthors}%
Akiba, T.%
, Sano, S.%
, Yanase, T.%
, Ohta, T.%
\BCBL {}\ \BBA {} Koyama, M.%
\end{APACrefauthors}%
\unskip\
\newblock
\APACrefYearMonthDay{2019}{}{}.
\newblock
{\BBOQ}\APACrefatitle {{Optuna: A Next-generation Hyperparameter Optimization Framework}} {{Optuna: A Next-generation Hyperparameter Optimization Framework}}.{\BBCQ}
\newblock
\BIn{} \APACrefbtitle {{Proceedings of the 25th {ACM} {SIGKDD} International Conference on Knowledge Discovery and Data Mining}.} {{Proceedings of the 25th {ACM} {SIGKDD} International Conference on Knowledge Discovery and Data Mining}.}
\PrintBackRefs{\CurrentBib}

\bibitem [\protect \citeauthoryear {%
Akiya%
, Ishihara%
\BCBL {}\ \BBA {} Yamamoto%
}{%
Akiya%
\ \protect \BOthers {.}}{%
{\protect \APACyear {2024}}%
}]{%
akiya2024comparison}
\APACinsertmetastar {%
akiya2024comparison}%
\begin{APACrefauthors}%
Akiya, I.%
, Ishihara, T.%
\BCBL {}\ \BBA {} Yamamoto, K.%
\end{APACrefauthors}%
\unskip\
\newblock
\APACrefYearMonthDay{2024}{}{}.
\newblock
{\BBOQ}\APACrefatitle {A Comparison of Synthetic Data Generation Techniques for Control Group Survival Data in Oncology Clinical Trials: Simulation Study.} {A comparison of synthetic data generation techniques for control group survival data in oncology clinical trials: Simulation study.}{\BBCQ}
\newblock
\APACjournalVolNumPages{JMIR Medical Informatics}{}{}{}.
\PrintBackRefs{\CurrentBib}

\bibitem [\protect \citeauthoryear {%
Alaa%
, Van~Breugel%
, Saveliev%
\BCBL {}\ \BBA {} van~der Schaar%
}{%
Alaa%
\ \protect \BOthers {.}}{%
{\protect \APACyear {2022}}%
}]{%
alaa2022faithful}
\APACinsertmetastar {%
alaa2022faithful}%
\begin{APACrefauthors}%
Alaa, A.%
, Van~Breugel, B.%
, Saveliev, E\BPBI S.%
\BCBL {}\ \BBA {} van~der Schaar, M.%
\end{APACrefauthors}%
\unskip\
\newblock
\APACrefYearMonthDay{2022}{}{}.
\newblock
{\BBOQ}\APACrefatitle {{How faithful is your synthetic data? Sample-level metrics for evaluating and auditing generative models}} {{How faithful is your synthetic data? Sample-level metrics for evaluating and auditing generative models}}.{\BBCQ}
\newblock
\BIn{} \APACrefbtitle {{International Conference on Machine Learning}} {{International Conference on Machine Learning}}\ (\BPGS\ 290--306).
\PrintBackRefs{\CurrentBib}

\bibitem [\protect \citeauthoryear {%
Ankan%
\ \BBA {} Panda%
}{%
Ankan%
\ \BBA {} Panda%
}{%
{\protect \APACyear {2015}}%
}]{%
ankan2015pgmpy}
\APACinsertmetastar {%
ankan2015pgmpy}%
\begin{APACrefauthors}%
Ankan, A.%
\BCBT {}\ \BBA {} Panda, A.%
\end{APACrefauthors}%
\unskip\
\newblock
\APACrefYearMonthDay{2015}{}{}.
\newblock
{\BBOQ}\APACrefatitle {pgmpy: Probabilistic graphical models using {Python}} {pgmpy: Probabilistic graphical models using {Python}}.{\BBCQ}
\newblock
\BIn{} \APACrefbtitle {{Proceedings of the 14th Python in Science Conference (SCIPY 2015)}.} {{Proceedings of the 14th Python in Science Conference (SCIPY 2015)}.}
\PrintBackRefs{\CurrentBib}

\bibitem [\protect \citeauthoryear {%
Assefa%
\ \protect \BOthers {.}}{%
Assefa%
\ \protect \BOthers {.}}{%
{\protect \APACyear {2021}}%
}]{%
assefa2021finance}
\APACinsertmetastar {%
assefa2021finance}%
\begin{APACrefauthors}%
Assefa, S\BPBI A.%
, Dervovic, D.%
, Mahfouz, M.%
, Tillman, R\BPBI E.%
, Reddy, P.%
\BCBL {}\ \BBA {} Veloso, M.%
\end{APACrefauthors}%
\unskip\
\newblock
\APACrefYearMonthDay{2021}{}{}.
\newblock
{\BBOQ}\APACrefatitle {{Generating Synthetic Data in Finance: Opportunities, Challenges and Pitfalls}} {{Generating Synthetic Data in Finance: Opportunities, Challenges and Pitfalls}}.{\BBCQ}
\newblock
\BIn{} \APACrefbtitle {{Proceedings of the First ACM International Conference on AI in Finance}.} {{Proceedings of the First ACM International Conference on AI in Finance}.}
\newblock
\APACaddressPublisher{}{Association for Computing Machinery}.
\PrintBackRefs{\CurrentBib}

\bibitem [\protect \citeauthoryear {%
Becker%
\ \BBA {} Kohavi%
}{%
Becker%
\ \BBA {} Kohavi%
}{%
{\protect \APACyear {1996}}%
}]{%
misc_adult_2}
\APACinsertmetastar {%
misc_adult_2}%
\begin{APACrefauthors}%
Becker, B.%
\BCBT {}\ \BBA {} Kohavi, R.%
\end{APACrefauthors}%
\unskip\
\newblock
\APACrefYearMonthDay{1996}{}{}.
\newblock
\APACrefbtitle {{Adult}.} {{Adult}.}
\newblock
\APAChowpublished {UCI Machine Learning Repository}.
\PrintBackRefs{\CurrentBib}

\bibitem [\protect \citeauthoryear {%
Bellovin%
, Dutta%
\BCBL {}\ \BBA {} Reitinger%
}{%
Bellovin%
\ \protect \BOthers {.}}{%
{\protect \APACyear {2019}}%
}]{%
bellovin2019privacy}
\APACinsertmetastar {%
bellovin2019privacy}%
\begin{APACrefauthors}%
Bellovin, S\BPBI M.%
, Dutta, P\BPBI K.%
\BCBL {}\ \BBA {} Reitinger, N.%
\end{APACrefauthors}%
\unskip\
\newblock
\APACrefYearMonthDay{2019}{}{}.
\newblock
{\BBOQ}\APACrefatitle {Privacy and synthetic datasets} {Privacy and synthetic datasets}.{\BBCQ}
\newblock
\APACjournalVolNumPages{Stanford Technology Law Review}{22}{}{1}.
\PrintBackRefs{\CurrentBib}

\bibitem [\protect \citeauthoryear {%
Bickel%
, Klaassen%
, Ritov%
\BCBL {}\ \BBA {} Wellner%
}{%
Bickel%
\ \protect \BOthers {.}}{%
{\protect \APACyear {1993}}%
}]{%
brickel1993efficient}
\APACinsertmetastar {%
brickel1993efficient}%
\begin{APACrefauthors}%
Bickel, P.%
, Klaassen, C.%
, Ritov, Y.%
\BCBL {}\ \BBA {} Wellner, J.%
\end{APACrefauthors}%
\unskip\
\newblock
\APACrefYear{1993}.
\newblock
\APACrefbtitle {{Efficient and Adaptive Estimation for Semiparametric Models}} {{Efficient and Adaptive Estimation for Semiparametric Models}}\ (\BVOL~4).
\newblock
\APACaddressPublisher{}{Baltimore Johns Hopkins University Press}.
\PrintBackRefs{\CurrentBib}

\bibitem [\protect \citeauthoryear {%
Bourou%
, El~Saer%
, Velivasaki%
, Voulkidis%
\BCBL {}\ \BBA {} Zahariadis%
}{%
Bourou%
\ \protect \BOthers {.}}{%
{\protect \APACyear {2021}}%
}]{%
bourou2021tabular}
\APACinsertmetastar {%
bourou2021tabular}%
\begin{APACrefauthors}%
Bourou, S.%
, El~Saer, A.%
, Velivasaki, T.%
, Voulkidis, A.%
\BCBL {}\ \BBA {} Zahariadis, T.%
\end{APACrefauthors}%
\unskip\
\newblock
\APACrefYearMonthDay{2021}{}{}.
\newblock
{\BBOQ}\APACrefatitle {{A Review of Tabular Data Synthesis Using {GANs} on an {IDS} Dataset}} {{A Review of Tabular Data Synthesis Using {GANs} on an {IDS} Dataset}}.{\BBCQ}
\newblock
\APACjournalVolNumPages{Information}{12}{}{375}.
\PrintBackRefs{\CurrentBib}

\bibitem [\protect \citeauthoryear {%
Brain%
\ \BBA {} Webb%
}{%
Brain%
\ \BBA {} Webb%
}{%
{\protect \APACyear {1999}}%
}]{%
brain1999effect}
\APACinsertmetastar {%
brain1999effect}%
\begin{APACrefauthors}%
Brain, D.%
\BCBT {}\ \BBA {} Webb, G\BPBI I.%
\end{APACrefauthors}%
\unskip\
\newblock
\APACrefYearMonthDay{1999}{}{}.
\newblock
{\BBOQ}\APACrefatitle {On the effect of data set size on bias and variance in classification learning} {On the effect of data set size on bias and variance in classification learning}.{\BBCQ}
\newblock
\BIn{} \APACrefbtitle {{Proceedings of the Fourth Australian Knowledge Acquisition Workshop, University of New South Wales}} {{Proceedings of the Fourth Australian Knowledge Acquisition Workshop, University of New South Wales}}\ (\BPGS\ 117--128).
\PrintBackRefs{\CurrentBib}

\bibitem [\protect \citeauthoryear {%
Chalé%
\ \BBA {} Bastian%
}{%
Chalé%
\ \BBA {} Bastian%
}{%
{\protect \APACyear {2022}}%
}]{%
chale2022intrusion}
\APACinsertmetastar {%
chale2022intrusion}%
\begin{APACrefauthors}%
Chalé, M.%
\BCBT {}\ \BBA {} Bastian, N\BPBI D.%
\end{APACrefauthors}%
\unskip\
\newblock
\APACrefYearMonthDay{2022}{}{}.
\newblock
{\BBOQ}\APACrefatitle {Generating realistic cyber data for training and evaluating machine learning classifiers for network intrusion detection systems} {Generating realistic cyber data for training and evaluating machine learning classifiers for network intrusion detection systems}.{\BBCQ}
\newblock
\APACjournalVolNumPages{Expert Systems with Applications}{207}{}{117936}.
\PrintBackRefs{\CurrentBib}

\bibitem [\protect \citeauthoryear {%
Chen%
, Lu%
, Chen%
, Williamson%
\BCBL {}\ \BBA {} Mahmood%
}{%
Chen%
\ \protect \BOthers {.}}{%
{\protect \APACyear {2021}}%
}]{%
chen2021synthetic}
\APACinsertmetastar {%
chen2021synthetic}%
\begin{APACrefauthors}%
Chen, R.%
, Lu, M.%
, Chen, T.%
, Williamson, D.%
\BCBL {}\ \BBA {} Mahmood, F.%
\end{APACrefauthors}%
\unskip\
\newblock
\APACrefYearMonthDay{2021}{}{}.
\newblock
{\BBOQ}\APACrefatitle {Synthetic data in machine learning for medicine and healthcare} {Synthetic data in machine learning for medicine and healthcare}.{\BBCQ}
\newblock
\APACjournalVolNumPages{Nature Biomedical Engineering}{5}{}{1-5}.
\PrintBackRefs{\CurrentBib}

\bibitem [\protect \citeauthoryear {%
Chernozhukov%
\ \protect \BOthers {.}}{%
Chernozhukov%
\ \protect \BOthers {.}}{%
{\protect \APACyear {2018}}%
}]{%
chernozhukov_double/debiased_2018}
\APACinsertmetastar {%
chernozhukov_double/debiased_2018}%
\begin{APACrefauthors}%
Chernozhukov, V.%
, Chetverikov, D.%
, Demirer, M.%
, Duflo, E.%
, Hansen, C.%
, Newey, W.%
\BCBL {}\ \BBA {} Robins, J.%
\end{APACrefauthors}%
\unskip\
\newblock
\APACrefYearMonthDay{2018}{}{}.
\newblock
{\BBOQ}\APACrefatitle {{Double/debiased machine learning for treatment and structural parameters}} {{Double/debiased machine learning for treatment and structural parameters}}.{\BBCQ}
\newblock
\APACjournalVolNumPages{The Econometrics Journal}{21}{1}{C1-C68}.
\newblock
\begin{APACrefDOI} \doi{10.1111/ectj.12097} \end{APACrefDOI}
\PrintBackRefs{\CurrentBib}

\bibitem [\protect \citeauthoryear {%
Chow%
\ \BBA {} Liu%
}{%
Chow%
\ \BBA {} Liu%
}{%
{\protect \APACyear {1968}}%
}]{%
chowliu}
\APACinsertmetastar {%
chowliu}%
\begin{APACrefauthors}%
Chow, C.%
\BCBT {}\ \BBA {} Liu, C.%
\end{APACrefauthors}%
\unskip\
\newblock
\APACrefYearMonthDay{1968}{}{}.
\newblock
{\BBOQ}\APACrefatitle {Approximating discrete probability distributions with dependence trees} {Approximating discrete probability distributions with dependence trees}.{\BBCQ}
\newblock
\APACjournalVolNumPages{IEEE Transactions on Information Theory}{14}{3}{462-467}.
\PrintBackRefs{\CurrentBib}

\bibitem [\protect \citeauthoryear {%
Drechsler%
}{%
Drechsler%
}{%
{\protect \APACyear {2011}}%
}]{%
drechsler2011synthetic}
\APACinsertmetastar {%
drechsler2011synthetic}%
\begin{APACrefauthors}%
Drechsler, J.%
\end{APACrefauthors}%
\unskip\
\newblock
\APACrefYear{2011}.
\newblock
\APACrefbtitle {Synthetic datasets for statistical disclosure control: theory and implementation} {Synthetic datasets for statistical disclosure control: theory and implementation}\ (\BVOL~201).
\newblock
\APACaddressPublisher{}{Springer Science \& Business Media}.
\PrintBackRefs{\CurrentBib}

\bibitem [\protect \citeauthoryear {%
Drechsler%
\ \BBA {} Haensch%
}{%
Drechsler%
\ \BBA {} Haensch%
}{%
{\protect \APACyear {2023}}%
}]{%
drechsler30yearsof}
\APACinsertmetastar {%
drechsler30yearsof}%
\begin{APACrefauthors}%
Drechsler, J.%
\BCBT {}\ \BBA {} Haensch, A\BHBI C.%
\end{APACrefauthors}%
\unskip\
\newblock
\APACrefYearMonthDay{2023}{}{}.
\newblock
{\BBOQ}\APACrefatitle {30 Years of Synthetic Data} {30 years of synthetic data}.{\BBCQ}
\newblock
\APACjournalVolNumPages{arXiv preprint arXiv:2304.02107}{}{}{}.
\PrintBackRefs{\CurrentBib}

\bibitem [\protect \citeauthoryear {%
Drechsler%
\ \BBA {} Reiter%
}{%
Drechsler%
\ \BBA {} Reiter%
}{%
{\protect \APACyear {2009}}%
}]{%
drechsler2009disclosure}
\APACinsertmetastar {%
drechsler2009disclosure}%
\begin{APACrefauthors}%
Drechsler, J.%
\BCBT {}\ \BBA {} Reiter, J.%
\end{APACrefauthors}%
\unskip\
\newblock
\APACrefYearMonthDay{2009}{}{}.
\newblock
{\BBOQ}\APACrefatitle {Disclosure risk and data utility for partially synthetic data: An empirical study using the German IAB Establishment Survey} {Disclosure risk and data utility for partially synthetic data: An empirical study using the german iab establishment survey}.{\BBCQ}
\newblock
\APACjournalVolNumPages{Journal of Official Statistics}{25}{4}{589--603}.
\PrintBackRefs{\CurrentBib}

\bibitem [\protect \citeauthoryear {%
Dwork%
\ \BBA {} Roth%
}{%
Dwork%
\ \BBA {} Roth%
}{%
{\protect \APACyear {2014}}%
}]{%
algorithmic2014dwork}
\APACinsertmetastar {%
algorithmic2014dwork}%
\begin{APACrefauthors}%
Dwork, C.%
\BCBT {}\ \BBA {} Roth, A.%
\end{APACrefauthors}%
\unskip\
\newblock
\APACrefYearMonthDay{2014}{}{}.
\newblock
{\BBOQ}\APACrefatitle {{The Algorithmic Foundations of Differential Privacy}} {{The Algorithmic Foundations of Differential Privacy}}.{\BBCQ}
\newblock
\APACjournalVolNumPages{Foundations and Trends in Theoretical Computer Science}{9}{3–4}{211–407}.
\PrintBackRefs{\CurrentBib}

\bibitem [\protect \citeauthoryear {%
El~Emam%
}{%
El~Emam%
}{%
{\protect \APACyear {2020}}%
}]{%
el2020seven}
\APACinsertmetastar {%
el2020seven}%
\begin{APACrefauthors}%
El~Emam, K.%
\end{APACrefauthors}%
\unskip\
\newblock
\APACrefYearMonthDay{2020}{}{}.
\newblock
{\BBOQ}\APACrefatitle {Seven ways to evaluate the utility of synthetic data} {Seven ways to evaluate the utility of synthetic data}.{\BBCQ}
\newblock
\APACjournalVolNumPages{IEEE Security \& Privacy}{18}{4}{56--59}.
\PrintBackRefs{\CurrentBib}

\bibitem [\protect \citeauthoryear {%
El~Emam%
, Mosquera%
, Fang%
\BCBL {}\ \BBA {} El-Hussuna%
}{%
El~Emam%
\ \protect \BOthers {.}}{%
{\protect \APACyear {2024}}%
}]{%
el2024evaluation}
\APACinsertmetastar {%
el2024evaluation}%
\begin{APACrefauthors}%
El~Emam, K.%
, Mosquera, L.%
, Fang, X.%
\BCBL {}\ \BBA {} El-Hussuna, A.%
\end{APACrefauthors}%
\unskip\
\newblock
\APACrefYearMonthDay{2024}{}{}.
\newblock
{\BBOQ}\APACrefatitle {An evaluation of the replicability of analyses using synthetic health data} {An evaluation of the replicability of analyses using synthetic health data}.{\BBCQ}
\newblock
\APACjournalVolNumPages{Scientific Reports}{14}{1}{6978}.
\PrintBackRefs{\CurrentBib}

\bibitem [\protect \citeauthoryear {%
Endres%
, Mannarapotta~Venugopal%
\BCBL {}\ \BBA {} Tran%
}{%
Endres%
\ \protect \BOthers {.}}{%
{\protect \APACyear {2022}}%
}]{%
endres2022synthetic}
\APACinsertmetastar {%
endres2022synthetic}%
\begin{APACrefauthors}%
Endres, M.%
, Mannarapotta~Venugopal, A.%
\BCBL {}\ \BBA {} Tran, T\BPBI S.%
\end{APACrefauthors}%
\unskip\
\newblock
\APACrefYearMonthDay{2022}{}{}.
\newblock
{\BBOQ}\APACrefatitle {Synthetic data generation: a comparative study} {Synthetic data generation: a comparative study}.{\BBCQ}
\newblock
\BIn{} \APACrefbtitle {Proceedings of the 26th International Database Engineered Applications Symposium} {Proceedings of the 26th international database engineered applications symposium}\ (\BPGS\ 94--102).
\PrintBackRefs{\CurrentBib}

\bibitem [\protect \citeauthoryear {%
Espinosa%
\ \BBA {} Figueira%
}{%
Espinosa%
\ \BBA {} Figueira%
}{%
{\protect \APACyear {2023}}%
}]{%
espinosa_2023}
\APACinsertmetastar {%
espinosa_2023}%
\begin{APACrefauthors}%
Espinosa, E.%
\BCBT {}\ \BBA {} Figueira, A.%
\end{APACrefauthors}%
\unskip\
\newblock
\APACrefYearMonthDay{2023}{}{}.
\newblock
{\BBOQ}\APACrefatitle {{On the Quality of Synthetic Generated Tabular Data}} {{On the Quality of Synthetic Generated Tabular Data}}.{\BBCQ}
\newblock
\APACjournalVolNumPages{Mathematics}{11}{15}{}.
\PrintBackRefs{\CurrentBib}

\bibitem [\protect \citeauthoryear {%
Figueira%
\ \BBA {} Vaz%
}{%
Figueira%
\ \BBA {} Vaz%
}{%
{\protect \APACyear {2022}}%
}]{%
figueira2022survey}
\APACinsertmetastar {%
figueira2022survey}%
\begin{APACrefauthors}%
Figueira, A.%
\BCBT {}\ \BBA {} Vaz, B.%
\end{APACrefauthors}%
\unskip\
\newblock
\APACrefYearMonthDay{2022}{}{}.
\newblock
{\BBOQ}\APACrefatitle {{Survey on Synthetic Data Generation, Evaluation Methods and GANs}} {{Survey on Synthetic Data Generation, Evaluation Methods and GANs}}.{\BBCQ}
\newblock
\APACjournalVolNumPages{Mathematics}{10}{15}{}.
\PrintBackRefs{\CurrentBib}

\bibitem [\protect \citeauthoryear {%
Ghalebikesabi%
\ \protect \BOthers {.}}{%
Ghalebikesabi%
\ \protect \BOthers {.}}{%
{\protect \APACyear {2022}}%
}]{%
ghalebikesabi2022mitigating}
\APACinsertmetastar {%
ghalebikesabi2022mitigating}%
\begin{APACrefauthors}%
Ghalebikesabi, S.%
, Wilde, H.%
, Jewson, J.%
, Doucet, A.%
, Vollmer, S.%
\BCBL {}\ \BBA {} Holmes, C.%
\end{APACrefauthors}%
\unskip\
\newblock
\APACrefYearMonthDay{2022}{}{}.
\newblock
\APACrefbtitle {Mitigating Statistical Bias within Differentially Private Synthetic Data.} {Mitigating statistical bias within differentially private synthetic data.}
\PrintBackRefs{\CurrentBib}

\bibitem [\protect \citeauthoryear {%
Ghosheh%
, Li%
\BCBL {}\ \BBA {} Zhu%
}{%
Ghosheh%
\ \protect \BOthers {.}}{%
{\protect \APACyear {2022}}%
}]{%
ghosheh2022review}
\APACinsertmetastar {%
ghosheh2022review}%
\begin{APACrefauthors}%
Ghosheh, G.%
, Li, J.%
\BCBL {}\ \BBA {} Zhu, T.%
\end{APACrefauthors}%
\unskip\
\newblock
\APACrefYearMonthDay{2022}{}{}.
\newblock
{\BBOQ}\APACrefatitle {{A review of Generative Adversarial Networks for Electronic Health Records: applications, evaluation measures and data sources}} {{A review of Generative Adversarial Networks for Electronic Health Records: applications, evaluation measures and data sources}}.{\BBCQ}
\newblock
\APACjournalVolNumPages{arXiv preprint}{}{}{}.
\PrintBackRefs{\CurrentBib}

\bibitem [\protect \citeauthoryear {%
Goodfellow%
\ \protect \BOthers {.}}{%
Goodfellow%
\ \protect \BOthers {.}}{%
{\protect \APACyear {2014}}%
}]{%
goodfellow2014generative}
\APACinsertmetastar {%
goodfellow2014generative}%
\begin{APACrefauthors}%
Goodfellow, I.%
, Pouget-Abadie, J.%
, Mirza, M.%
, Xu, B.%
, Warde-Farley, D.%
, Ozair, S.%
\BDBL {}Bengio, Y.%
\end{APACrefauthors}%
\unskip\
\newblock
\APACrefYearMonthDay{2014}{}{}.
\newblock
{\BBOQ}\APACrefatitle {Generative adversarial nets} {Generative adversarial nets}.{\BBCQ}
\newblock
\APACjournalVolNumPages{Advances in neural information processing systems}{27}{}{}.
\PrintBackRefs{\CurrentBib}

\bibitem [\protect \citeauthoryear {%
Hernandez%
, Epelde%
, Alberdi%
, Cilla%
\BCBL {}\ \BBA {} Rankin%
}{%
Hernandez%
\ \protect \BOthers {.}}{%
{\protect \APACyear {2022}}%
}]{%
hernandez2022synthetic}
\APACinsertmetastar {%
hernandez2022synthetic}%
\begin{APACrefauthors}%
Hernandez, M.%
, Epelde, G.%
, Alberdi, A.%
, Cilla, R.%
\BCBL {}\ \BBA {} Rankin, D.%
\end{APACrefauthors}%
\unskip\
\newblock
\APACrefYearMonthDay{2022}{}{}.
\newblock
{\BBOQ}\APACrefatitle {{Synthetic data generation for tabular health records: A systematic review}} {{Synthetic data generation for tabular health records: A systematic review}}.{\BBCQ}
\newblock
\APACjournalVolNumPages{Neurocomputing}{493}{}{28--45}.
\PrintBackRefs{\CurrentBib}

\bibitem [\protect \citeauthoryear {%
Hines%
, Dukes%
, Diaz-Ordaz%
\BCBL {}\ \BBA {} Vansteelandt%
}{%
Hines%
\ \protect \BOthers {.}}{%
{\protect \APACyear {2022}}%
}]{%
demystifying2022hines}
\APACinsertmetastar {%
demystifying2022hines}%
\begin{APACrefauthors}%
Hines, O.%
, Dukes, O.%
, Diaz-Ordaz, K.%
\BCBL {}\ \BBA {} Vansteelandt, S.%
\end{APACrefauthors}%
\unskip\
\newblock
\APACrefYearMonthDay{2022}{}{}.
\newblock
{\BBOQ}\APACrefatitle {Demystifying statistical learning based on efficient influence functions} {Demystifying statistical learning based on efficient influence functions}.{\BBCQ}
\newblock
\APACjournalVolNumPages{AMERICAN STATISTICIAN}{76}{3}{292--304}.
\PrintBackRefs{\CurrentBib}

\bibitem [\protect \citeauthoryear {%
Jordon%
, Yoon%
\BCBL {}\ \BBA {} Van Der~Schaar%
}{%
Jordon%
\ \protect \BOthers {.}}{%
{\protect \APACyear {2018}}%
}]{%
jordon2018pate}
\APACinsertmetastar {%
jordon2018pate}%
\begin{APACrefauthors}%
Jordon, J.%
, Yoon, J.%
\BCBL {}\ \BBA {} Van Der~Schaar, M.%
\end{APACrefauthors}%
\unskip\
\newblock
\APACrefYearMonthDay{2018}{}{}.
\newblock
{\BBOQ}\APACrefatitle {PATE-GAN: Generating synthetic data with differential privacy guarantees} {Pate-gan: Generating synthetic data with differential privacy guarantees}.{\BBCQ}
\newblock
\BIn{} \APACrefbtitle {International conference on learning representations.} {International conference on learning representations.}
\PrintBackRefs{\CurrentBib}

\bibitem [\protect \citeauthoryear {%
Kaloskampis%
, Joshi%
, Cheung%
\BCBL {}\ \BBA {} Nolan%
}{%
Kaloskampis%
\ \protect \BOthers {.}}{%
{\protect \APACyear {2020}}%
}]{%
kaloskampis2020synthetic}
\APACinsertmetastar {%
kaloskampis2020synthetic}%
\begin{APACrefauthors}%
Kaloskampis, I.%
, Joshi, C.%
, Cheung, C.%
\BCBL {}\ \BBA {} Nolan, L.%
\end{APACrefauthors}%
\unskip\
\newblock
\APACrefYearMonthDay{2020}{}{}.
\newblock
{\BBOQ}\APACrefatitle {Synthetic data in the civil service} {Synthetic data in the civil service}.{\BBCQ}
\newblock
\APACjournalVolNumPages{Significance}{17}{6}{18--23}.
\PrintBackRefs{\CurrentBib}

\bibitem [\protect \citeauthoryear {%
Kingma%
\ \BBA {} Welling%
}{%
Kingma%
\ \BBA {} Welling%
}{%
{\protect \APACyear {2013}}%
}]{%
kingma2013auto}
\APACinsertmetastar {%
kingma2013auto}%
\begin{APACrefauthors}%
Kingma, D\BPBI P.%
\BCBT {}\ \BBA {} Welling, M.%
\end{APACrefauthors}%
\unskip\
\newblock
\APACrefYearMonthDay{2013}{}{}.
\newblock
{\BBOQ}\APACrefatitle {Auto-encoding variational bayes} {Auto-encoding variational bayes}.{\BBCQ}
\newblock
\APACjournalVolNumPages{arXiv preprint}{}{}{}.
\PrintBackRefs{\CurrentBib}

\bibitem [\protect \citeauthoryear {%
Klein%
\ \BBA {} Sinha%
}{%
Klein%
\ \BBA {} Sinha%
}{%
{\protect \APACyear {2015}}%
}]{%
klein2015likelihood}
\APACinsertmetastar {%
klein2015likelihood}%
\begin{APACrefauthors}%
Klein, M.%
\BCBT {}\ \BBA {} Sinha, B.%
\end{APACrefauthors}%
\unskip\
\newblock
\APACrefYearMonthDay{2015}{}{}.
\newblock
{\BBOQ}\APACrefatitle {Likelihood based finite sample inference for singly imputed synthetic data under the multivariate normal and multiple linear regression models} {Likelihood based finite sample inference for singly imputed synthetic data under the multivariate normal and multiple linear regression models}.{\BBCQ}
\newblock
\APACjournalVolNumPages{Journal of Privacy and Confidentiality}{7}{1}{}.
\PrintBackRefs{\CurrentBib}

\bibitem [\protect \citeauthoryear {%
Lehmann%
\ \BBA {} Casella%
}{%
Lehmann%
\ \BBA {} Casella%
}{%
{\protect \APACyear {2006}}%
}]{%
lehmann2006theory}
\APACinsertmetastar {%
lehmann2006theory}%
\begin{APACrefauthors}%
Lehmann, E\BPBI L.%
\BCBT {}\ \BBA {} Casella, G.%
\end{APACrefauthors}%
\unskip\
\newblock
\APACrefYear{2006}.
\newblock
\APACrefbtitle {Theory of point estimation} {Theory of point estimation}.
\newblock
\APACaddressPublisher{}{Springer Science \& Business Media}.
\PrintBackRefs{\CurrentBib}

\bibitem [\protect \citeauthoryear {%
Liu%
, Qian%
, Berrevoets%
\BCBL {}\ \BBA {} van~der Schaar%
}{%
Liu%
\ \protect \BOthers {.}}{%
{\protect \APACyear {2022}}%
}]{%
liu2022goggle}
\APACinsertmetastar {%
liu2022goggle}%
\begin{APACrefauthors}%
Liu, T.%
, Qian, Z.%
, Berrevoets, J.%
\BCBL {}\ \BBA {} van~der Schaar, M.%
\end{APACrefauthors}%
\unskip\
\newblock
\APACrefYearMonthDay{2022}{}{}.
\newblock
{\BBOQ}\APACrefatitle {{GOGGLE}: Generative modelling for tabular data by learning relational structure} {{GOGGLE}: Generative modelling for tabular data by learning relational structure}.{\BBCQ}
\newblock
\BIn{} \APACrefbtitle {{The Eleventh International Conference on Learning Representations}.} {{The Eleventh International Conference on Learning Representations}.}
\PrintBackRefs{\CurrentBib}

\bibitem [\protect \citeauthoryear {%
Nowok%
, Raab%
\BCBL {}\ \BBA {} Dibben%
}{%
Nowok%
\ \protect \BOthers {.}}{%
{\protect \APACyear {2016}}%
}]{%
synthpop2016}
\APACinsertmetastar {%
synthpop2016}%
\begin{APACrefauthors}%
Nowok, B.%
, Raab, G\BPBI M.%
\BCBL {}\ \BBA {} Dibben, C.%
\end{APACrefauthors}%
\unskip\
\newblock
\APACrefYearMonthDay{2016}{}{}.
\newblock
{\BBOQ}\APACrefatitle {{Synthpop}: Bespoke Creation of Synthetic Data in {R}} {{Synthpop}: Bespoke creation of synthetic data in {R}}.{\BBCQ}
\newblock
\APACjournalVolNumPages{Journal of Statistical Software}{74}{11}{1--26}.
\PrintBackRefs{\CurrentBib}

\bibitem [\protect \citeauthoryear {%
Ohm%
}{%
Ohm%
}{%
{\protect \APACyear {2009}}%
}]{%
ohm2009broken}
\APACinsertmetastar {%
ohm2009broken}%
\begin{APACrefauthors}%
Ohm, P.%
\end{APACrefauthors}%
\unskip\
\newblock
\APACrefYearMonthDay{2009}{}{}.
\newblock
{\BBOQ}\APACrefatitle {Broken promises of privacy: Responding to the surprising failure of anonymization} {Broken promises of privacy: Responding to the surprising failure of anonymization}.{\BBCQ}
\newblock
\APACjournalVolNumPages{UCLA l. Rev.}{57}{}{1701}.
\PrintBackRefs{\CurrentBib}

\bibitem [\protect \citeauthoryear {%
Patki%
, Wedge%
\BCBL {}\ \BBA {} Veeramachaneni%
}{%
Patki%
\ \protect \BOthers {.}}{%
{\protect \APACyear {2016}}%
}]{%
SDV}
\APACinsertmetastar {%
SDV}%
\begin{APACrefauthors}%
Patki, N.%
, Wedge, R.%
\BCBL {}\ \BBA {} Veeramachaneni, K.%
\end{APACrefauthors}%
\unskip\
\newblock
\APACrefYearMonthDay{2016}{}{}.
\newblock
{\BBOQ}\APACrefatitle {The Synthetic data vault} {The synthetic data vault}.{\BBCQ}
\newblock
\BIn{} \APACrefbtitle {{IEEE International Conference on Data Science and Advanced Analytics (DSAA)}} {{IEEE International Conference on Data Science and Advanced Analytics (DSAA)}}\ (\BPG~399-410).
\PrintBackRefs{\CurrentBib}

\bibitem [\protect \citeauthoryear {%
Pearl%
}{%
Pearl%
}{%
{\protect \APACyear {2011}}%
}]{%
pearl2011BN}
\APACinsertmetastar {%
pearl2011BN}%
\begin{APACrefauthors}%
Pearl, J.%
\end{APACrefauthors}%
\unskip\
\newblock
\APACrefYear{2011}.
\newblock
\APACrefbtitle {Bayesian Networks} {Bayesian networks}.
\PrintBackRefs{\CurrentBib}

\bibitem [\protect \citeauthoryear {%
Qian%
, Cebere%
\BCBL {}\ \BBA {} van~der Schaar%
}{%
Qian%
\ \protect \BOthers {.}}{%
{\protect \APACyear {2023}}%
}]{%
Synthcity}
\APACinsertmetastar {%
Synthcity}%
\begin{APACrefauthors}%
Qian, Z.%
, Cebere, B\BHBI C.%
\BCBL {}\ \BBA {} van~der Schaar, M.%
\end{APACrefauthors}%
\unskip\
\newblock
\APACrefYearMonthDay{2023}{}{}.
\newblock
{\BBOQ}\APACrefatitle {Synthcity: facilitating innovative use cases of synthetic data in different data modalities} {Synthcity: facilitating innovative use cases of synthetic data in different data modalities}.{\BBCQ}
\newblock
\APACjournalVolNumPages{arXiv preprint}{}{}{}.
\PrintBackRefs{\CurrentBib}

\bibitem [\protect \citeauthoryear {%
Raab%
, Nowok%
\BCBL {}\ \BBA {} Dibben%
}{%
Raab%
\ \protect \BOthers {.}}{%
{\protect \APACyear {2016}}%
}]{%
raab2016practical}
\APACinsertmetastar {%
raab2016practical}%
\begin{APACrefauthors}%
Raab, G\BPBI M.%
, Nowok, B.%
\BCBL {}\ \BBA {} Dibben, C.%
\end{APACrefauthors}%
\unskip\
\newblock
\APACrefYearMonthDay{2016}{}{}.
\newblock
{\BBOQ}\APACrefatitle {Practical data synthesis for large samples} {Practical data synthesis for large samples}.{\BBCQ}
\newblock
\APACjournalVolNumPages{Journal of Privacy and Confidentiality}{7}{3}{67--97}.
\PrintBackRefs{\CurrentBib}

\bibitem [\protect \citeauthoryear {%
Raghunathan%
}{%
Raghunathan%
}{%
{\protect \APACyear {2021}}%
}]{%
raghunathan2021synthetic}
\APACinsertmetastar {%
raghunathan2021synthetic}%
\begin{APACrefauthors}%
Raghunathan, T\BPBI E.%
\end{APACrefauthors}%
\unskip\
\newblock
\APACrefYearMonthDay{2021}{}{}.
\newblock
{\BBOQ}\APACrefatitle {Synthetic data} {Synthetic data}.{\BBCQ}
\newblock
\APACjournalVolNumPages{Annual review of statistics and its application}{8}{}{129--140}.
\PrintBackRefs{\CurrentBib}

\bibitem [\protect \citeauthoryear {%
Raghunathan%
, Reiter%
\BCBL {}\ \BBA {} Rubin%
}{%
Raghunathan%
\ \protect \BOthers {.}}{%
{\protect \APACyear {2003}}%
}]{%
raghunathan2003multiple}
\APACinsertmetastar {%
raghunathan2003multiple}%
\begin{APACrefauthors}%
Raghunathan, T\BPBI E.%
, Reiter, J\BPBI P.%
\BCBL {}\ \BBA {} Rubin, D\BPBI B.%
\end{APACrefauthors}%
\unskip\
\newblock
\APACrefYearMonthDay{2003}{}{}.
\newblock
{\BBOQ}\APACrefatitle {Multiple imputation for statistical disclosure limitation} {Multiple imputation for statistical disclosure limitation}.{\BBCQ}
\newblock
\APACjournalVolNumPages{Journal of official statistics}{19}{1}{1}.
\PrintBackRefs{\CurrentBib}

\bibitem [\protect \citeauthoryear {%
R{\"a}is{\"a}%
, J{\"a}lk{\"o}%
, Kaski%
\BCBL {}\ \BBA {} Honkela%
}{%
R{\"a}is{\"a}%
\ \protect \BOthers {.}}{%
{\protect \APACyear {2023}}%
}]{%
raisa2023noise}
\APACinsertmetastar {%
raisa2023noise}%
\begin{APACrefauthors}%
R{\"a}is{\"a}, O.%
, J{\"a}lk{\"o}, J.%
, Kaski, S.%
\BCBL {}\ \BBA {} Honkela, A.%
\end{APACrefauthors}%
\unskip\
\newblock
\APACrefYearMonthDay{2023}{}{}.
\newblock
{\BBOQ}\APACrefatitle {Noise-Aware Statistical Inference with Differentially Private Synthetic Data} {Noise-aware statistical inference with differentially private synthetic data}.{\BBCQ}
\newblock
\BIn{} \APACrefbtitle {International Conference on Artificial Intelligence and Statistics} {International conference on artificial intelligence and statistics}\ (\BPGS\ 3620--3643).
\PrintBackRefs{\CurrentBib}

\bibitem [\protect \citeauthoryear {%
Rajabi%
\ \BBA {} Garibay%
}{%
Rajabi%
\ \BBA {} Garibay%
}{%
{\protect \APACyear {2022}}%
}]{%
rajabi2022tabfairgan}
\APACinsertmetastar {%
rajabi2022tabfairgan}%
\begin{APACrefauthors}%
Rajabi, A.%
\BCBT {}\ \BBA {} Garibay, O\BPBI O.%
\end{APACrefauthors}%
\unskip\
\newblock
\APACrefYearMonthDay{2022}{}{}.
\newblock
{\BBOQ}\APACrefatitle {{{TabFairGAN}: Fair Tabular Data Generation with Generative Adversarial Networks}} {{{TabFairGAN}: Fair Tabular Data Generation with Generative Adversarial Networks}}.{\BBCQ}
\newblock
\APACjournalVolNumPages{Machine Learning and Knowledge Extraction}{4}{2}{488--501}.
\PrintBackRefs{\CurrentBib}

\bibitem [\protect \citeauthoryear {%
Reiter%
}{%
Reiter%
}{%
{\protect \APACyear {2005}}%
}]{%
reiter2005releasing}
\APACinsertmetastar {%
reiter2005releasing}%
\begin{APACrefauthors}%
Reiter, J\BPBI P.%
\end{APACrefauthors}%
\unskip\
\newblock
\APACrefYearMonthDay{2005}{}{}.
\newblock
{\BBOQ}\APACrefatitle {Releasing multiply imputed, synthetic public use microdata: an illustration and empirical study} {Releasing multiply imputed, synthetic public use microdata: an illustration and empirical study}.{\BBCQ}
\newblock
\APACjournalVolNumPages{Journal of the Royal Statistical Society Series A: Statistics in Society}{168}{1}{185--205}.
\PrintBackRefs{\CurrentBib}

\bibitem [\protect \citeauthoryear {%
Reiter%
\ \BBA {} Mitra%
}{%
Reiter%
\ \BBA {} Mitra%
}{%
{\protect \APACyear {2009}}%
}]{%
reiter2009estimating}
\APACinsertmetastar {%
reiter2009estimating}%
\begin{APACrefauthors}%
Reiter, J\BPBI P.%
\BCBT {}\ \BBA {} Mitra, R.%
\end{APACrefauthors}%
\unskip\
\newblock
\APACrefYearMonthDay{2009}{}{}.
\newblock
{\BBOQ}\APACrefatitle {Estimating risks of identification disclosure in partially synthetic data} {Estimating risks of identification disclosure in partially synthetic data}.{\BBCQ}
\newblock
\APACjournalVolNumPages{Journal of Privacy and Confidentiality}{1}{1}{}.
\PrintBackRefs{\CurrentBib}

\bibitem [\protect \citeauthoryear {%
Rubin%
}{%
Rubin%
}{%
{\protect \APACyear {1993}}%
}]{%
rubin1993statistical}
\APACinsertmetastar {%
rubin1993statistical}%
\begin{APACrefauthors}%
Rubin, D\BPBI B.%
\end{APACrefauthors}%
\unskip\
\newblock
\APACrefYearMonthDay{1993}{}{}.
\newblock
{\BBOQ}\APACrefatitle {Statistical disclosure limitation} {Statistical disclosure limitation}.{\BBCQ}
\newblock
\APACjournalVolNumPages{Journal of official Statistics}{9}{2}{461--468}.
\PrintBackRefs{\CurrentBib}

\bibitem [\protect \citeauthoryear {%
Stadler%
, Oprisanu%
\BCBL {}\ \BBA {} Troncoso%
}{%
Stadler%
\ \protect \BOthers {.}}{%
{\protect \APACyear {2022}}%
}]{%
stadler2022synthetic}
\APACinsertmetastar {%
stadler2022synthetic}%
\begin{APACrefauthors}%
Stadler, T.%
, Oprisanu, B.%
\BCBL {}\ \BBA {} Troncoso, C.%
\end{APACrefauthors}%
\unskip\
\newblock
\APACrefYearMonthDay{2022}{}{}.
\newblock
{\BBOQ}\APACrefatitle {Synthetic data--anonymisation groundhog day} {Synthetic data--anonymisation groundhog day}.{\BBCQ}
\newblock
\BIn{} \APACrefbtitle {31st USENIX Security Symposium (USENIX Security 22)} {31st usenix security symposium (usenix security 22)}\ (\BPGS\ 1451--1468).
\PrintBackRefs{\CurrentBib}

\bibitem [\protect \citeauthoryear {%
Tao%
, McKenna%
, Hay%
, Machanavajjhala%
\BCBL {}\ \BBA {} Miklau%
}{%
Tao%
\ \protect \BOthers {.}}{%
{\protect \APACyear {2021}}%
}]{%
tao2021benchmarking}
\APACinsertmetastar {%
tao2021benchmarking}%
\begin{APACrefauthors}%
Tao, Y.%
, McKenna, R.%
, Hay, M.%
, Machanavajjhala, A.%
\BCBL {}\ \BBA {} Miklau, G.%
\end{APACrefauthors}%
\unskip\
\newblock
\APACrefYearMonthDay{2021}{}{}.
\newblock
{\BBOQ}\APACrefatitle {{Benchmarking Differentially Private Synthetic Data Generation Algorithms}} {{Benchmarking Differentially Private Synthetic Data Generation Algorithms}}.{\BBCQ}
\newblock
\APACjournalVolNumPages{arXiv preprint}{}{}{}.
\PrintBackRefs{\CurrentBib}

\bibitem [\protect \citeauthoryear {%
van Breugel%
, Qian%
\BCBL {}\ \BBA {} van~der Schaar%
}{%
van Breugel%
\ \protect \BOthers {.}}{%
{\protect \APACyear {2023}}%
}]{%
van2023synthetic}
\APACinsertmetastar {%
van2023synthetic}%
\begin{APACrefauthors}%
van Breugel, B.%
, Qian, Z.%
\BCBL {}\ \BBA {} van~der Schaar, M.%
\end{APACrefauthors}%
\unskip\
\newblock
\APACrefYearMonthDay{2023}{}{}.
\newblock
{\BBOQ}\APACrefatitle {{Synthetic data, real errors: How (not) to publish and use synthetic data}} {{Synthetic data, real errors: How (not) to publish and use synthetic data}}.{\BBCQ}
\newblock
\APACjournalVolNumPages{arXiv preprint}{}{}{}.
\PrintBackRefs{\CurrentBib}

\bibitem [\protect \citeauthoryear {%
van~der Laan%
\ \BBA {} Rose%
}{%
van~der Laan%
\ \BBA {} Rose%
}{%
{\protect \APACyear {2011}}%
}]{%
van_der_laan_targeted_2011}
\APACinsertmetastar {%
van_der_laan_targeted_2011}%
\begin{APACrefauthors}%
van~der Laan, M\BPBI J.%
\BCBT {}\ \BBA {} Rose, S.%
\end{APACrefauthors}%
\unskip\
\newblock
\APACrefYear{2011}.
\newblock
\APACrefbtitle {Targeted {Learning}} {Targeted {Learning}}.
\newblock
\APACaddressPublisher{New York, NY}{Springer New York}.
\newblock
\begin{APACrefDOI} \doi{10.1007/978-1-4419-9782-1} \end{APACrefDOI}
\PrintBackRefs{\CurrentBib}

\bibitem [\protect \citeauthoryear {%
Van~der Laan%
\ \BBA {} Rose%
}{%
Van~der Laan%
\ \BBA {} Rose%
}{%
{\protect \APACyear {2011}}%
}]{%
van2011targeted}
\APACinsertmetastar {%
van2011targeted}%
\begin{APACrefauthors}%
Van~der Laan, M\BPBI J.%
\BCBT {}\ \BBA {} Rose, S.%
\end{APACrefauthors}%
\unskip\
\newblock
\APACrefYear{2011}.
\newblock
\APACrefbtitle {Targeted learning: causal inference for observational and experimental data} {Targeted learning: causal inference for observational and experimental data}\ (\BVOL~4).
\newblock
\APACaddressPublisher{}{Springer}.
\PrintBackRefs{\CurrentBib}

\bibitem [\protect \citeauthoryear {%
Vansteelandt%
\ \BBA {} Dukes%
}{%
Vansteelandt%
\ \BBA {} Dukes%
}{%
{\protect \APACyear {2022}}%
}]{%
vansteelandt2022assumption}
\APACinsertmetastar {%
vansteelandt2022assumption}%
\begin{APACrefauthors}%
Vansteelandt, S.%
\BCBT {}\ \BBA {} Dukes, O.%
\end{APACrefauthors}%
\unskip\
\newblock
\APACrefYearMonthDay{2022}{}{}.
\newblock
{\BBOQ}\APACrefatitle {Assumption-lean inference for generalised linear model parameters} {Assumption-lean inference for generalised linear model parameters}.{\BBCQ}
\newblock
\APACjournalVolNumPages{Journal of the Royal Statistical Society Series B: Statistical Methodology}{84}{3}{657--685}.
\PrintBackRefs{\CurrentBib}

\bibitem [\protect \citeauthoryear {%
Wan%
, Zhang%
\BCBL {}\ \BBA {} He%
}{%
Wan%
\ \protect \BOthers {.}}{%
{\protect \APACyear {2017}}%
}]{%
wan2017variational}
\APACinsertmetastar {%
wan2017variational}%
\begin{APACrefauthors}%
Wan, Z.%
, Zhang, Y.%
\BCBL {}\ \BBA {} He, H.%
\end{APACrefauthors}%
\unskip\
\newblock
\APACrefYearMonthDay{2017}{}{}.
\newblock
{\BBOQ}\APACrefatitle {Variational autoencoder based synthetic data generation for imbalanced learning} {Variational autoencoder based synthetic data generation for imbalanced learning}.{\BBCQ}
\newblock
\BIn{} \APACrefbtitle {{2017 IEEE symposium series on computational intelligence (SSCI)}} {{2017 IEEE symposium series on computational intelligence (SSCI)}}\ (\BPGS\ 1--7).
\PrintBackRefs{\CurrentBib}

\bibitem [\protect \citeauthoryear {%
Wilde%
, Jewson%
, Vollmer%
\BCBL {}\ \BBA {} Holmes%
}{%
Wilde%
\ \protect \BOthers {.}}{%
{\protect \APACyear {2021}}%
}]{%
wilde2021foundations}
\APACinsertmetastar {%
wilde2021foundations}%
\begin{APACrefauthors}%
Wilde, H.%
, Jewson, J.%
, Vollmer, S.%
\BCBL {}\ \BBA {} Holmes, C.%
\end{APACrefauthors}%
\unskip\
\newblock
\APACrefYearMonthDay{2021}{}{}.
\newblock
{\BBOQ}\APACrefatitle {Foundations of Bayesian learning from synthetic data} {Foundations of bayesian learning from synthetic data}.{\BBCQ}
\newblock
\BIn{} \APACrefbtitle {International Conference on Artificial Intelligence and Statistics} {International conference on artificial intelligence and statistics}\ (\BPGS\ 541--549).
\PrintBackRefs{\CurrentBib}

\bibitem [\protect \citeauthoryear {%
Xie%
, Lin%
, Wang%
, Wang%
\BCBL {}\ \BBA {} Zhou%
}{%
Xie%
\ \protect \BOthers {.}}{%
{\protect \APACyear {2018}}%
}]{%
xie2018differentially}
\APACinsertmetastar {%
xie2018differentially}%
\begin{APACrefauthors}%
Xie, L.%
, Lin, K.%
, Wang, S.%
, Wang, F.%
\BCBL {}\ \BBA {} Zhou, J.%
\end{APACrefauthors}%
\unskip\
\newblock
\APACrefYearMonthDay{2018}{}{}.
\newblock
{\BBOQ}\APACrefatitle {Differentially private generative adversarial network} {Differentially private generative adversarial network}.{\BBCQ}
\newblock
\APACjournalVolNumPages{arXiv preprint arXiv:1802.06739}{}{}{}.
\PrintBackRefs{\CurrentBib}

\bibitem [\protect \citeauthoryear {%
Xu%
, Skoularidou%
, Cuesta-Infante%
\BCBL {}\ \BBA {} Veeramachaneni%
}{%
Xu%
\ \protect \BOthers {.}}{%
{\protect \APACyear {2019}}%
}]{%
xu2019modeling}
\APACinsertmetastar {%
xu2019modeling}%
\begin{APACrefauthors}%
Xu, L.%
, Skoularidou, M.%
, Cuesta-Infante, A.%
\BCBL {}\ \BBA {} Veeramachaneni, K.%
\end{APACrefauthors}%
\unskip\
\newblock
\APACrefYearMonthDay{2019}{}{}.
\newblock
{\BBOQ}\APACrefatitle {Modeling tabular data using conditional {GAN}} {Modeling tabular data using conditional {GAN}}.{\BBCQ}
\newblock
\APACjournalVolNumPages{Advances in neural information processing systems}{32}{}{}.
\PrintBackRefs{\CurrentBib}

\bibitem [\protect \citeauthoryear {%
Yan%
\ \protect \BOthers {.}}{%
Yan%
\ \protect \BOthers {.}}{%
{\protect \APACyear {2022}}%
}]{%
yan2022multifaceted}
\APACinsertmetastar {%
yan2022multifaceted}%
\begin{APACrefauthors}%
Yan, C.%
, Yan, Y.%
, Wan, Z.%
, Zhang, Z.%
, Omberg, L.%
, Guinney, J.%
\BDBL {}Malin, B\BPBI A.%
\end{APACrefauthors}%
\unskip\
\newblock
\APACrefYearMonthDay{2022}{}{}.
\newblock
{\BBOQ}\APACrefatitle {A multifaceted benchmarking of synthetic electronic health record generation models} {A multifaceted benchmarking of synthetic electronic health record generation models}.{\BBCQ}
\newblock
\APACjournalVolNumPages{Nature communications}{13}{1}{7609}.
\PrintBackRefs{\CurrentBib}

\bibitem [\protect \citeauthoryear {%
J.~Zhang%
, Cormode%
, Procopiuc%
, Srivastava%
\BCBL {}\ \BBA {} Xiao%
}{%
J.~Zhang%
\ \protect \BOthers {.}}{%
{\protect \APACyear {2017}}%
}]{%
zhang2017privbayes}
\APACinsertmetastar {%
zhang2017privbayes}%
\begin{APACrefauthors}%
Zhang, J.%
, Cormode, G.%
, Procopiuc, C\BPBI M.%
, Srivastava, D.%
\BCBL {}\ \BBA {} Xiao, X.%
\end{APACrefauthors}%
\unskip\
\newblock
\APACrefYearMonthDay{2017}{}{}.
\newblock
{\BBOQ}\APACrefatitle {Privbayes: Private data release via bayesian networks} {Privbayes: Private data release via bayesian networks}.{\BBCQ}
\newblock
\APACjournalVolNumPages{ACM Transactions on Database Systems (TODS)}{42}{4}{1--41}.
\PrintBackRefs{\CurrentBib}

\bibitem [\protect \citeauthoryear {%
Z.~Zhang%
, Yan%
, Mesa%
, Sun%
\BCBL {}\ \BBA {} Malin%
}{%
Z.~Zhang%
\ \protect \BOthers {.}}{%
{\protect \APACyear {2020}}%
}]{%
zhang2020ensuring}
\APACinsertmetastar {%
zhang2020ensuring}%
\begin{APACrefauthors}%
Zhang, Z.%
, Yan, C.%
, Mesa, D\BPBI A.%
, Sun, J.%
\BCBL {}\ \BBA {} Malin, B\BPBI A.%
\end{APACrefauthors}%
\unskip\
\newblock
\APACrefYearMonthDay{2020}{}{}.
\newblock
{\BBOQ}\APACrefatitle {Ensuring electronic medical record simulation through better training, modeling, and evaluation} {Ensuring electronic medical record simulation through better training, modeling, and evaluation}.{\BBCQ}
\newblock
\APACjournalVolNumPages{Journal of the American Medical Informatics Association}{27}{1}{99--108}.
\PrintBackRefs{\CurrentBib}

\end{thebibliography}

% References
% \bibliography{uai2024-template}

\newpage

\onecolumn
\setcounter{section}{0}
\renewcommand{\thesection}{\arabic{section}}

\raggedbottom % wat doet dit?

\appendix
\title{Appendix}

\maketitle

\section{Derivation of Corrected Standard Error}
\label{sec:APPENDIX_derivation}

In this proof, we study the large sample behaviour of a $\sqrt{n}$-consistent estimator for a scalar parameter $\theta$ calculated on the synthetic data. Let $P$ refer to the true distribution of the data. Then $\theta$ can be viewed as a functional of $P$; we will denote it $\theta(P)$. For instance, the population mean of an outcome $Y$ can be written as $\theta(P)=\int ydP(y)$. The observed data forms a sample from the distribution $P$. They enable us to obtain an estimator of $P$, denoted $P_n$, so that estimators of $\theta$ can be viewed as the functional $\theta(.)$ evaluated at $P_n$: $\theta(P_n)$ \citep{brickel1993efficient}. For instance, with $P_n$ the empirical distribution of the data, which assigns point mass to each data point, we have that 
$\theta(P_n)=\int ydP_n(y)=n^{-1}\sum_{i=1}^n Y_i$. The fact that we consider $\sqrt{n}$-consistent estimators implies that $E\left\{\theta(P_n)\right\}=\theta(P)+o(n^{-1/2})$ and that $\mbox{\rm Var}\left\{\theta(P_n)\right\}=\sigma^2(P)/n+o(n^{-1})$ for some constant $\sigma^2(P)$ (which need not represent the outcome variance).

When using synthetic data, we first construct an estimator $\hat{P}$ of $P$, next sample $m$ independent data from $\hat{P}$ and finally obtain an estimator of $\hat{P}$, denoted $\hat{P}_m$, so that the estimator obtained from the synthetic data can be written as $\theta(\hat{P}_m)$. We then have that 
\begin{eqnarray*}
\mbox{\rm Var}\left\{\theta(\hat{P}_m)\right\}
&=&E\left[\mbox{\rm Var}\left\{\theta(\hat{P}_m)|\hat{P}\right\}\right]
+\mbox{\rm Var}\left[E\left\{\theta(\hat{P}_m)|\hat{P}\right\}\right]\\
&=&E\left[\sigma^2(\hat{P})/m+o(m^{-1})\right]
+\mbox{\rm Var}\left[\theta(\hat{P})+o(m^{-1/2})\right]\\
&=&\sigma^2(P)/m+o(n^{-1}m^{-1})+o(m^{-1})
+\sigma^2(P)/n+o(m^{-1})+o(n^{-1})\\
&=&\sigma^2(P)\left(\frac{1}{m}+\frac{1}{n}\right)+o(m^{-1})+o(n^{-1}).
\end{eqnarray*}
Here, we use that $\theta(.)$ and $\sigma^2(.)$ are smooth functionals of $P$ (in the sense of being path-wise differentiable parameters \citep{demystifying2022hines}) and $\hat{P}$ being a $\sqrt{n}$-consistent estimator of $P$ (which will generally be satisfied when a parametric synthetic data generation method is used, but not otherwise). We conclude that the standard error of the estimator for $\theta$ as obtained on synthetic data can be approximated in large samples as 
\[\sigma(P)\sqrt{\frac{1}{m}+\frac{1}{n}};\]
note that `large sample' here refers to both the original and synthetic data size being `large'.
The fact that a naive analysis will deliver a standard error equal to $\sigma(P)/\sqrt{m}$, explains the correction reported in the main text.

\newpage 

\section{Synthetic Data Generation Methods} \label{subsec:APPENDIX_sim_synthmethods}

\setcounter{figure}{0}
\renewcommand{\thefigure}{B\arabic{figure}}
\setcounter{table}{0}
\renewcommand{\thetable}{B\arabic{table}}

We elaborate on the generative models used to create synthetic data in our study. These are split into statistical approaches and deep learning (DL) approaches, following the categorisation suggested in \cite{hernandez2022synthetic}. All models were trained on our internal cluster using a single GPU (NVIDIA Ampere A100; only utilised by our DL methods) and eight CPUs (AMD EPYC 7413), taking less than $24$ hours to complete.

\subsection{Statistical approaches}
\label{subsubsec:APPENDIX_sim_synthmethods_Stat}

Our first statistical approach \codebold{Synthpop} uses the synthetic data generation framework built into the \codepackage{R} package \codepackage{Synthpop} \citep{synthpop2016}. Specifically, we rely on its default parametric method. In this approach, the user can provide the assumed dependencies between the variables in the form of a Directed Acyclic Graph (DAG), whose topological ordering prescribes the sequential order in which to synthesise the variables. In our simulation study, the true structure of the DAG as depicted in Figure \ref{DAG_simulated_data} is provided. In our case study, we work with the Adult Census Income Dataset, for which the true dependency structure between the variables is unknown. In this case, we do not pass a DAG to the \codepackage{Synthpop} method, which then falls back on using the arbitrary ordering of the columns in the dataset ($age$ first, $income$ last) as a variable ordering. 

Once the sequential ordering is fixed, each variable is modelled by fitting a parametric or non-parametric representation based on the original data $R$, conditionally on its parent variables (except for the root nodes, which come first in the sequence and do not have any parents, and are generated using bootstrap samples). The type of model used to fit each representation is based on the data type of the considered variable. By default, \codepackage{Synthpop} uses distribution-preserving linear regression, logistic regression, unordered polytomous regression, and ordered polytomous regression models for continuous, binary, unordered categorical, and ordered categorical variables, respectively. These synthesising methods were used in our simulation study, whereas in the case study classification and regression trees (non-parametric) were used for the generation of unordered categorical variables. This is because the Adult Census Income dataset contains multiple variables with many categories that cannot be readily fitted by the default unordered polytomous regression model.

Parallel to the fitting of each conditional distribution, synthetic data are generated for that particular variable. The exact implementation of this procedure depends on the assumed model and is based on the synthesised values of all variables preceding it in the sequential ordering. In addition, the process of generating data can be based on either a `proper' or `simple' synthesis, referring to whether each method samples from the posterior distribution of the parameters of the conditional models or not, respectively. In both the simulation study and case study, we opted for simple synthesis. For a more detailed explanation of \codepackage{Synthpop}, we refer the interested reader to the package's source code and \cite{synthpop2016}.

Both the second and third statistical approach are based on creating synthetic data through Bayesian Networks (BNs). We implement both a method where the DAG is pre-specified by the user (\codebold{BN DAG}) and a method where the DAG is estimated by using an algorithm (\codebold{BN}). In the former case, we again provide the true DAG structure as depicted in Figure \ref{DAG_simulated_data} for our simulation study. In the latter case, the unknown DAG is estimated based on a tree search using the Chow-Liu algorithm \citep{chowliu} (note that this Bayesian approach is therefore not a `pure' statistical method). For both Bayesian Network implementations, the conditional probability distributions (CPDs) are estimated using Maximum Likelihood Estimation (MLE) and synthetic data are generated via forward sampling. More information on the algorithm and estimators can be found in the documentation of \codepackage{PGMPY} \citep{ankan2015pgmpy}.

Both \codebold{BN} and \codebold{BN DAG} are included in order to investigate whether the availability of the correct DAG would result in better performance (e.g. less variability in estimators) compared with a BN that does not have prior knowledge and needs to rely on data-adaptive DAG discovery.
In many practical settings, the (full) DAG cannot be provided upfront since causal relationships between the variables are unknown. In those cases, dependency structure discovery methods like the Chow-Liu algorithm are often used to recover the DAG. 
Note that the performance of a model that follows this paradigm %is included, even if its performance 
is upper bounded by the performance of the model that receives the DAG upfront.

There are multiple differences between \codebold{Synthpop} and the \codebold{BN} (with or without DAG) implementation. In a BN, a joint probability is obtained through factorisation. When all variables are discrete, natural estimates for the CPDs are the relative frequencies, which coincide with the MLE of a multinomial model. Within our BN implementation, continuous variables undergo discretisation. It is possible to avoid this, but practically this means imposing a linear Gaussian CPD for all variables, including the discrete ones, which undermines the representation power of the BN. In Synthpop, the joint distribution is also defined in terms of a series of conditional distributions. With its parametric methods, Synthpop imposes a specific distribution and parametric regression model depending on the variable type. Therefore, the likelihood can now be written as a function of these regression models, instead of just the multinomial likelihood function seen in BN, and the corresponding parameters of these regression models are estimated via MLE. Depending on the variable type, different parametric models are possible, as opposed to BNs, where the distribution is either multinomial for discrete variables or Gaussian when (non-discretised) continuous variables are included in the mix. Thus, the difference between \codebold{Synthpop} and \codebold{BN} (with or without DAG) lies in the flexibility of the assumed parametric distribution and the way each method deals with mixed variable types.

\subsection{Deep learning approaches}
\label{subsubsec:APPENDIX_sim_synthmethods_DL}

We focus on two commonly used deep generative models, namely Generative Adversarial Networks (GANs) \citep{goodfellow2014generative} and Variational Autoencoders (VAEs) \citep{kingma2013auto}. 

A GAN consists of two competing neural networks, a generator and discriminator, and aims to achieve an equilibrium between both \citep{hernandez2022synthetic}. This translates to a mini-max game, since the generator aims to minimise the difference between the real and generated data, while the discriminator aims to maximise the possibility to distinguish the real and generated data \citep{goodfellow2014generative}. We use the \codebold{CTGAN} implementation that was designed specifically for tabular data, proposed by \cite{xu2019modeling}. 

A VAE is a deep latent variable model, consisting of an encoder and a decoder \citep{kingma2013auto}. The encoder models the approximate posterior distribution of the latent variables given an input instance, whereby typically a standard normal prior is assumed for the latent variables. The decoder allows reconstructing an input instance, based on a sample from the predicted latent space distribution. Encoder and decoder can be jointly trained by maximising the \emph{Evidence Lower BOund} (ELBO), i.e. the marginal likelihood of the training instances. Maximising the ELBO corresponds to minimising the KL-divergence between the predicted latent variable distribution for a given input instance and the standard normal priors, and minimising the reconstruction error of the input instance at the decoder output. Once again, we use the tabular implementation of a VAE (\codebold{TVAE}) proposed by \cite{xu2019modeling}. 

As these DL approaches are very expressive models, especially compared to the low-dimensional data that were used in the simulation study, tuning these models is an important step towards prevention of overfitting. For this reason, the next section outlines the strategy followed to tune the hyperparameters of our \code{CTGAN} and \code{TVAE} models. To study the effect of hyperparameter tuning, we also considered untuned versions of both DL methods (\code{Default CTGAN} and \code{Default TVAE}). Here, we used the implementation from the \codepackage{Synthcity} library, with the hyperparameters set to their default values \citep{Synthcity}.

\subsection{Hyperparameter tuning}
\label{subsubsection:APPENDIX_sim_synthmethods_DL_hpo}

Since \codepackage{Synthcity}'s implementation did not allow us to tune all hyperparameters we wanted, we implemented an extended version of the CTGAN and TVAE modules, where we used \codepackage{Synthetic Data Vault}'s implementation as a baseline \citep{SDV}. This made it possible to tune additional regularisation hyperparameters: generator and discriminator dropout were added to the CTGAN module, and encoder and decoder dropout were added to the TVAE module. 
Note that \codepackage{SDV} implements the CTGAN and TVAE modules as originally proposed by \cite{xu2019modeling}, where a cluster-based normaliser is used to preprocess numerical features.

\paragraph{Hyperparameters CTGAN} The following hyperparameters were tuned: number of hidden layers of generator $\in \{1,2,3,4\}$, number of nodes per hidden layer of generator $\in \{8,16,32,64,128,256,512\}$, number of hidden layers of discriminator $\in \{1,2,3,4\}$, number of nodes per hidden layer of discriminator $\in \{8,16,32,64,128,256,512\}$, number of epochs log-uniformly $\in [5,300]$, number of iterations in the discriminator per iteration of generator $\in \{1,5,10\}$, learning rate (the same for generator and discriminator) log-uniformly $\in [1\mathrm{e}{-6},1\mathrm{e}{-2}]$, dropout in generator uniformly $\in [0,1]$, dropout in discriminator uniformly $\in [0,1]$, weight decay of generator log-uniformly $\in [1\mathrm{e}{-6},1]$, and weight decay of discriminator log-uniformly $\in [1\mathrm{e}{-6},1]$. Batch size was fixed at min$(200, n)$, with $n$ the sample size.

\paragraph{Hyperparameters TVAE} The following hyperparameters were tuned: embedding dimension $\in \{32,64,128,256,512\}$, number of hidden layers (the same for encoder and decoder) $\in \{1,2,3,4\}$, number of nodes per hidden layer (the same for encoder and decoder) $\in \{32,64,128,256,512\}$, number of epochs log-uniformly $\in [200, 1000]$, reconstruction error loss factor $\in \{1,2,5,10\}$, dropout in encoder uniformly $\in [0,1]$, dropout in decoder uniformly $\in [0,1]$, and weight decay (the same for encoder and decoder) log-uniformly $\in [1\mathrm{e}{-6},1]$. Batch size was fixed at min$(200, n)$, with $n$ the sample size.

\paragraph{Objective score}
The average inverse of the Kullback–Leibler divergence (IKLD) between the original and the synthetic dataset was used as metric. 5-fold cross-validation was used as follows: each time four training folds of original data were used to train the generative model and one validation fold of original data was used to calculate the IKLD with the synthetic data (of equal size as the validation fold) generated by the model. To make the IKLD metric independent of data dimension (lower IKLDs are typically seen for smaller sample sizes even if the synthetic datasets would be sampled directly from the ground truth population), the 5-fold cross-validated IKLD was normalised by the 5-fold cross-validated IKLD of a generative model that simply generates bootstrap samples of the original data. Finally, this procedure was repeated and averaged over five seeds to make the performance independent of seed initialisation (used for split in train and validation sets during cross-validation and for generative model initialisation). The objective score is thus the normalised 5-fold cross-validated IKLD averaged over five seeds. Note that we opted to use this score as the IKLD is a widely used measure, though this choice remains rather arbitrary.
%this score was an arbitrary choice made a priori as the IKLD is a widely used measure. 
Although alternative tuning objectives could impact the convergence rate of the SEs, we expect that they remain still slower than 1 over root-$n$ due to the highly data-adaptive nature of deep generative models.

\paragraph{Optimisation algorithm}
The \codepackage{Optuna} package \citep{optuna_2019} was used to optimise the objective score in the hyperparameter space. First $100$ hyperparameter configurations were randomly sampled from the specified hyperparameter search space. Subsequently, the Tree-structured Parzen Estimator algorithm (a Bayesian optimisation algorithm that uses a Gaussian mixture model as surrogate model) was applied to propose promising hyperparameter configurations until the hyperparameter optimisation study exceeded $12$ hours (on a single NVIDIA Ampere A100 GPU) for each generative model. To reduce computation costs, median pruning was enabled after $10$ hyperparameter proposals: if the hyperparameter configuration proposed yielded a moving average of the normalised 5-fold cross-validated IKLD after $x$ seed initialisation(s) that was worse than the median of the moving average obtained by previous hyperparameter configurations after the same number of seed initialisations, then this hyperparameter configuration was discarded.

\paragraph{Performance}
The optimisation study was performed for a random sample of size $n=500$ from the population. $419$ and $1812$ hyperparameter configurations were proposed for CTGAN and TVAE, respectively, of which $280$ and $1409$ were discarded by the pruning algorithm. The top three configurations were then evaluated on a random sample of sizes $n=50$ and $n=5000$ to check the applicability of the proposed hyperparameter configurations to other sample sizes. Based on this, the following configuration was chosen for \code{CTGAN}: $3$ hidden layers in the generator, each with $512$ nodes, $3$ hidden layers in the discriminator, each with $128$ nodes, $58$ epochs, $10$ iterations in the discriminator per iteration of generator, learning rate of $1.7\mathrm{e}{-5}$ (the same for generator and discriminator), dropout in generator of $88.9\%$, dropout in discriminator of $38.4\%$, weight decay of generator of $6.9\mathrm{e}{-6}$, and weight decay of discriminator of $1.4\mathrm{e}{-3}$. The following configuration was chosen for \code{TVAE}: embedding dimension of $64$, $1$ hidden layer with $512$ nodes each (the same for encoder and decoder), $961$ epochs, loss factor of $10$, dropout in encoder of $7.3\%$, dropout in decoder of $77.5\%$, and weight decay of $1.3\mathrm{e}{-4}$ (the same for encoder and decoder).

\newpage

\section{Simulation study}

\setcounter{figure}{0}
\renewcommand{\thefigure}{C\arabic{figure}}
\setcounter{table}{0}
\renewcommand{\thetable}{C\arabic{table}}

\subsection{Data generating mechanism} \label{sec:app_data_gen_mechanism}

Inspired by an applied medical setting, we create a hypothetical disease, defined by a low-dimensional tabular data generation mechanism. The dependency structure depicted by the Directed Acyclic Graph in Figure \ref{DAG_simulated_data} in the main text displays the presence of five variables, each of them chosen to obtain a mix of data types. In our hypothetical disease, it is assumed that a patient is observed at a given point in time. At this time, patient data about \emph{age}, \emph{disease stage}, \emph{biomarker}, and the random assignment of \emph{therapy} is gathered. The binary outcome variable \emph{death} is evaluated at a later time point, making this design a simplification, since we do not consider the data as longitudinal. 

The exact routine to reconstruct this data generating mechanism is presented in the pseudo-code in Algorithm \ref{algo:data_gen_mechanism}. \emph{Age} (continuous) follows a normal distribution with mean $50$ and standard deviation $10$. \emph{Disease stage} (ordinal) was generated according to a proportional odds cumulative logit model where an increase in \emph{age} causes an increase in the odds of having a \emph{disease stage} higher than a given stage $k$ ($\nu_{age}=-0.05$ and intercepts $\nu_{stage} = \{2, 3, 4\}$ for stage I-III). The variable \emph{biomarker} (continuous) is a quantification of the \emph{disease stage} and was also based on a generalised linear model, where \emph{biomarker} follows a gamma distribution and its mean changes in function of \emph{disease stage}. It was constructed in such a way that a higher \emph{disease stage} results in higher values for the \emph{biomarker} ($\gamma_{0} = 4$, $\gamma_{stage}=\{0, -1, -2, -3\}$ for stage I-IV, respectively). \emph{Therapy} (binary) is considered to be 1:1 randomly assigned and is therefore sampled from a Bernouilli distribution with a probability of $0.50$. The last variable, \emph{death} (binary), is generated by using a binomial logistic regression model in which the odds of \emph{death} increase with an increasing \emph{age} ($\beta_{age}=0.05$), a higher \emph{disease stage} ($\beta_{stage}=\{0, 0.50, 1.00, 1.50\}$ for stage I-IV, respectively), and absence of \emph{therapy} ($\beta_{therapy}=-0.50$).

\begin{algorithm}[H]
\SetAlgoLined
\SetKwInOut{Input}{input}
\SetKwInOut{Output}{output}
\DontPrintSemicolon
\BlankLine
\Input{Requested number of data records $n$.}
\Output{Dataframe $D$ with $n$ records, each made up of 5 attributes: $age$, $stage$, $biomarker$, $therapy$, $death$.}
\BlankLine
$D \leftarrow Empty\ dataframe$\;
\For{$i\gets1$ \KwTo $n$}{
    $age \leftarrow Normal(mean=50, std=10)$\;
    \BlankLine
    $\nu_{age} \leftarrow 0.05$\;
    $\nu_{I}, \nu_{II}, \nu_{III} \leftarrow 2, 3, 4$\;
    $cp_{I}, cp_{II}, cp_{III} \leftarrow Sigmoid(\nu_{I}- \nu_{age}\times age), Sigmoid(\nu_{II}-\nu_{age}\times age), Sigmoid(\nu_{III}-\nu_{age}\times age)$\;
    %$p_{I}, p_{II}, p_{III}, p_{IV} = cp_{I}, cp_{II}-cp_{I}, cp_{III}-cp_{II}, 1-cp_{III}$\;
    $stage \leftarrow Categorical(cat=[I, II, III, IV], probs=[cp_{I}, cp_{II}-cp_{I}, cp_{III}-cp_{II}, 1-cp_{III}])$\;
    \BlankLine
    $\gamma_0 \leftarrow 4$\;
    $\gamma_{I}, \gamma_{II}, \gamma_{III}, \gamma_{IV} \leftarrow 0, -1, -2, -3$\;
    $biomarker \leftarrow Gamma(shape = 25, scale = \frac{1}{25 \times (\gamma_0+\gamma_{stage})})$\;
    \BlankLine
    $therapy \leftarrow Categorical(cat=[False, True], p=[0.5, 0.5])$\;
    \BlankLine
    $\beta_{age}, \beta_{therapy} \leftarrow 0.05, -0.50$\;
    $\beta_{I}, \beta_{II}, \beta_{III}, \beta_{IV} \leftarrow 0, 0.50, 1.00, 1.50$\;
    $p_{death} \leftarrow Sigmoid(-3 + \beta_{age} \times age + \beta_{stage} + \beta_{therapy} \times therapy)$\;
    $death \leftarrow Categorical(cat=[False, True], p=[1-p_{death}, p_{death}])$
    \BlankLine
    $D_i \leftarrow \{age, stage, biomarker, therapy, death\}$
}
\BlankLine
\caption{Data Generating Mechanism for Hypothetical Disease.}
\label{algo:data_gen_mechanism}
\end{algorithm}

\subsection{Quality of synthetic data} \label{sec:app_quality_synth_data} 

We performed some additional analyses to assess the quality of the synthetic data obtained in our simulation study. 

\paragraph{Average IKLD} The inverse of the Kullback-Leibler divergence (IKLD) between original and synthetic data, averaged over 200 Monte Carlo runs and standardised between $0$ and $1$, is presented in Table \ref{table:appendix_IKLD}, where the tuned \code{CTGAN} and \code{TVAE} have higher IKLD than their default versions. However, the statistical approaches still seem to perform slightly better.

 \begin{table}[t]
     \caption{\label{table:appendix_IKLD} The IKLD between original and synthetic data, averaged over 200 Monte Carlo runs and standardised between 0 and 1. Higher values indicate similar datasets in terms of underlying distribution.}
     % \shrink \shrink
     \begin{center}
         \begin{tabular}{lccccc}
         \toprule
        \textbf{Generator}  & $n=50$ & $n=160$ & $n=500$ & $n=1600$ & $n=5000$ \\
        \midrule
        \textbf{Synthpop} & 0.939 & 0.976 & 0.994 & 0.995 & 0.996 \\
        \textbf{BN} & 0.934 & 0.984 & 0.997 & 0.998 & 0.999 \\
        \textbf{BN DAG} & 0.936 & 0.986 & 0.996 & 0.998 & 0.999 \\
        \textbf{Default CTGAN} & 0.853 & 0.905 & 0.861 & 0.903 & 0.933 \\
        \textbf{CTGAN} & 0.918 & 0.952 & 0.975 & 0.984 & 0.988 \\
        \textbf{Default TVAE} & 0.822 & 0.861 & 0.915 & 0.959 & 0.983 \\
        \textbf{TVAE} & 0.838 & 0.898 & 0.979 & 0.996 & 0.998 \\   
         \bottomrule
         \bottomrule
         \end{tabular}%
     \end{center}
 \end{table}

\paragraph{Failed generators} Our \code{Default CTGAN} model, as implemented in \codepackage{Synthcity}, could not be trained in one run (run $51$ for $n=500$) due to an internal error in the package. As such, it was not possible to generate synthetic data with \code{Default CTGAN} in this run. This comprises $0.1\%$ $(1/1000)$ of all \code{Default CTGANs} trained and $0.01\%$ $(1/8000)$ of all generators trained. The other generative models did not produce errors during training, so that synthetic data could be generated in every run.

\paragraph{Exact memorisation} A sanity check was conducted to ensure that no records of the original data were memorised by the generative model. \code{BN} made the following number of exact copies of the original data in the synthetic data: one record ($2.00\%$) for $n=50$ in five runs (runs $66, 103, 139, 147, 178$), one record ($0.63\%$) for $n=160$ in one run (run $55$), and one record ($0.02\%$) for $n=5000$ in one run (run $90$). \code{BN DAG} made the following exact copies: one record ($2.00\%$) for $n=50$ in two runs (runs $66, 103$). The other generative models did not make exact copies.

\paragraph{Non-estimable estimators} Due to sparse data, especially for small sample sizes, some of the $17$ estimators considered could not be estimated in a small subset of the $7999$ (original and synthetic) datasets, producing extremely small ($<1\mathrm{e}{-10}$) or large ($>1\mathrm{e}{2}$) standard errors. Overall, $0.35\%$ ($481/\num{135983}$) estimates could not be obtained, mainly for $n=50$ ($1.71\%; 466/\num{27200})$ and to a much lesser extent for $n=160$ ($0.02\%; 6/\num{27200})$ and $n=500$ ($0.03\%; 9/\num{27183})$. The number of estimates that could not be obtained are presented per estimator in Table \ref{non_estimable_per_estimator} and per generator in Table \ref{non_estimable_per_generator} for each sample size.

\begin{table}[H]
\caption{\label{non_estimable_per_estimator} The number of estimates that could not be obtained (due to sparse data) in the (original or synthetic) sample per sample size $n$ and per estimator.}
\begin{center}
\begin{tabular}{lcccccc}
\toprule
 &  $n = 50$ &  $n = 160$ &  $n = 500$ &  $n = 1600$ &  $n = 5000$ &  \textbf{{   All   }} \\ 
 \midrule
\textit{Proportion} & & & &\\
Proportion stage II &    11 &    0 &    0 & 0 &    0 &   \textbf{11} \\
Proportion stage III &    7 &    0 &    0 & 0 &    0 &    \textbf{7} \\
Proportion stage IV &   16 &    0 &    0 & 0 &    0 &   \textbf{16} \\ [0.15cm]
\textit{Gamma regression} & & & &\\
Effect stage II on death  &    11 &    0 &    0 & 0 &    0 &    \textbf{11} \\
Effect stage III on death &    7 &    0 &    0 & 0 &    0 &    \textbf{7} \\
Effect stage IV on death  &   16 &    0 &    0 & 0 &    0 &   \textbf{16} \\ [0.15cm]
\textit{Logistic regression} & & & &\\
Effect therapy on death       &    11 &    0 &    0 & 0 &    0 &    \textbf{11} \\
Effect stage II on death      &   91 &    1 &    5 & 0 &    0 &  \textbf{97} \\
Effect stage III on death     &  116 &    3 &    1 & 0 &    0 &  \textbf{120} \\
Effect stage IV on death      &  180 &    2 &    3 & 0 &    0 &  \textbf{185} \\ 
\midrule
\rule{0pt}{2.5ex}   
All                         &  466 &    6 &    9 & 0 &    0 &  \textbf{481} \\
\bottomrule
\bottomrule
\end{tabular}
\end{center}
\end{table}

\begin{table}[H]
\caption{\label{non_estimable_per_generator} The number of estimates that could not be obtained (due to sparse data) in the (original or synthetic) sample per sample size $n$ and per generator.}
\begin{center}
\begin{tabular}{lcccccccccc}
\toprule
&  Original &  Synthpop &  BN &  BN DAG &  \makecell[t]{Default\\CTGAN}  & CTGAN & \makecell[t]{Default\\TVAE}  & TVAE & & \textbf{{    All    }} \\ 
\midrule
$n = 50$ & 19 & 48 & 57 & 4 & 137 & 23 & 91 & 87 && \textbf{466} \\
$n = 160$ & 0 & 0 & 0 & 0 & 5 & 0 & 1 & 0 && \textbf{6} \\
$n = 500$ & 0 & 0 & 0 & 0 & 9 & 0 & 0 & 0 && \textbf{9} \\ 
$n = 1600$ & 0 & 0 & 0 & 0 & 0 & 0 & 0 & 0 && \textbf{0} \\ 
$n = 5000$ & 0 & 0 & 0 & 0 & 0 & 0 & 0 & 0 && \textbf{0} \\ 
\midrule
All & 19 & 48 & 57 & 4 & 151 & 23 & 92 & 87 && \textbf{481} \\
\bottomrule
\bottomrule
\end{tabular}
\end{center}
\end{table}

\subsection{Additional Results} \label{subsec:APPENDIX_Results}

\begin{figure}[H]
    \centering
    \includegraphics[width=\textwidth]{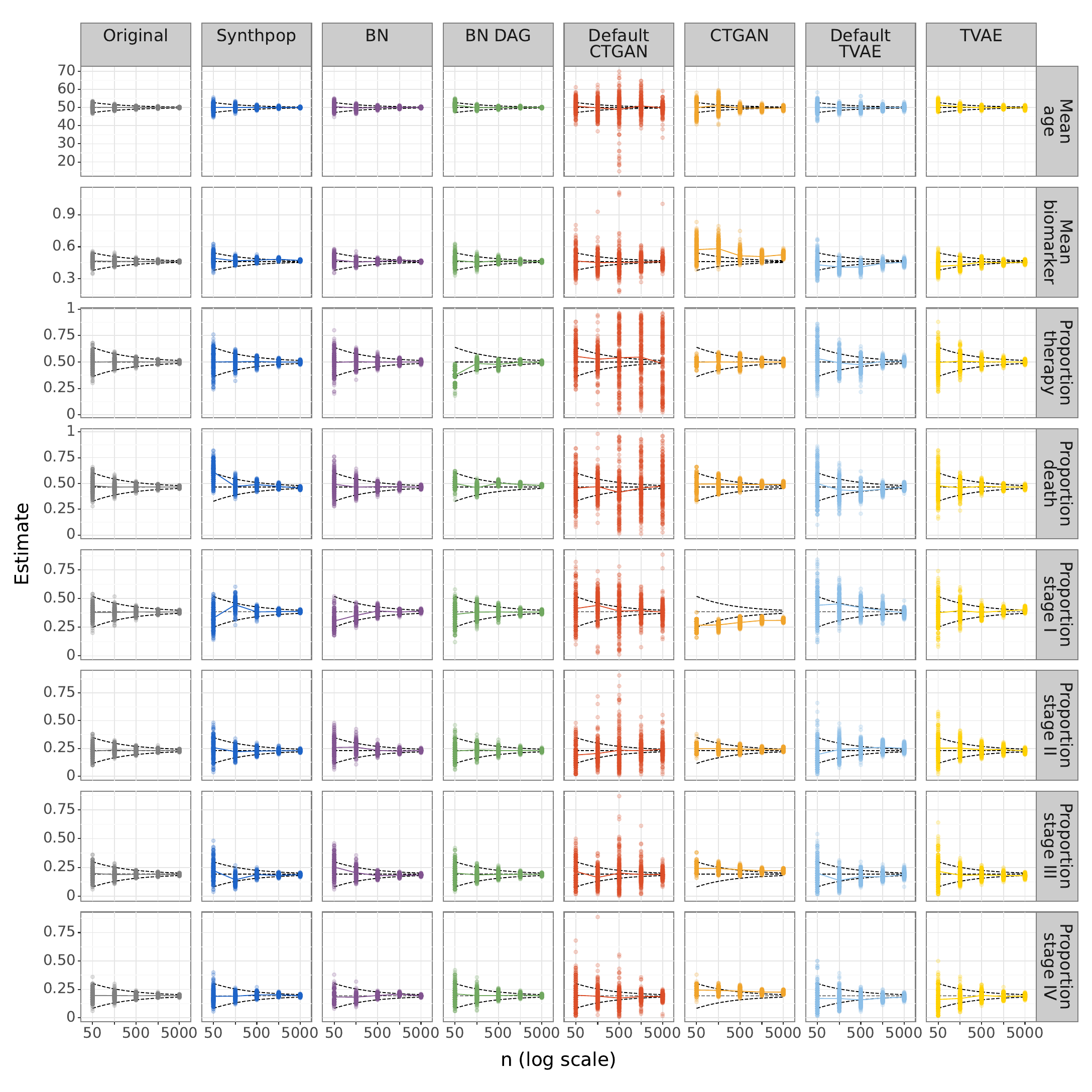}
    % \shrink
    \caption{Simulation study results for all mean and proportion estimators. Each dot is an estimate per Monte Carlo run (200 dots in total per value of $n$). The population parameter is represented by the horizontal dashed line. The dashed funnel indicates the behaviour of an unbiased and $\sqrt{n}$-consistent estimator based on observed data.}
    \label{appendix:bias_plot_1}
\end{figure}

\newpage 

\begin{figure}[H]
    \centering
    \includegraphics[width=\textwidth]{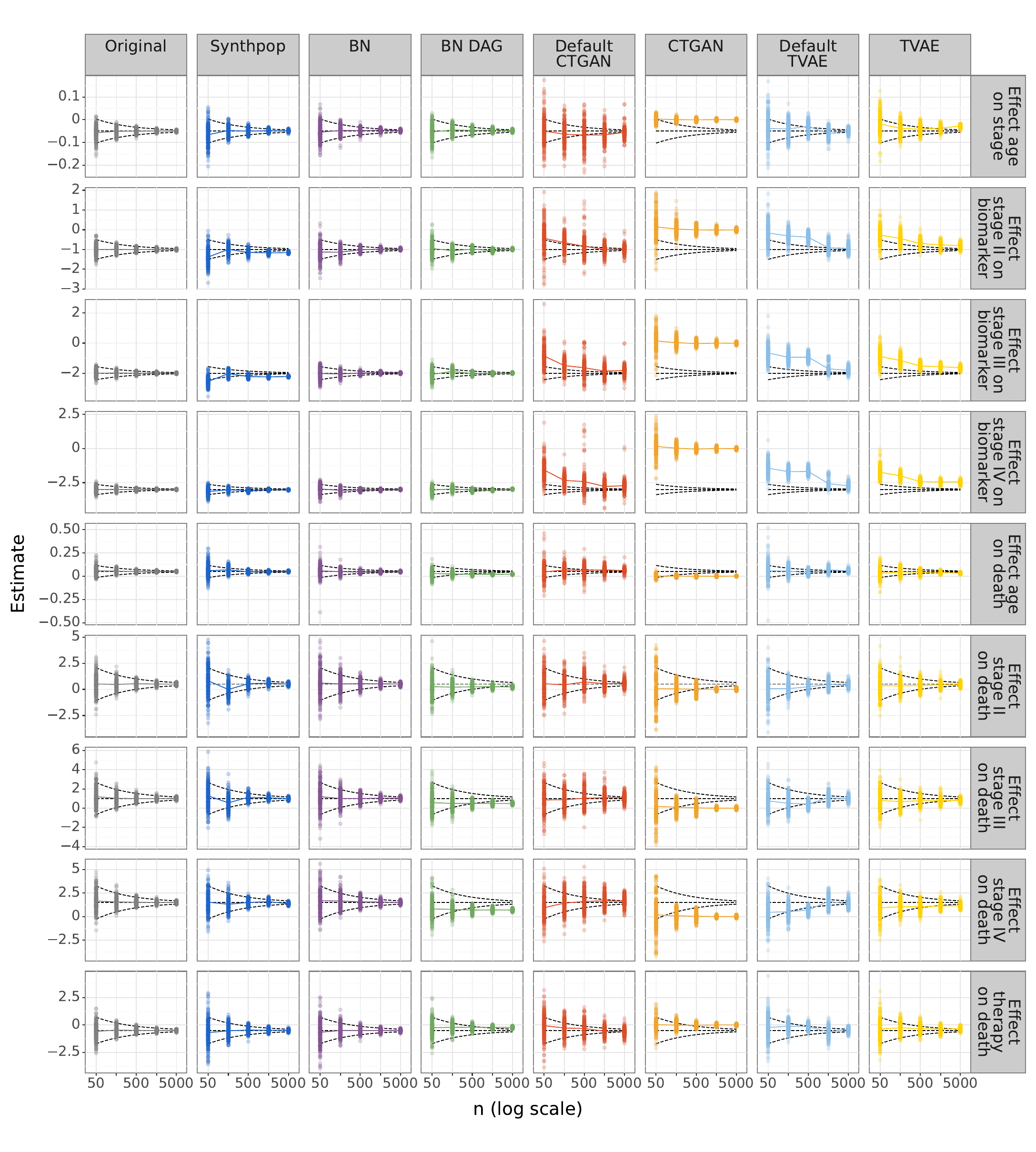}
    % \shrink
    \caption{Simulation study results for all regression coefficient estimators. Each dot is an estimate per Monte Carlo run (200 dots in total per value of $n$). The population parameter is represented by the horizontal dashed line. The dashed funnel indicates the behaviour of an unbiased and $\sqrt{n}$-consistent estimator based on observed data.}
    \label{appendix:bias_plot_2}
\end{figure}

\begin{landscape}
% statistical approaches (synthpop, BN en BN DAG)
\begin{table}
\caption{\label{tab:app_RE_all_stat} Relative error (RE) for all statistical approaches and all estimators, averaged over 200 Monte Carlo runs. $\text{RE}_{\hat{\theta}}$ is the relative bias of the estimates $\hat{\theta}$. $\text{RE}_{\hat{\sigma}_{\hat{\theta}}}$ is the relative error between the naive model-based ($\hat{\sigma}_{\hat{\theta}, naive}$) and the empirical standard error. Positive and negative values indicate a relative over- and underestimation.}
\resizebox{\columnwidth}{!}{
\begin{tabular}{lcccccccccccc}
\toprule
& \multicolumn{4}{c}{\textbf{Synthpop}} 
& \multicolumn{4}{c}{\textbf{BN}}
& \multicolumn{4}{c}{\textbf{BN DAG}} \\ 
{} & \multicolumn{2}{c}{$\text{RE}_{\hat{\theta}}$ (\%)} & \multicolumn{2}{c}{$\text{RE}_{\hat{\sigma}_{\hat{\theta}}}$ (\%)} & \multicolumn{2}{c}{$\text{RE}_{\hat{\theta}}$ (\%)} & \multicolumn{2}{c}{$\text{RE}_{\hat{\sigma}_{\hat{\theta}}}$ (\%)} & 
\multicolumn{2}{c}{$\text{RE}_{\hat{\theta}}$ (\%)} & \multicolumn{2}{c}{$\text{RE}_{\hat{\sigma}_{\hat{\theta}}}$ (\%)} \\
\textbf{Estimator}  & $n=50$ & $n=5000$ & $n=50$ & $n=5000$ & $n=50$ & $n=5000$ & $n=50$ & $n=5000$ &  $n=50$ & $n=5000$ &  $n=50$ & $n=5000$ \\
\midrule
\textit{Mean} &&&&&&&&&&&&\\ 
Mean age & 0.15 & 0.03 & -40.31 & -26.55 & 1.05 & 0.02 & -27.71 & -31.21 & 2.59 & -0.10 & -6.17 & -2.32 \\
Mean biomarker & 6.86 & 2.16 & -14.23 & 1.16 & 3.58 & -0.11 & -5.65 & 5.32 & 2.32 & 0.05 & -21.31 & -26.83 \\[0.15cm]
\textit{Proportion} &&&&&&&&&&&&\\
Proportion therapy & 0.18 & -0.12 & -33.26 & -31.47 & -0.92 & 0.09 & -33.05 & -33.94 & -25.28 & -0.93 & 7.00 & -4.56 \\
Proportion death & 31.39 & -2.35 & -25.40 & -14.07 & 5.76 & -0.26 & -29.85 & -34.33 & 7.27 & 3.23 & 45.38 & 25.23 \\
Proportion stage I & -14.22 & 0.83 & -27.41 & -7.91 & -21.00 & 1.51 & -9.42 & -18.97 & -5.89 & -0.02 & -24.20 & -29.00 \\
Proportion stage II & 11.15 & -0.77 & -31.54 & -23.17 & 11.68 & -0.13 & -28.55 & -26.40 & -2.22 & -0.07 & -24.85 & -33.21 \\
Proportion stage III & 17.03 & -1.48 & -30.17 & -25.81 & 32.74 & -2.86 & -27.63 & -19.00 & 5.57 & -0.26 & -30.35 & -22.50 \\
Proportion stage IV & -1.89 & 0.74 & -11.41 & -11.20 & -4.59 & -0.01 & 2.20 & -6.50 & 8.90 & 0.40 & -22.85 & -26.96 \\[0.15cm]
\textit{Cumulative regression} &&&&&&&&&&&&\\
Effect age on stage & 31.92 & -1.51 & -30.07 & -28.01 & 9.37 & -3.77 & -25.69 & -25.85 & 7.60 & -3.05 & -19.11 & -31.98 \\[0.15cm]
\textit{Gamma regression} &&&&&&&&&&&&\\
Effect stage II on biomarker & 38.07 & 15.61 & -14.43 & 3.93 & 12.84 & -1.63 & -26.98 & -15.29 & -5.45 & -1.75 & -11.56 & 2.40 \\
Effect stage III on biomarker & 27.82 & 11.13 & 9.38 & 14.45 & 1.08 & -0.19 & -26.03 & -11.83 & 3.92 & -0.27 & -16.07 & -3.35 \\
Effect stage IV on biomarker & 4.96 & 0.89 & 12.95 & 31.81 & -1.25 & -0.01 & -23.09 & -13.33 & 1.59 & -0.72 & 0.74 & -2.41 \\[0.15cm]
\textit{Logistic regression} &&&&&&&&&&&&\\
Effect age on death & 19.07 & 2.35 & -33.90 & -30.04 & 13.74 & -3.18 & -30.01 & -26.43 & -41.76 & -57.81 & -9.98 & -3.13 \\
Effect stage II on death & 62.75 & -1.30 & -29.50 & -24.99 & 20.75 & 2.10 & -21.82 & -29.32 & -49.14 & -54.56 & -12.22 & -6.64 \\
Effect stage III on death & 28.07 & -0.46 & -15.91 & -21.00 & 19.98 & 3.19 & -19.33 & -31.96 & -42.65 & -51.54 & -6.69 & 0.39 \\
Effect stage IV on death & 2.23 & -8.20 & 0.08 & -20.22 & 12.70 & 0.65 & -12.17 & -31.24 & -45.20 & -53.25 & -9.68 & -16.36 \\
Effect therapy on death & 38.53 & -2.74 & -32.78 & -30.45 & 31.88 & 0.71 & -27.53 & -31.80 & -50.16 & -55.24 & -11.63 & -15.83 \\
\bottomrule
\bottomrule
\end{tabular}
}
\end{table}
\end{landscape}

% DL approaches (default CTGAN, CTGAN, default TVAE, TVAE)
\begin{table}[H]
\caption{\label{tab:app_RE_all_DL} Relative error (RE) for all deep learning approaches and all estimators, averaged over 200 Monte Carlo runs. $\text{RE}_{\hat{\theta}}$ is the relative bias of the estimates $\hat{\theta}$. $\text{RE}_{\hat{\sigma}_{\hat{\theta}}}$ is the relative error between the naive model-based ($\hat{\sigma}_{\hat{\theta}, naive}$) and the empirical standard error. Positive and negative values indicate a relative over- and underestimation.}
\resizebox{16.70cm}{!}{
\begin{tabular}{lcccccccc}
\toprule
& \multicolumn{4}{c}{\textbf{Default CTGAN}} & \multicolumn{4}{c}{\textbf{CTGAN}} \\ 
{} & \multicolumn{2}{c}{$\text{RE}_{\hat{\theta}}$ (\%)} & \multicolumn{2}{c}{$\text{RE}_{\hat{\sigma}_{\hat{\theta}}}$ (\%)} & \multicolumn{2}{c}{$\text{RE}_{\hat{\theta}}$ (\%)} & \multicolumn{2}{c}{$\text{RE}_{\hat{\sigma}_{\hat{\theta}}}$ (\%)} \\
\textbf{Estimator}  & $n=50$ & $n=5000$ & $n=50$ & $n=5000$ &  $n=50$ & $n=5000$ &  $n=50$ & $n=5000$ \\
\midrule
\textit{Mean} &&&&&&&&\\ 
Mean age & 1.51 & -0.44 & -66.29 & -95.54 & -0.03 & -0.73 & -46.38 & -78.85 \\
Mean biomarker & 1.11 & -0.12 & -60.86 & -92.56 & 24.14 & 13.87 & -35.77 & -76.00 \\[0.15cm]
\textit{Proportion} &&&&&&&&\\
Proportion therapy & 10.64 & -3.10 & -52.41 & -97.84 & -0.42 & -0.18 & 131.41 & -56.79 \\
Proportion death & -2.39 & 2.09 & -56.80 & -96.64 & 6.04 & 5.29 & -9.52 & -47.44 \\
Proportion stage I & 7.50 & -0.58 & -53.63 & -91.41 & -30.49 & -19.44 & 37.80 & -43.39 \\
Proportion stage II & -17.69 & 4.88 & -48.00 & -92.24 & 6.63 & 4.60 & 46.34 & -52.61 \\
Proportion stage III & 13.00 & -0.97 & -48.69 & -84.79 & 26.72 & 16.35 & 33.84 & -51.59 \\
Proportion stage IV & 2.79 & -3.74 & -54.53 & -81.30 & 26.37 & 17.06 & 48.08 & -47.06 \\[0.15cm]
\textit{Cumulative regression} &&&&&&&&\\
Effect age on stage & -0.13 & 11.58 & -40.15 & -85.57 & -105.97 & -100.03 & 13.91 & -3.23 \\[0.15cm]
\textit{Gamma regression} &&&&&&&&\\
Effect stage II on biomarker & -56.70 & -2.90 & -23.39 & -77.69 & -114.42 & -99.61 & -20.44 & -7.70 \\
Effect stage III on biomarker & -57.18 & -9.09 & -28.55 & -87.97 & -107.06 & -99.61 & -14.75 & -8.60 \\
Effect stage IV on biomarker & -47.37 & -9.21 & -40.31 & -91.84 & -104.96 & -99.62 & -22.89 & -7.25 \\[0.15cm]
\textit{Logistic regression} &&&&&&&&\\
Effect age on death & -8.24 & 21.74 & -38.49 & -78.71 & -96.42 & -97.81 & -9.41 & -1.18 \\
Effect stage II on death & 6.99 & 14.60 & -17.44 & -71.45 & -88.74 & -99.15 & -44.22 & -11.08 \\
Effect stage III on death & -14.20 & 7.92 & -20.60 & -74.80 & -82.06 & -99.44 & -43.57 & -18.69 \\
Effect stage IV on death & -36.57 & 10.05 & -16.64 & -78.24 & -87.18 & -99.48 & -48.96 & -9.38 \\
Effect therapy on death & -87.68 & 24.40 & -20.59 & -72.08 & -104.72 & -101.02 & 17.50 & -13.89 \\
\midrule
& \multicolumn{4}{c}{\textbf{Default TVAE}} & \multicolumn{4}{c}{\textbf{TVAE}} \\ 
{} & \multicolumn{2}{c}{$\text{RE}_{\hat{\theta}}$ (\%)} & \multicolumn{2}{c}{$\text{RE}_{\hat{\sigma}_{\hat{\theta}}}$ (\%)} & \multicolumn{2}{c}{$\text{RE}_{\hat{\theta}}$ (\%)} & \multicolumn{2}{c}{$\text{RE}_{\hat{\sigma}_{\hat{\theta}}}$ (\%)} \\
\textbf{Estimator}  & $n=50$ & $n=5000$ & $n=50$ & $n=5000$ &  $n=50$ & $n=5000$ &  $n=50$ & $n=5000$ \\
\midrule
\textit{Mean} &&&&&&&&\\ 
Mean age & 0.04 & 0.14 & -65.82 & -85.71 & 2.79 & -0.38 & -40.97 & -78.55 \\
Mean biomarker & -6.49 & -1.34 & -58.62 & -77.81 & -9.67 & -1.09 & -48.37 & -63.51 \\[0.15cm]
\textit{Proportion} &&&&&&&&\\
Proportion therapy & 6.40 & 1.26 & -57.92 & -65.27 & -1.16 & -0.14 & -46.33 & -42.72 \\
Proportion death & 7.03 & 0.15 & -58.67 & -66.89 & 2.72 & -0.48 & -48.78 & -41.14 \\
Proportion stage I & 14.86 & -2.10 & -53.17 & -71.80 & -2.17 & 4.75 & -42.91 & -50.70 \\
Proportion stage II & -15.73 & 8.37 & -53.19 & -77.58 & 9.11 & -2.04 & -46.27 & -56.65 \\
Proportion stage III & 4.73 & -0.04 & -49.99 & -77.58 & 13.68 & -6.59 & -52.44 & -57.94 \\
Proportion stage IV & -11.56 & -5.83 & -48.23 & -66.16 & -16.18 & -0.50 & -45.01 & -59.32 \\[0.15cm]
\textit{Cumulative regression} &&&&&&&&\\
Effect age on stage & -22.49 & -3.86 & -26.09 & -72.83 & -62.29 & -43.48 & -15.99 & -53.27 \\[0.15cm]
\textit{Gamma regression} &&&&&&&&\\
Effect stage II on biomarker & -83.35 & -4.19 & -4.47 & -80.60 & -73.01 & -19.87 & 2.13 & -60.54 \\
Effect stage III on biomarker & -67.21 & -9.37 & -14.38 & -80.47 & -55.66 & -18.20 & -21.37 & -62.77 \\
Effect stage IV on biomarker & -52.00 & -9.01 & -22.71 & -83.32 & -42.03 & -18.43 & -22.65 & -66.43 \\[0.15cm]
\textit{Logistic regression} &&&&&&&&\\
Effect age on death & 17.87 & 15.48 & -34.93 & -74.68 & -36.16 & -35.79 & -11.26 & -44.03 \\
Effect stage II on death & -85.24 & -10.42 & -21.16 & -65.43 & -28.67 & -8.57 & -10.28 & -52.84 \\
Effect stage III on death & -24.49 & -4.02 & -11.77 & -67.33 & -19.32 & -20.20 & -9.94 & -53.38 \\
Effect stage IV on death & -69.64 & -10.59 & -13.54 & -74.70 & -37.17 & -23.28 & -11.38 & -64.08 \\
Effect therapy on death & -56.22 & -18.13 & -27.70 & -68.74 & -26.43 & -35.56 & -24.90 & -56.18 \\
\bottomrule
\bottomrule
\end{tabular}
}
\end{table}

\begin{landscape}
\begin{table}[htbp]
\caption{\label{tab:app_convergence_rate_SE} Estimated exponent $a$ for the power law convergence rate $n^{-a}$ for the empirical standard error (SE).}
\resizebox{\columnwidth}{!}{
\begin{tabular}{lcccccccc}
\toprule
&& \multicolumn{7}{c}{\textbf{Generator}} \\ 
\cmidrule(r){3-9}
&  \textbf{Original} &  \textbf{Synthpop} &  \makecell[t]{\textbf{BN}} & \makecell[t]{\textbf{BN DAG}} &  \makecell[t]{\textbf{Default}\\\textbf{CTGAN}} &  \textbf{CTGAN} &  \makecell[t]{\textbf{Default}\\\textbf{TVAE}} & \textbf{TVAE} \\
\textbf{Estimator, SE} &  &  &  & &  & & & \\
\midrule
\textit{Mean} &&&&&&&&\\ 
Mean age &  0.49 [0.47; 0.52] &  0.53 [0.50; 0.55] &  0.49 [0.45; 0.52] &    0.49 [0.47; 0.52] &   0.03 [-0.34; 0.41] &   0.40 [0.13; 0.66] &   0.23 [0.11; 0.36] &  0.22 [0.03; 0.42] \\
Mean biomarker &  0.48 [0.44; 0.53] &  0.51 [0.44; 0.58] &  0.51 [0.48; 0.53] &    0.49 [0.47; 0.51] &   0.12 [-0.10; 0.35] &   0.34 [0.26; 0.42] &   0.30 [0.17; 0.43] &  0.36 [0.29; 0.42] \\[0.15cm]
\textit{Proportion} &&&&&&&&\\ 
Proportion therapy &  0.50 [0.43; 0.56] &  0.50 [0.46; 0.54] &  0.48 [0.46; 0.51] &    0.46 [0.38; 0.54] &  -0.18 [-0.37; 0.01] &  0.19 [-0.14; 0.52] &   0.46 [0.36; 0.55] &  0.51 [0.42; 0.61] \\
Proportion death &  0.51 [0.49; 0.53] &  0.52 [0.49; 0.54] &  0.50 [0.44; 0.55] &    0.48 [0.40; 0.55] &  -0.09 [-0.28; 0.10] &   0.41 [0.28; 0.53] &   0.45 [0.40; 0.50] &  0.53 [0.44; 0.62] \\
Proportion stage I &  0.48 [0.43; 0.52] &  0.54 [0.51; 0.57] &  0.46 [0.41; 0.52] &    0.48 [0.43; 0.53] &   0.13 [-0.02; 0.29] &   0.28 [0.11; 0.44] &   0.38 [0.34; 0.41] &  0.47 [0.41; 0.53] \\
Proportion stage II   &  0.48 [0.46; 0.51] &  0.51 [0.47; 0.56] &  0.51 [0.49; 0.53] &    0.47 [0.45; 0.49] &   0.06 [-0.17; 0.29] &   0.26 [0.13; 0.39] &   0.31 [0.26; 0.37] &  0.44 [0.39; 0.49] \\
Proportion stage III &  0.51 [0.46; 0.56] &  0.51 [0.41; 0.61] &  0.56 [0.47; 0.64] &    0.51 [0.45; 0.56] &   0.18 [-0.16; 0.52] &   0.29 [0.17; 0.41] &   0.31 [0.20; 0.42] &  0.47 [0.37; 0.57] \\
Proportion stage IV &  0.48 [0.44; 0.52] &  0.47 [0.38; 0.57] &  0.49 [0.42; 0.55] &    0.50 [0.46; 0.54] &    0.29 [0.16; 0.43] &   0.29 [0.16; 0.41] &   0.39 [0.35; 0.43] &  0.41 [0.38; 0.44] \\[0.15cm]
\textit{Cumulative regression} &&&&&&&&\\ 
Effect age on stage &  0.53 [0.49; 0.56] &  0.54 [0.48; 0.60] &  0.51 [0.48; 0.55] &    0.48 [0.41; 0.55] &    0.20 [0.06; 0.33] &   0.40 [0.30; 0.50] &   0.34 [0.20; 0.48] &  0.44 [0.38; 0.51] \\[0.15cm]
\textit{Gamma regression} &&&&&&&&\\ 
Effect stage II on biomarker &  0.51 [0.47; 0.56] &  0.54 [0.50; 0.58] &  0.55 [0.51; 0.60] &    0.55 [0.52; 0.59] &    0.27 [0.08; 0.46] &   0.55 [0.48; 0.61] &  0.18 [-0.04; 0.41] &  0.26 [0.23; 0.30] \\
Effect stage III on biomarker &  0.50 [0.47; 0.54] &  0.52 [0.46; 0.58] &  0.56 [0.52; 0.61] &    0.55 [0.52; 0.58] &    0.16 [0.00; 0.31] &   0.52 [0.44; 0.61] &  0.19 [-0.01; 0.39] &  0.32 [0.24; 0.40] \\
Effect stage IV on biomarker  &  0.50 [0.47; 0.53] &  0.54 [0.49; 0.58] &  0.57 [0.50; 0.64] &    0.52 [0.49; 0.55] &   0.10 [-0.19; 0.39] &   0.55 [0.43; 0.67] &  0.15 [-0.08; 0.37] &  0.28 [0.14; 0.43] \\[0.15cm]
\textit{Logistic regression} &&&&&&&&\\ 
Effect age on death  &  0.56 [0.49; 0.63] &  0.59 [0.45; 0.72] &  0.56 [0.45; 0.67] &    0.57 [0.55; 0.59] &    0.28 [0.12; 0.44] &   0.47 [0.31; 0.63] &   0.37 [0.13; 0.61] &  0.51 [0.45; 0.58] \\
Effect stage II on death  &  0.52 [0.47; 0.57] &  0.56 [0.49; 0.63] &  0.55 [0.52; 0.57] &    0.55 [0.50; 0.59] &    0.30 [0.23; 0.36] &   0.68 [0.41; 0.95] &   0.34 [0.09; 0.60] &  0.37 [0.28; 0.47] \\
Effect stage III on death &  0.52 [0.46; 0.59] &  0.55 [0.47; 0.64] &  0.52 [0.48; 0.57] &    0.55 [0.50; 0.60] &    0.24 [0.13; 0.35] &   0.65 [0.38; 0.93] &   0.30 [0.05; 0.54] &  0.36 [0.27; 0.46] \\
Effect stage IV on death &  0.53 [0.48; 0.57] &  0.51 [0.43; 0.59] &  0.50 [0.49; 0.51] &    0.52 [0.47; 0.56] &    0.20 [0.14; 0.27] &   0.71 [0.44; 0.97] &  0.23 [-0.09; 0.56] &  0.34 [0.28; 0.40] \\
Effect therapy on death &  0.53 [0.48; 0.58] &  0.56 [0.49; 0.63] &  0.55 [0.47; 0.62] &    0.53 [0.46; 0.59] &    0.25 [0.19; 0.31] &   0.51 [0.40; 0.61] &  0.29 [-0.07; 0.66] &  0.40 [0.30; 0.50] \\
\bottomrule
\bottomrule
\end{tabular}
}
\end{table}

\end{landscape}

\begin{landscape}
\begin{table}[htbp]
\caption{\label{tab:app_convergence_rate_bias} Estimated exponent $a$ for the power law convergence rate $n^{-a}$ for the empirical bias.}
\resizebox{\columnwidth}{!}{
\begin{tabular}{lcccccccc}
\toprule
&& \multicolumn{7}{c}{\textbf{Generator}} \\
\cmidrule(r){3-9}
&  \textbf{Original} &  \textbf{Synthpop} &  \makecell[t]{\textbf{BN}} & \makecell[t]{\textbf{BN DAG}} &  \makecell[t]{\textbf{Default}\\\textbf{CTGAN}} &  \textbf{CTGAN} &  \makecell[t]{\textbf{Default}\\\textbf{TVAE}} & \textbf{TVAE} \\
\textbf{Estimator, bias} &  &  &  & &  & & & \\
\midrule
\textit{Mean} &&&&&&&&\\ 
Mean age &   0.64 [0.40; 0.89] &    0.38 [0.18; 0.58] &   0.86 [0.39; 1.32] &    0.45 [-0.61; 1.50] &   0.08 [-0.61; 0.78] &   -0.42 [-1.62; 0.77] &  -0.09 [-0.99; 0.81] &   0.54 [-0.37; 1.45] \\
Mean biomarker &   0.47 [0.17; 0.77] &   0.10 [-0.43; 0.63] &  0.52 [-0.40; 1.44] &     0.81 [0.20; 1.41] &   0.42 [-0.34; 1.18] &    0.18 [-0.07; 0.43] &   0.37 [-0.14; 0.88] &   0.47 [-0.03; 0.96] \\[0.15cm]
\textit{Proportion} &&&&&&&&\\ 
Proportion therapy &   0.42 [0.09; 0.76] &   0.12 [-0.52; 0.75] &   0.63 [0.13; 1.14] &    0.81 [-0.12; 1.74] &   0.17 [-0.18; 0.52] &    0.10 [-0.34; 0.54] &   0.29 [-0.17; 0.76] &    0.63 [0.05; 1.22] \\
Proportion death &   1.24 [0.64; 1.85] &   0.45 [-0.52; 1.43] &  0.43 [-0.47; 1.34] &    0.02 [-0.63; 0.67] &  -0.06 [-0.88; 0.75] &    0.04 [-0.01; 0.09] &   0.69 [-0.29; 1.67] &   0.58 [-0.29; 1.44] \\
Proportion stage I &  0.65 [-0.02; 1.32] &   0.79 [-0.52; 2.09] &   0.72 [0.01; 1.42] &    0.88 [-0.41; 2.18] &   0.58 [-0.02; 1.19] &     0.11 [0.06; 0.17] &    0.49 [0.20; 0.78] &  -0.12 [-0.53; 0.30] \\
Proportion stage II &   0.92 [0.49; 1.36] &    0.57 [0.46; 0.67] &  0.91 [-0.24; 2.05] &     0.76 [0.34; 1.17] &   0.41 [-0.48; 1.30] &     0.10 [0.02; 0.19] &  -0.00 [-0.52; 0.51] &   0.34 [-0.02; 0.70] \\
Proportion stage III &  0.46 [-0.51; 1.43] &   0.57 [-0.12; 1.27] &   0.47 [0.14; 0.81] &     0.70 [0.40; 0.99] &    0.66 [0.31; 1.02] &     0.13 [0.06; 0.19] &   0.90 [-0.87; 2.66] &   0.11 [-0.42; 0.63] \\
Proportion stage IV &  0.21 [-0.10; 0.51] &   0.04 [-0.69; 0.77] &  1.01 [-0.76; 2.78] &    0.47 [-0.23; 1.17] &  -0.12 [-0.72; 0.48] &     0.11 [0.07; 0.15] &   0.14 [-0.15; 0.42] &    0.72 [0.29; 1.15] \\[0.15cm]
\textit{Cumulative regression} &&&&&&&&\\ 
Effect age on stage &   0.76 [0.27; 1.25] &   0.55 [-0.28; 1.39] &  0.12 [-0.17; 0.42] &    0.17 [-0.10; 0.44] &  -0.80 [-2.48; 0.88] &    0.01 [-0.01; 0.03] &   0.29 [-0.33; 0.92] &   0.05 [-0.41; 0.51] \\[0.15cm]
\textit{Gamma regression} &&&&&&&&\\ 
Effect stage II on biomarker &  0.05 [-0.71; 0.81] &  -0.09 [-1.29; 1.12] &  0.54 [-0.11; 1.19] &    0.20 [-0.25; 0.64] &   1.02 [-0.51; 2.56] &    0.03 [-0.00; 0.06] &    0.70 [0.23; 1.17] &    0.30 [0.20; 0.40] \\
Effect stage III on biomarker &  0.02 [-0.87; 0.90] &  -0.03 [-1.02; 0.95] &  0.35 [-0.33; 1.04] &     0.58 [0.09; 1.08] &    0.43 [0.17; 0.69] &    0.01 [-0.00; 0.03] &    0.46 [0.16; 0.75] &    0.25 [0.14; 0.36] \\
Effect stage IV on biomarker &  0.37 [-0.29; 1.03] &   0.34 [-0.52; 1.19] &  0.90 [-0.36; 2.15] &    0.15 [-0.05; 0.34] &    0.38 [0.15; 0.61] &    0.01 [-0.01; 0.02] &    0.40 [0.13; 0.67] &    0.20 [0.05; 0.34] \\[0.15cm]
\textit{Logistic regression} &&&&&&&&\\ 
Effect age on death &  0.76 [-0.21; 1.74] &   0.52 [-0.01; 1.05] &  0.16 [-0.48; 0.79] &   -0.07 [-0.14; 0.01] &  -0.17 [-0.60; 0.25] &   -0.00 [-0.02; 0.01] &  -0.08 [-0.50; 0.34] &  -0.05 [-0.55; 0.45] \\
Effect stage II on death &  0.49 [-0.27; 1.25] &   0.81 [-0.00; 1.63] &  0.34 [-0.34; 1.01] &   -0.01 [-0.06; 0.05] &  -0.08 [-0.69; 0.52] &  -0.03 [-0.05; -0.00] &    0.58 [0.10; 1.07] &    0.30 [0.08; 0.52] \\
Effect stage III on death &  0.57 [-0.31; 1.45] &    0.87 [0.14; 1.60] &  0.14 [-1.65; 1.92] &   -0.03 [-0.08; 0.01] &   0.14 [-0.37; 0.65] &  -0.04 [-0.07; -0.01] &   0.47 [-0.21; 1.15] &   0.03 [-0.15; 0.20] \\
Effect stage IV on death &   0.62 [0.45; 0.79] &  -0.16 [-1.37; 1.05] &   0.63 [0.32; 0.93] &  -0.03 [-0.06; -0.00] &   0.14 [-0.52; 0.80] &  -0.03 [-0.05; -0.01] &    0.51 [0.07; 0.95] &    0.10 [0.05; 0.15] \\
Effect therapy on death &   0.70 [0.53; 0.87] &    0.52 [0.11; 0.93] &  0.20 [-2.46; 2.86] &   -0.02 [-0.07; 0.03] &   0.34 [-0.01; 0.68] &    0.00 [-0.03; 0.04] &   0.29 [-0.03; 0.61] &   0.02 [-0.24; 0.27] \\
\bottomrule
\bottomrule
\end{tabular}
}
\end{table}
\end{landscape}

\begin{figure}[H]
    \centering
    \includegraphics[width=\textwidth]{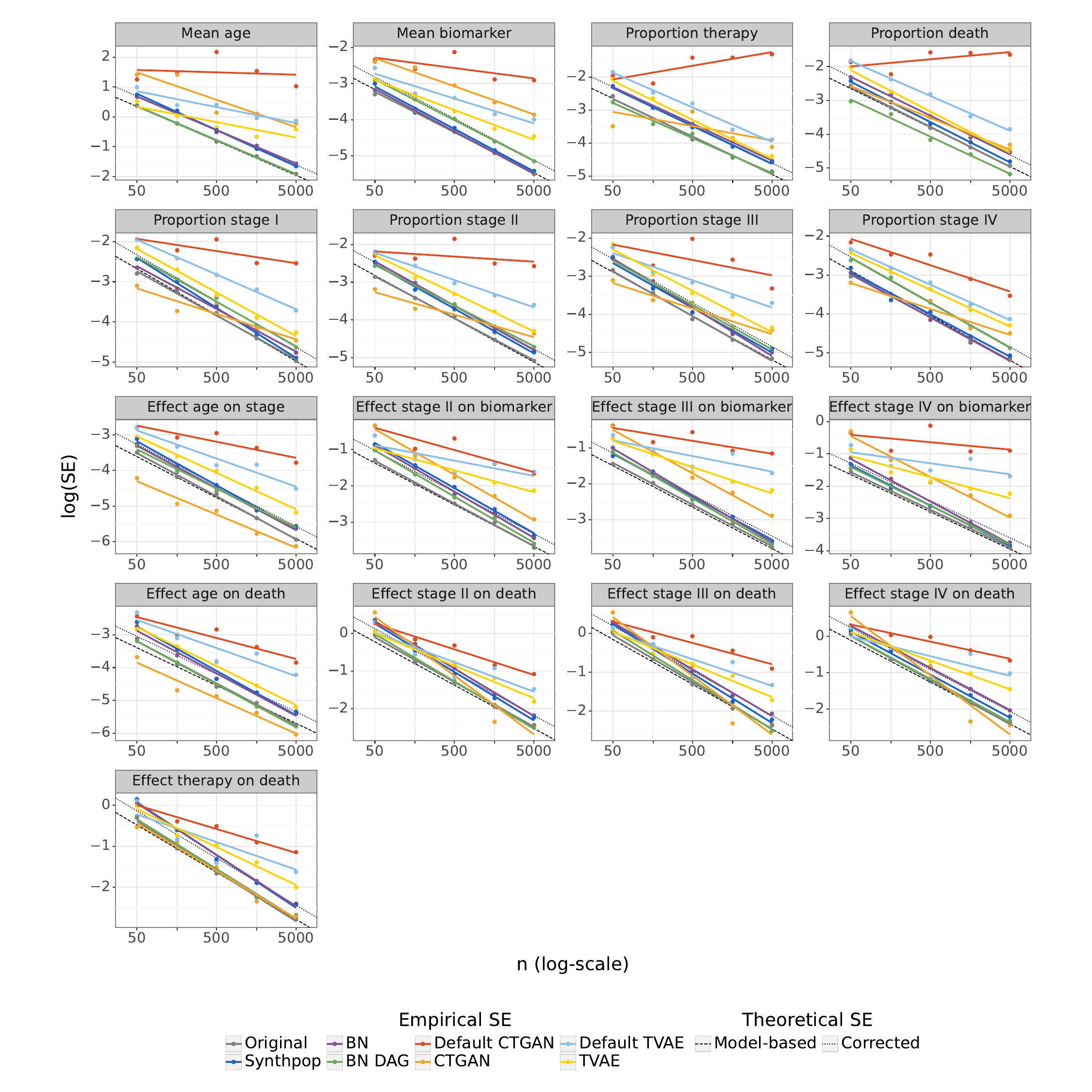}
    \caption{Convergence rate of the empirical standard error (SE). If the SE is of the form SE $= cn^{-a}$, where $c$ is a constant, then $\log{(SE)} = \log{c} + (-a) \log{(n)}$. Therefore slope $a$ represents the convergence rate and the vertical offset $\log{c}$ the log asymptotic variance. The dashed line indicates the behaviour of the SE of an unbiased and $\sqrt{n}$-consistent estimator based on observed data, whereas the dotted line indicates the assumed behaviour of the SE of the same estimator based on synthetic data, following the correction proposed by \cite{raab2016practical}. Note that the asymptotic variances of the effect of \emph{age} on \emph{stage} and the effect of \emph{age} on \emph{death} by \code{CTGAN}, and the proportion of \emph{death} by \code{BN DAG} are smaller than on the observed data, as they deliver a biased effect due to generative model misspecification.}
    \label{appendix:plot_convergence_se}
\end{figure}

\newpage 

\section{Case study}

\setcounter{figure}{0}
\renewcommand{\thefigure}{D\arabic{figure}}
\setcounter{table}{0}
\renewcommand{\thetable}{D\arabic{table}}

\subsection{Quality of synthetic data} \label{sec:app_quality_synth_data_case_study}

We performed some additional analyses to assess the quality of the synthetic data obtained in our case study.

\paragraph{Average IKLD} The inverse of the Kullback-Leibler divergence (IKLD) between original and synthetic data, averaged over 200 Monte Carlo runs and standardised between $0$ and $1$, is $0.923$ for \code{Default CTGAN} and $0.955$ for \code{Synthpop}.

\paragraph{Failed generators} Our \code{Default CTGAN} model, as implemented in \codepackage{Synthcity}, could not be trained in three runs (runs $147, 156, 175$) due to an internal error in the package. As such, it was not possible to generate synthetic data with \code{Default CTGAN} in these runs. This comprises $1.50\%$ $(3/200)$ of all \code{Default CTGANs} trained and $0.75\%$ $(3/400)$ of all generators trained. \code{Synthpop} did not produce errors during training, so that synthetic data could be generated in every run.

\paragraph{Exact memorisation} A sanity check was conducted to ensure that no records of the original data were memorised by the generative model. \code{Synthpop} copied one original record ($0.02\%$) in three runs (runs $5, 162, 197$). \code{Default CTGAN} did not make exact copies.

\subsubsection{Non-estimable estimators}

The sample mean of $age$ and the sample effect of $age$ on $income$ (estimated via a logistic regression model) could be estimated in all (original and synthetic) datasets.

\subsection{Additional results} \label{sec:app_results_case_study}

\begin{figure}[H]
\centering
\includegraphics[scale=0.525]{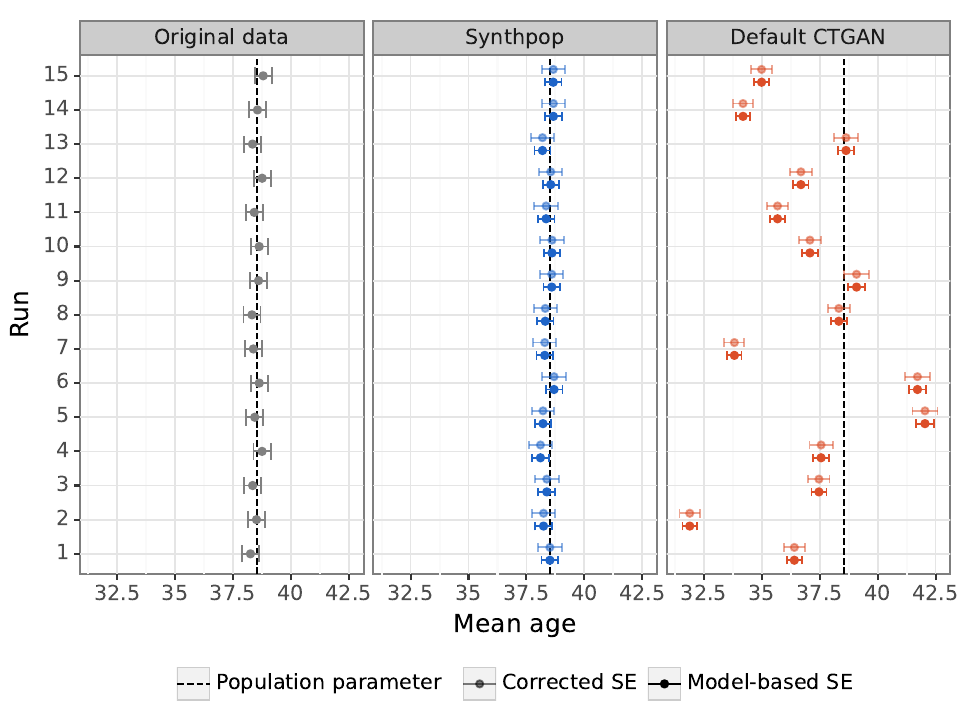}
\caption{Empirical coverage of $95\%$ CIs for mean of $age$, with model-based and corrected standard error (SE). }
\label{fig:adult_dataset}
\end{figure}

\newpage 

\section{Differentially private generators} \label{sec:app_dp_methods}

\setcounter{figure}{0}
\renewcommand{\thefigure}{E\arabic{figure}}
\setcounter{table}{0}
\renewcommand{\thetable}{E\arabic{table}}

Here, we present the results for three state-of-the-art differentially private models: \code{PrivBayes} \citep{zhang2017privbayes}, \code{DP-GAN} \citep{xie2018differentially}, and \code{PATE-GAN} \citep{jordon2018pate}.

\begin{figure}[H]
    \centering
    \includegraphics[width=0.60\textwidth]{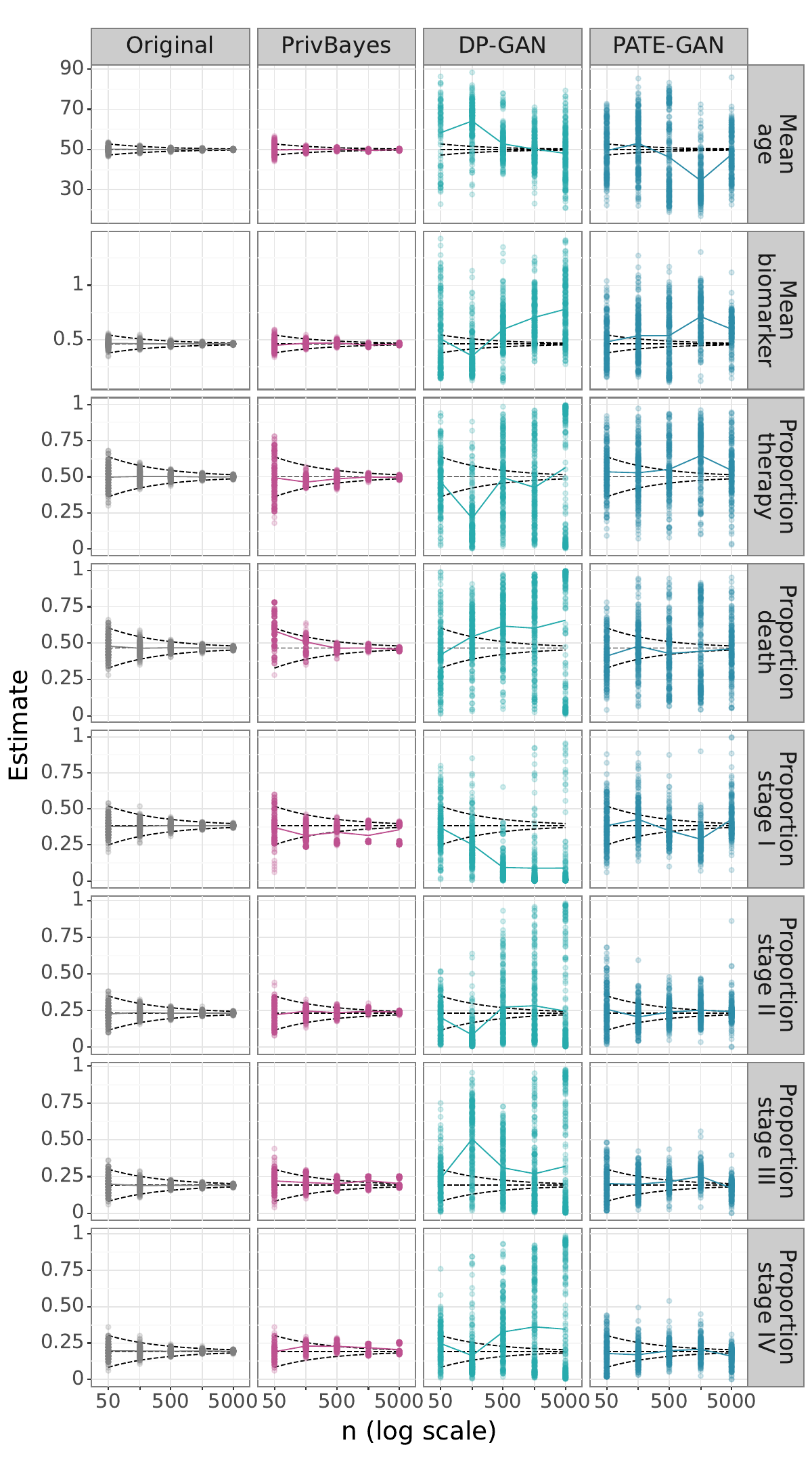}
    % \shrink
    \caption{Simulation study results for all mean and proportion estimators. Each dot is an estimate per Monte Carlo run (200 dots in total per value of $n$). The population parameter is represented by the horizontal dashed line. The dashed funnel indicates the behaviour of an unbiased and $\sqrt{n}$-consistent estimator based on observed data.}
    \label{appendix:bias_plot_3}
\end{figure}

\begin{figure}[H]
    \centering
    \includegraphics[width=0.60\textwidth]{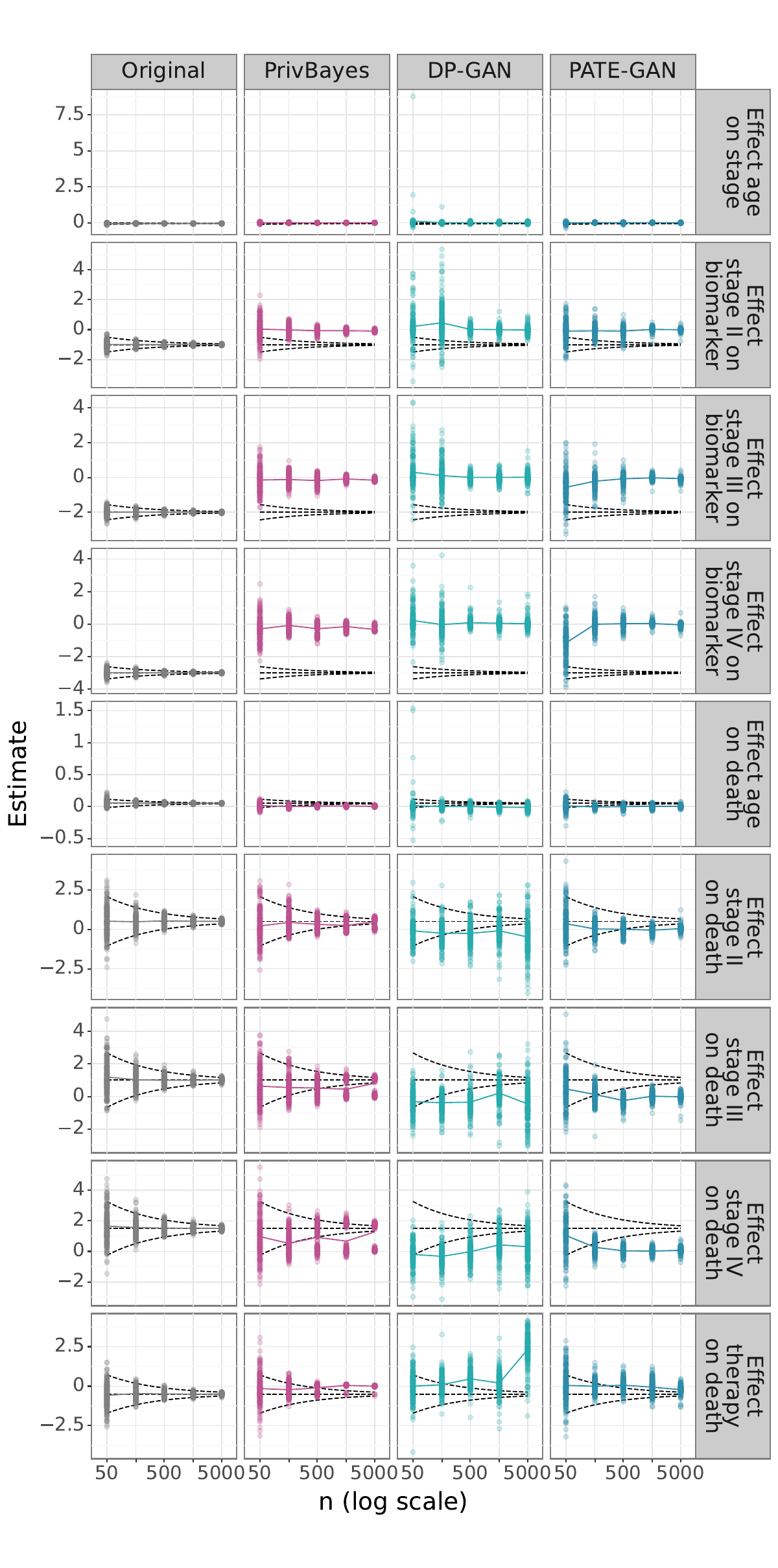}
    % \shrink
    \caption{Simulation study results for all regression coefficient estimators. Each dot is an estimate per Monte Carlo run (200 dots in total per value of $n$). The population parameter is represented by the horizontal dashed line. The dashed funnel indicates the behaviour of an unbiased and $\sqrt{n}$-consistent estimator based on observed data.}
    \label{appendix:bias_plot_4}
\end{figure}

\begin{table}[htbp]
\caption{\label{tab:app_convergence_rate_se_DP} Estimated exponent $a$ for the power law convergence rate $n^{-a}$ for the empirical standard error (SE).}
\resizebox{\columnwidth}{!}{
\begin{tabular}{lcccc}
\toprule
&& \multicolumn{3}{c}{\textbf{Generator}} \\
\cmidrule(r){3-5}
&  \textbf{Original} &  \textbf{PrivBayes} & \textbf{DP-GAN} & \textbf{PATE-GAN} \\
\textbf{Estimator, SE} &  &  &  & \\
\midrule
\textit{Mean} &&&&\\ 
Mean age &  0.49 [0.47; 0.52] &   0.50 [0.27; 0.73] &    0.06 [-0.17; 0.30] &   0.04 [-0.18; 0.26] \\
Mean biomarker &  0.48 [0.44; 0.53] &   0.45 [0.37; 0.53] &    0.02 [-0.18; 0.22] &   0.07 [-0.11; 0.25] \\[0.15cm]
\textit{Proportion} &&&&\\ 
Proportion therapy &  0.50 [0.43; 0.56] &   0.60 [0.36; 0.83] &  -0.18 [-0.30; -0.07] &  -0.02 [-0.08; 0.03] \\
Proportion death &  0.51 [0.49; 0.53] &   0.58 [0.45; 0.70] &   -0.14 [-0.31; 0.02] &  -0.08 [-0.25; 0.09] \\
Proportion stage I &  0.48 [0.43; 0.52] &  0.09 [-0.08; 0.25] &   -0.07 [-0.29; 0.15] &   0.05 [-0.11; 0.20] \\
Proportion stage II &  0.48 [0.46; 0.51] &   0.55 [0.41; 0.69] &  -0.26 [-0.44; -0.09] &   0.10 [-0.07; 0.27] \\
Proportion stage III &  0.51 [0.46; 0.56] &   0.15 [0.01; 0.29] &  -0.18 [-0.34; -0.02] &    0.14 [0.01; 0.27] \\
Proportion stage IV &  0.48 [0.44; 0.52] &  0.05 [-0.20; 0.30] &  -0.26 [-0.29; -0.23] &   0.12 [-0.00; 0.25] \\[0.15cm]
\textit{Cumulative regression} &&&&\\ 
Effect age on stage &  0.53 [0.49; 0.56] &   0.52 [0.50; 0.54] &    0.60 [-0.12; 1.31] &    0.36 [0.01; 0.71] \\[0.15cm]
\textit{Gamma regression} &&&&\\ 
Effect stage II on biomarker &  0.51 [0.47; 0.56] &   0.53 [0.44; 0.61] &    0.41 [-0.04; 0.85] &    0.42 [0.29; 0.56] \\
Effect stage III on biomarker &  0.50 [0.47; 0.54] &   0.47 [0.35; 0.59] &    0.33 [-0.05; 0.70] &    0.45 [0.31; 0.59] \\
Effect stage IV on biomarker &  0.50 [0.47; 0.53] &   0.34 [0.21; 0.47] &    0.25 [-0.03; 0.53] &    0.46 [0.26; 0.66] \\[0.15cm]
\textit{Logistic regression} &&&&\\ 
Effect age on death&  0.56 [0.49; 0.63] &   0.34 [0.15; 0.54] &    0.27 [-0.19; 0.73] &   0.21 [-0.14; 0.55] \\
Effect stage II on death &  0.52 [0.47; 0.57] &   0.28 [0.12; 0.44] &   -0.07 [-0.21; 0.07] &    0.38 [0.20; 0.55] \\
Effect stage III on death &  0.52 [0.46; 0.59] &   0.18 [0.03; 0.33] &   -0.10 [-0.26; 0.07] &    0.36 [0.18; 0.54] \\
Effect stage IV on death &  0.53 [0.48; 0.57] &  0.08 [-0.09; 0.25] &   -0.08 [-0.21; 0.04] &    0.42 [0.15; 0.69] \\
Effect therapy on death &  0.53 [0.48; 0.58] &   0.55 [0.39; 0.71] &   -0.04 [-0.27; 0.19] &   0.25 [-0.03; 0.52] \\
\bottomrule
\bottomrule
\end{tabular}
}
\end{table}

\begin{table}[htbp]
\caption{\label{tab:app_convergence_rate_bias_DP} Estimated exponent $a$ for the power law convergence rate $n^{-a}$ for the empirical bias.}
\resizebox{\columnwidth}{!}{
\begin{tabular}{lcccc}
\toprule
&& \multicolumn{3}{c}{\textbf{Generator}} \\
\cmidrule(r){3-5}
&  \textbf{Original} &  \textbf{PrivBayes} & \textbf{DP-GAN} & \textbf{PATE-GAN} \\
\textbf{Estimator, bias} &  &  &  & \\
\midrule
\textit{Mean} &&&&\\ 
Mean age  &   0.64 [0.40; 0.89] &   0.05 [-0.77; 0.87] &    0.63 [-0.45; 1.71] &   -0.39 [-1.21; 0.43] \\
Mean biomarker &   0.47 [0.17; 0.77] &   0.46 [-0.30; 1.23] &  -0.40 [-0.55; -0.26] &  -0.45 [-0.89; -0.00] \\[0.15cm]
\textit{Proportion} &&&&\\ 
Proportion therapy &   0.42 [0.09; 0.76] &   0.37 [-0.58; 1.32] &    0.01 [-1.21; 1.23] &   -0.19 [-0.67; 0.29] \\
Proportion death &   1.24 [0.64; 1.85] &   0.86 [-0.76; 2.49] &  -0.29 [-0.47; -0.12] &    0.35 [-0.23; 0.92] \\
Proportion stage I &  0.65 [-0.02; 1.32] &  -0.13 [-0.69; 0.43] &   -0.58 [-1.20; 0.04] &   -0.71 [-1.75; 0.33] \\
Proportion stage II &   0.92 [0.49; 1.36] &   0.15 [-0.41; 0.71] &    0.23 [-0.43; 0.90] &    0.18 [-0.25; 0.62] \\
Proportion stage III &  0.46 [-0.51; 1.43] &   0.11 [-0.31; 0.54] &   -0.06 [-0.68; 0.56] &   -0.30 [-0.85; 0.25] \\
Proportion stage IV &  0.21 [-0.10; 0.51] &  -0.24 [-1.15; 0.67] &   -0.33 [-0.75; 0.09] &   -0.10 [-0.74; 0.53] \\[0.15cm]
\textit{Cumulative regression} &&&&\\ 
Effect age on stage &   0.76 [0.27; 1.25] &  -0.03 [-0.12; 0.07] &    0.19 [-0.15; 0.53] &   -0.08 [-0.27; 0.10] \\[0.15cm]
\textit{Gamma regression} &&&&\\ 
Effect stage II on biomarker &  0.05 [-0.71; 0.81] &    0.03 [0.01; 0.05] &    0.07 [-0.03; 0.17] &   -0.03 [-0.05; 0.00] \\
Effect stage III on biomarker &  0.02 [-0.87; 0.90] &  -0.00 [-0.02; 0.02] &    0.03 [-0.00; 0.06] &   -0.06 [-0.12; 0.00] \\
Effect stage IV on biomarker &  0.37 [-0.29; 1.03] &   0.00 [-0.03; 0.04] &    0.01 [-0.01; 0.03] &   -0.08 [-0.22; 0.05] \\[0.15cm]
\textit{Logistic regression} &&&&\\ 
Effect age on death &  0.76 [-0.21; 1.74] &  -0.03 [-0.16; 0.09] &  -0.12 [-0.19; -0.04] &   -0.02 [-0.13; 0.09] \\
Effect stage II on death &  0.49 [-0.27; 1.25] &   0.36 [-0.50; 1.22] &   -0.07 [-0.22; 0.08] &   -0.21 [-0.56; 0.14] \\
Effect stage III on death &  0.57 [-0.31; 1.45] &   0.09 [-0.24; 0.43] &    0.03 [-0.18; 0.25] &   -0.12 [-0.32; 0.08] \\
Effect stage IV on death &   0.62 [0.45; 0.79] &   0.16 [-0.26; 0.59] &     0.11 [0.00; 0.21] &   -0.21 [-0.49; 0.07] \\
Effect therapy on death &   0.70 [0.53; 0.87] &  -0.12 [-0.27; 0.02] &   -0.32 [-0.65; 0.02] &    0.13 [-0.01; 0.27] \\
\bottomrule
\bottomrule
\end{tabular}
}
\end{table}

\end{document}